\documentclass[12pt,journal,compsoc,onecolumn,draftclsnofoot]{IEEEtran}   

\usepackage[utf8]{inputenc} 
\usepackage[T1]{fontenc}    

\usepackage{colortbl}
\usepackage{url}
\usepackage{color}
\usepackage{times}
\usepackage{epsfig}
\usepackage{booktabs}
\usepackage{hyphenat}
\usepackage{graphicx}
\usepackage{cite} 
\usepackage{tabularx}
\usepackage{epstopdf}
\usepackage{amssymb,amsmath,amsthm}
\usepackage{multirow}
\usepackage{hhline}
\usepackage{sidecap}
\usepackage{setspace}
\usepackage{cleveref}
\usepackage{subcaption}
\usepackage{pdflscape}
\usepackage{pdfpages}
\usepackage{longtable}
\usepackage{comment}
\usepackage{import}
\usepackage[ruled,vlined]{algorithm2e}
\usepackage{stackengine}

\newcommand{\norm}[1]{\left\lVert#1\right\rVert}
\newcommand{\abs}[1]{\left|#1\right|}
\newcommand{\enoisy}{$\varepsilon$-noisy~}
\newcommand{\CAPTION}[1]{\caption{\small{#1}}}

\newcommand{\PU}{\mathcal{P_U}}

\newcommand{\PS}{\mathcal{P_S}}
\newcommand{\PC}{\mathcal{P_C}}

\newcommand{\etal}{{\em et~al.}}

\long\def\symbolfootnote[#1]#2{\begingroup%
\def\thefootnote{\fnsymbol{footnote}}\footnote[#1]{#2}\endgroup}

\begin{document}

\title{Pictorial and apictorial polygonal jigsaw puzzles from arbitrary number of crossing cuts}

\author{Peleg Harel, 
        Ofir Itzhak Shahar,
        and~Ohad Ben-Shahar 
\IEEEcompsocitemizethanks{\IEEEcompsocthanksitem The authors are with the Department
of Computer Science, Ben-Gurion University of the Negev,Beer Sheva, Israel.
Corresponding author: ben-shahar@cs.bgu.ac.il}}

\date{}

\maketitle

\begin{abstract}
    Jigsaw puzzle solving, the problem of constructing a coherent whole from a set of non-overlapping unordered visual fragments, is fundamental to numerous applications, and yet most of the literature of the last two decades has focused thus far on less realistic puzzles whose pieces are identical squares. Here we formalize a new type of jigsaw puzzle where the pieces are general convex polygons generated by cutting through a global polygonal shape/image with an arbitrary number of straight cuts, a generation model inspired by the celebrated Lazy caterer's sequence. We analyze the theoretical properties of such puzzles, including the inherent challenges in solving them once pieces are contaminated with geometrical noise. To cope with such difficulties and obtain tractable solutions, we abstract the problem as a multi-body spring-mass dynamical system endowed with hierarchical loop constraints and a layered reconstruction process. We define evaluation metrics and present experimental results on both apictorial and pictorial puzzles to show that they are solvable completely automatically.
\end{abstract}

\newpage   



\section{Introduction}
\label{sec:introduction}

It happens often in real life that an orderless set of given fragments should be matched correctly (typically with no overlaps) to reconstruct a desired (known or unknown) coherent shape. Indeed, this broader (yet informal) problem description of the common jigsaw puzzle game fits countless real-world applications, including in (but not limited to)
archaeology~\cite{Willis2008ComputationalRO,kleber2009scientific,Sizikova2016WallPR}, 
biology~\cite{gassner1996test,marande2007mitochondrial}, 
earth sciences~\cite{lindstrom2019geological},
paleontology~\cite{Warren_etal_2014_Geology}, 
security and forensics~\cite{gao2010novel,ali2014development,gioe2017more}, artificial intelligence~\cite{Zhao2020aJigsaw},
speech processing~\cite{zhao2007puzzle},
document reconstruction~\cite{Kleber_etal_2009_ICDAR},
not to mention direct image editing~\cite{cho2010probabilistic} 
and artistic expressions in general~\cite{Taggart2018thisartist}.
While one may conceive jigsaw puzzles of more abstract form, here we will refer to puzzles as {\em visuals} if both their fragments (a.k.a. pieces), and the reconstructed ``wholes'', are ``visual'', namely if they can be understood, analyzed, and reconstructed by a visual system (and in particular, the human visual system). In practical terms, this means that both the pieces and the reconstructed whole are geometrical entities, possibly endowed with some pictorial overlay. If only the global geometric shape of the fragments is used in the process, the puzzle is called `{\em apictorial}'. If, on the other hand, pictorial information on the fragments is also used for the reconstruction, the puzzle is called `{\em pictorial}'.

While solving real-life jigsaw puzzles has occupied humans for millennia, to our best knowledge it was first introduced as a \textit{computational} task in 1964 by Freeman and Garder~\cite{freeman1964apictorial}, who discussed
the attributes of \textit{apictorial} jigsaw puzzles and proposed a solver for puzzles of \textit{unrestricted shapes}. Limited by the computation power of the time, results were of course constrained to very simple and small puzzles.
In the last two decades, however, the focus in the computational literature has shifted towards puzzles of \textit{square} pieces that must be matched into a rectangular image. Since the pieces in such puzzles are all shaped identically as squares, their pictorial content becomes \textit{the only} source of information available for the reconstruction. While starting modestly, the suggested solvers for such puzzles evolved rapidly in the past decade, and although they cannot guarantee optimal solutions, contemporary methods exist to solve "square jigsaw puzzles" of virtually any practical size.

As discussed below in the related work, the significant gap between unrestricted puzzles and square jigsaw puzzles was rarely addressed in the literature, although there are numerous methodological and applicative advantages for doing so.
In this work, we attempt to do exactly that. We introduce a different puzzle formation process, and a new class of puzzles termed here {\em crossing cuts polygonal puzzles}, that are inspired by the celebrated Lazy Caterer's sequence. Specifically, we consider global convex polygonal shapes (for apictorial puzzles) or convex polygonal images (for pictorial puzzles) that are sliced through by multiple straight cuts of arbitrary positions and directions, thus dividing the puzzle shape into many convex polygonal fragments. 
We discuss the synthesis of such puzzles, their properties with and without geometric noise, the qualitatively different reconstruction challenges they present for reconstruction, and a novel solver formulation that is based on abstracting the puzzle as a physical mechanical system. We discuss evaluation measures and present both qualitative and quantitative results on large and novel datasets that are made available to the community for future research.




\section{Related work}
\label{sec:related_work}

As mentioned above, the problem of puzzle solving is one where an orderless set of given fragments should be organized correctly with no overlaps to reconstruct a desired (known or unknown) global shape and possibly also with (typically unknown) pictorial content. 
In such cases, the pictorial data is yet another reconstruction cue whose degree of importance can vary relative to the apictorial (geometric) ones. 
Decades after Freeman and Garder's seminal paper~\cite{freeman1964apictorial} the problem was proved NP-complete~\cite{Demaine2007}, leading the literature to focus on devising heuristics, ad-hoc methods, and computational schemes that indeed cannot guarantee optimal solutions in tractable time but nevertheless facilitate successful puzzle solving in many cases, including large scale puzzles of various types.

\begin{figure}[!ht]
    \centering
    \includegraphics[width=0.6\columnwidth]{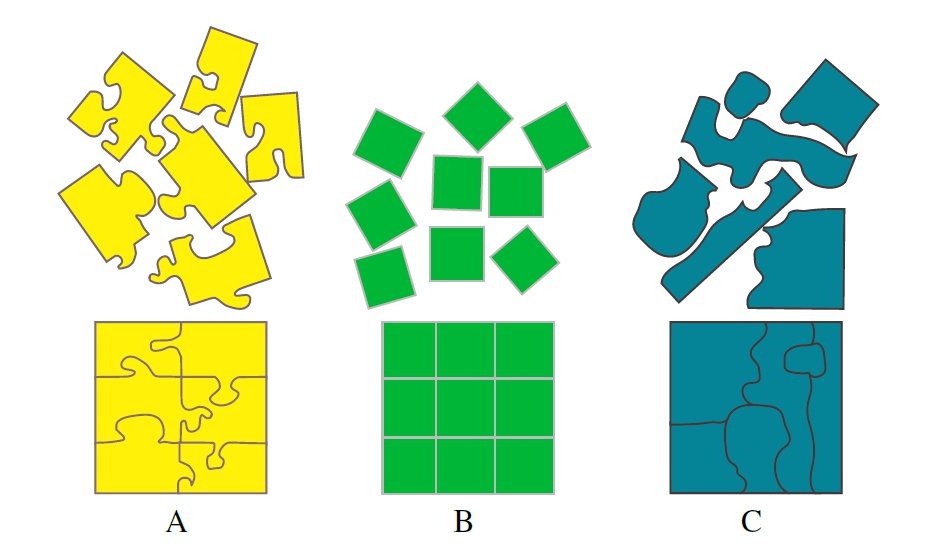}
    \CAPTION{
    The type of visual puzzle classes most studied in the literature. Shown are both the bag of pieces and the reconstructed puzzle. Sketched are only apictorial puzzles of the same global (in this case, rectangular) shape to emphasize the effect of the class on the possible geometry of the pieces.  
    \textbf{A:}~Commercial toy puzzle.
    \textbf{B:}~Square Jigsaw puzzle. Note that while the sketch focuses on their geometry, these puzzles must be pictorial to avoid trivial settings.
    \textbf{C:}~Unrestricted puzzle.
    }
    \label{fig:prior_art_puzzles}
\end{figure}

Broadly speaking, visual puzzles are generated by taking a coherent global (pictorial or apictorial) object and ``breaking'' it into a set of orderless or disorganized set of fragments. The details of this ``breaking'' are called in this paper the ``forward'' {\em puzzle generation process}, which 
sometimes incorporates additional actions before the puzzle is finalized, such as removing fragments, adding bogus ones, deforming fragments geometry, or distorting the pictorial content (what we will call ``geometric noise'' and ``pictorial noise'', respectively).%
With this in mind, the types of puzzles addressed in the prior art can be categorized into three different classes of such generation processes. We call these classes 
"Commercial toy puzzles", 
"Square Jigsaw puzzles", and
"Unrestricted puzzles"  
(see Fig.~\ref{fig:prior_art_puzzles}).
Since puzzle reconstruction algorithms attempt to ``reverse'' the puzzle generation process, the classification also implies that the reconstruction process may be done differently in each class. One evident example is Square Jigsaw puzzles, which unlike their counterparts \textit{must be} pictorial in order to escape trivial settings. In the following, we elaborate on each class, not necessarily according to their chronological order in the literature.

\subsection{Commercial toy puzzles}

Commercial toy puzzles, a set we denote $\PC$, include puzzles one can buy at toy stores and designated as a leisure time activity. As this type of puzzle is specifically meant to be solvable by humans, it follows a common set of very specific constraints and rules~\cite{goldberg2002global}. 
First, the outer border of the puzzle is rectangular. Second, the pieces in a reconstructed puzzle form a sort of rectangular grid so that all pieces except boundary ones have exactly four neighbors. Finally, pieces interlock with their neighbors by tabs (i.e., concave or convex protrusions), so that the shape of the matching tab allows a unique match\footnote{It is becoming more common in recent times to find commercial jigsaw puzzles that deviate from these rules. However, we are unaware of corresponding computational literature that addresses such variations and thus ignore them here.}. 

Although designed for humans, and very limited in their applications, Commercial Toy puzzles consume a fair share of the computational literature, and mostly after the mid-1980s. The uniqueness property suggests that a greedy approach can always solve such problems in low-order polynomial time simply by matching boundary curves. However, the need to scan the pieces and represent their boundaries numerically introduces geometric noise that leads to false positives during the search. 
With this in mind, the several toy puzzle solvers proposed in the literature share a common scheme.
First, the pieces are classified as either border or inner pieces by analyzing their boundary and counting straight segments. Border pieces are then assembled first (just as humans would tend to do), for example by reducing the problem to the traveling salesman problem and solving it via heuristic methods~\cite{wolfson1988solving}.  Once the border pieces are placed,
the dimensions of the puzzle grid can be deduced and the inner pieces are placed in a grid using either a greedy or an exhaustive search method.
Because noise could generate false positives, each piece placement
involves a test for geometric violations (e.g., overlaps between pieces), a type of event that results in backtracking.
Although the use of backtracking can entail exponential complexity,
the shape of the tab is expected to be unique enough to make false positives rare (or even impossible), thus retaining a tractable solution process.

Given that the matching geometry is unique, $\PC$ puzzles need not contain pictorial content for a computer (or for that matter, even humans) to solve, as indeed was the case in several prior studies on the topic
\cite{wolfson1988solving,burdea1989solving,webster1991isthmus,bunke1993jigsaw,de2004constructing,goldberg2002global}.
%
That being said, pictorial extensions \textit{do} exist, addressing the full real-life toy jigsaw puzzle challenge (except for the fact that toy puzzles usually include the reconstructed image printed on their box cover). Such pictorial toy puzzle solvers
\cite{kosiba1994automatic,chung1998jigsaw,yao2003shape,nielsen2008solving}
can clearly utilize an additional constraint of visual coherence (e.g., continuity) across neighboring pieces to improve the accuracy of potential mates, lower the risk of false positives, and thus reduce the search space.
Thus far, the biggest pictorial toy puzzle solved this way was sized at 320 pieces~\cite{nielsen2008solving}.
Unfortunately, given the possibility of solving such puzzles perfectly in tractable time, no performance metrics (other than testing for perfect reconstruction) or statistical benchmark experimentation are typically performed.

\subsection{Square jigsaw puzzles}

Square Jigsaw puzzles, denoted $\PS$, are the type of visual puzzles discussed most frequently in the last two decades.  They are the simplest geometrically and based on a generation process that cuts a rectangular image into a grid of identically shaped square pieces. The problem is considerably different than the commercial toy puzzles since with identical boundaries to all pieces, the reconstruction must fully rely on the pictorial content. 

Square Jigsaw puzzles also tend to share a similar algorithmic flow. 
First, a measure of dissimilarity between every two potential neighbors is pre-calculated. Then, the dissimilarity is used to derive neighbors' compatibility scores to represent the confidence that they should be paired. Then neighboring pieces are matched, placed, and aggregated to maximize global compatibility, either in a greedy fashion or by employing heuristics to globally search the solution space. 
Since state-of-the-art square-piece puzzle solving now tends to deal with rather large-scale problems, backtracking is typically avoided as the number of search paths is intractable.
The many variants of this general scheme include solvers for Square Jigsaw puzzles with pieces of \textit{known} size and piece orientation
\cite{toyama2002assembly,fei2007image,zhao2007puzzle,murakami2008assembly,alajlan2009solving,cho2010probabilistic,pomeranz2011fully,yang2011particle,sholomon2013genetic,adluru2015sequential,andalo2016psqp},
puzzles where the orientation of the pieces is \textit{unknown}
\cite{gallagher2012jigsaw,mondal2013robust,sholomon2014generalized,son2014solving,yu2015solving,son2016solving,son2018solving,rika2019novel},
challenges with mixed set of pieces from multiple puzzles
\cite{gallagher2012jigsaw,mondal2013robust,paikin2015solving,son2016solving,son2018solving},
missing pieces
\cite{gallagher2012jigsaw,mondal2013robust,paikin2015solving,son2016solving,son2018solving}
noisy pictorial content
\cite{mondal2013robust,brandao2016hot,rika2019novel,son2014solving,yu2015solving,son2018solving},
gaps between pieces
\cite{paumard2020deepzzle,derech2021solving},
and even restricted deformations to the shapes of fragments~\cite{gur2017square}.

Indeed, earlier attempts to address the problem assumed known dimensions, known piece orientation, and even some prior knowledge regarding the solution.
For example, Cho~\etal~\cite{cho2010probabilistic} used 
prior knowledge in the form of ground truth anchor pieces or low-resolution images of the solved puzzle. Color differences along abutting piece boundaries were used for the compatibility score and the reconstruction was based on achieving maximum likelihood for both piece compatibility and the prior knowledge. 
%
Shortly after, Pomeranz~\etal~\cite{pomeranz2011fully} were the first to solve the Square Jigsaw puzzle fully autonomously and \textit{without} any prior knowledge (except for the puzzle dimensions) and increased the size of solvable puzzles an order of magnitude over the prior art to puzzles of thousands of pieces.
Among the contributions were an iterative greedy approach, prediction of pictorial content across piece boundary for the dissimilarity metric, and the introduction of the {\em best-buddies} concept that influenced much of the later works and served as a precursor for the employment of loopy constraints to reduce the search space (see below).
Sholomon~\etal~\cite{sholomon2013genetic} introduced a different type of solver based on a genetic algorithm and the best-buddies idea, a combination that proved successful in solving even larger puzzles, exceeding the likely capacity of human solvers.

Pieces of unknown orientation add another layer of complexity to the Square Jigsaw puzzle problem, as the number of possible configurations increases by a factor of $4^K$ (for puzzles of $K$ pieces). Gallagher~\cite{gallagher2012jigsaw} was the first to tackle such puzzles while introducing a gradient-based dissimilarity score and a greedy spanning tree-based solver.
Yu~\etal~\cite{yu2015solving} also dealt with unknown orientation
by using a global optimization in the form of relaxed linear programming, and Sholomon~\etal~\cite{sholomon2014generalized} modified their original genetic algorithm to also handle unknown orientations in puzzles with a very large number of pieces.

Noise in the pictorial data increases the difficulty of the problem since the  dissimilarity metric becomes less reliable and false matches are even more likely, prompting some studies to seek more robust
compatibility metric (e.g., \cite{mondal2013robust,brandao2016hot,rika2019novel}).
Toward that end, Mondal~\etal~\cite{mondal2013robust} combined two previously used dissimilarity metrics while Brandão and Marques~\cite{brandao2016hot} measured the dissimilarity using a
heat-based affinity measure that utilizes a pixel environment larger than the piece boundary.
Rika~\etal~\cite{rika2019novel} used deep learning as a mechanism to assess the compatibility between pairs of pieces, taking the whole piece as input. 
Taking a different approach, Yu~\etal~\cite{yu2015solving} and Son~\etal~\cite{son2014solving,son2018solving} dealt with noise by applying a reconstruction algorithm that demands a consensus in an environment larger than the immediate neighbors of each piece.
The former used a relaxed linear programming algorithm that rewards global piece consensus while the latter introduced the notion of loopy constraint - a requirement for compatibility consensus in loops of pieces.

Present-day state-of-the-art solvers for the Square Jigsaw puzzle can solve puzzles with over $20,000$
pieces~\cite{sholomon2014generalized}. For historical reasons, most of the prior art experimented with square pieces of $28{\times}28$ pixels in size in order to allow enough pictorial data while measuring the compatibility of pieces. However, recent works now extend this convention to pieces as small as $7\times 7$ pixels \cite{son2014solving,son2018solving}.

\subsection{Unrestricted puzzles}

Unrestricted puzzles, the class we denote $\PU$, contain puzzles that do not have a formal generation process or constraints on the shape of their pieces.
In such puzzles,  the representation of the pieces is far more complex (than $\PS$ or $\PC$),
they can be matched to an arbitrary number of neighbors abutting arbitrary sections of their boundary,
and the reconstruction of such 2D puzzles can be described as a general planar adjacency graph of arbitrary maximal degree (unlike the degree 4 that characterizes the reconstructions of 2D puzzles in $\PC$ or $\PS$). 
Despite these complications, and somewhat unexpectedly, the very first work on computational puzzle
solving~\cite{freeman1964apictorial} belongs to this class.

\textit{Apictorial} unrestricted puzzle solvers typically use curve matching to find potential matching pieces
\cite{freeman1964apictorial,radack1982jigsaw,kong2001solving}.
As mentioned above, the first to explore such an approach (or computational puzzle solving in general) were Freeman and Garder~\cite{freeman1964apictorial}, who also introduced a solver capable of dealing with a large variety of piece shapes and junction types. 
Their solver matches curves using a chain encoding scheme and then assembles
the puzzle using a greedy algorithm that backtracks on errors, 
an exhaustive scheme possible only because of the very small scale problems considered. The solver tried to reconstruct coherently around junctions, thus seeking neighbors with loopy consensus, perhaps leading the way to the future use of loopy constraints in the field~\cite{son2014solving}.
Owing to the small computational resources of the time, the single evaluation on a 9-piece puzzle of highly discriminated pieces did not permit later experimental comparison to contemporary contributions.
Forty years later, Kong and Kimia~\cite{kong2001solving} used a coarse-to-fine approach to curve matching and a greedy merging of piece triplets and backtracking upon spatial overlap. While the geometrical treatment was significantly more rigorous, here too the experimentation was limited to few puzzles of up to 25 pieces, most of which had near-convex low-order polygonal shapes. An interesting question that emerges is whether or not such data represent realistic scenarios, at least on average. Of course, it is difficult to address such questions without some formal puzzle generation model, an observation that is key to the suggested research in this paper (see below).

Extending the basic computational flow of the above, solvers for unrestricted \textit{pictorial} puzzles
\cite{tsamoura2009automatic,liu2011automated,makridis2006new,saugirouglu2010optimization,zhang2014graph,le2019jigsawnet}
use the pictorial content as well as geometrical boundaries to match pieces and reconstruct the puzzle.
Sağıroğlu and Erçil~\cite{saugirouglu2010optimization} used an extrapolation method to approximate the content of the pictorial data in a band around each piece. This allowed for a pictorial score by comparing the extrapolated bands to the content of prospective neighbors. Then, the Fourier transform translation property was used to find an alignment that maximizes the correlation between pieces while satisfying the geometrical constraints.
The reconstruction itself was done in a greedy fashion, starting from a random configuration and improving the global score one piece at a time. To escape local minima, the reconstruction process was restarted multiple times with different random seed configurations. The experimental evaluation was limited to assemblies of 21 pieces, most of which had very distinctive boundaries. A related approach with fragment extrapolation for registration of neighboring candidates was proposed by Derech~\etal~\cite{derech2021solving}.

Recently, Le~\etal~\cite{le2019jigsawnet} introduced a novel approach for fragment matching using a Convolutional Neural Network that utilizes both boundary shape and pictorial data with a hierarchical loops approach for the reconstruction. The solver was tested successfully on puzzles of up to 400 pieces, significantly bigger than prior work. Moreover, evaluation was performed quantitatively and on a relatively large number of puzzle problems, two advances over the prior art in the unrestricted puzzle literature.  That being said, the test data published alongside the paper contains a relatively constrained shape for the pieces as all of them were roughly perturbed rectangles. 

It should be mentioned that much work on (typically apictorial) unrestricted puzzles is performed in the archaeological domain, where computational tools have generated a revolution~\cite{grosman2016reaching} and visual puzzles are typically not 2D, but either 2.5D (``thick'' 2D manifolds) or 3D. For their different focus we omit a detailed review of that literature, referring the reader to selected references from the last two decades~\cite{Papaioannou2001VirtualAA,Papaodysseus2002ContourshapeBR,Papaioannou2003OnTA,koller2006computer,Huang2006ReassemblingFO,Willis2008ComputationalRO,Kleber_etal_2009_ICDAR,Mellado2010SemiAutomaticGR,TolerFranklin2010MultifeatureMO,Castaeda2011GlobalCI,Funkhouser2011LearningHT,Oxholm2011ReassemblingTA,Brown2012ToolsFV,shin2012analyzing,Palmas2013ACC,pintus2014geometric,Mavridis2015FracturedOR,andalo2016automatic,brandao2016hot,Li2020PairwiseMF, Yilmaz_2022_Survey,Markaki2023}. That being said, the computational flow in most of these studies is similar and constitutes several steps, including scanning the artifacts to point clouds, processing these point clouds into meshes, segmenting the meshes to facets, and extracting geometrical features either on the facets or their boundary curves. Facets of different fragments are then registered through the raw geometrical point cloud data (e.g., using ICP) or the processed features. The final stage of combining the pairwise matches into a global assembly is done with a variety of methods, often including human intervention. 

\subsection{A missing link in the puzzle solving chain?}

The scientific background covered above suggests that even though the basic problem is one, research into visual puzzle solving has been conducted in ``parallel tracks'' that affected progress in ways that are not necessarily optimal. Related to this are observations like the following

\begin{itemize}

\item Solving commercial jigsaw puzzles computationally is very anecdotal in terms of its application value and serves mostly as an intellectual challenge.

\item Markedly inconsistent with the popularity of $\PS$ in the literature, there are almost no real-life applications that can be abstracted as strict Square Jigsaw puzzles. For example, although most studies of Square Jigsaw puzzles cite archaeology as a possible application domain, archeological puzzles are virtually never square, and to our best knowledge there is exactly one case in the computational archeology literature for which Square Jigsaw puzzle solvers may be relevant\cite{brandao2016hot}.

\item Unrestricted puzzles do have the potential to serve numerous applications, but their unrestricted generation model makes it difficult to study them rigorously, obtain useful insights, or allow any type of guarantees related to the solution process. Rigorous analysis means answering questions about puzzle properties (e.g., the expected number of pieces in a puzzle, the statistical properties of the area of puzzle pieces, etc.) and about the developed algorithms (e.g., how many false positive matches may occur to determine the worst time complexity of a particular algorithmic step). Such understanding of the problem or the solutions is typically missing for this class.

\item The fact that $\PU$, $\PS$, and $\PC$ are so different makes it practically impossible to apply tools (e.g., solvers or performance measures) from one class to another, even though both $\PS$ and $\PC$ are subsets of $\PU$ (and under certain relaxation one can even view $\PS$ as a subset of $\PC$). This gap manifests itself not only at the level of algorithms, but also in the representation of the problem (e.g., for I/O), data structures and formats used, and operational assumptions.
\end{itemize}

Generally speaking, a trade-off emerges between how constrained a puzzle class is, how relevant it is for real-life applications, how rigorous the analysis it permits, and how applicable are its solvers to other classes. 
Our present work tries to address this trade-off by suggesting a new puzzle generation model that is more restricted than $\PU$ puzzles and thus allows rigorous analysis while being much more general than $\PS$  and thus extends the applicability and usability to real-life challenges. In general, such an approach may refer to a new class of puzzles one may term {\em restricted modeled puzzles}, where some formal restrictions are defined for the puzzle generation process in a way that rigorous analysis is still possible while the range of applications remains viable. We hope that the present research will encourage the community to explore this direction further.




\section{Model formulation}
\label{sec:the_puzzle}

Recall that the pieces (or fragments) of square jigsaw puzzles are all identical in shape, a setup that drives all reconstruction decisions to be solely pictorial. However, real-world puzzles usually have pieces of a more general form  (e.g.,~\cite{shin2012analyzing}), leading to a different set of challenges. Here we try to formulate a new class of puzzles that is both general enough for more real-world cases and yet formal enough for rigorous analysis and exploration. We call this class the {\em crossing cuts puzzles}.

A crossing cuts puzzle is created by cutting through a convex polygon\footnote{We note that one could apply crossing cuts to an arbitrary \textit{non polygonal} convex shape, but then the curved edges would serve as a major clue for reconstruction, a relief we preferred to avoid in this work.} with $a \in \mathbb{N}$ arbitrary (random) straight cuts  $Cuts = \{c_1, \dots c_a\}$. The pieces of such puzzles are thus convex polygons where every piece (except border pieces) has a single neighbor along each of its edges. 
This puzzle generation model is inspired by the procedure that produces the Lazy Caterer's sequence\footnote{Each number $f(n)$ in the Lazy caterer's sequence, also known as the central polygonal numbers, is the {\em maximal} number of cake pieces a caterer can obtain by cutting the cake (or more abstractly, a disk) exactly $n$ cuts. To do so a caterer must be ``lazy'' since the cuts cannot all intersect the center of the disk, as is usually done while slicing cakes.}~\cite{wetzel1978division,Yaglom_1987_Challenging_Math_Problems}, but unlike the latter, in our case, the cuts are completely arbitrary and there is neither guarantee nor desire to maximize the number of pieces\footnote{In some sense, the caterer in our case is even ``lazier'' than the ``lazy caterer'', as she does not need to take measures to choose her cuts to maximize the number of pieces.}. This proposed model can also be used to address and simulate the \textit{realistic} and/or \textit{physical} generation of puzzles already discussed in literature, such as square jigsaw puzzles. For example, while using a pair of scissors or a ruler and a blade, one can create a real-life square jigsaw puzzle by cutting a picture, where cuts are not strictly parallel or equidistant due to human error or lack of sensitivity. The result will be a \textit{noisy} square jigsaw puzzle (see below), which is a crossing cut puzzle.

Geometrically, square piece puzzles are indeed a very special case of crossing cuts puzzles and thus the latter require a more general mechanism to represent them. Towards that end, and inspired by Freeman and Garder~\cite{freeman1964apictorial}, we define the \textit{mating graph} to be a planar graph whose nodes are the edges of the puzzle pieces and whose links\footnote{To avoid terminological confusion, we use the term 'links' for the edges of the mating graph, while we reserve the term 'edges' for the boundary segments of puzzle pieces.}, dubbed \textit{matings}, represent immediate neighborhood relationship. The connected pieces will be called \textit{neighbors} or \textit{neighboring pieces} while the edges matched by a mating will be called \textit{mates}. Note that in the ideal case, when no geometric noise is present, a mating in the solved puzzle represents two overlapping mates with identical lengths.

\begin{figure}[h!]
    \centering
    \includegraphics[width=0.9\columnwidth]{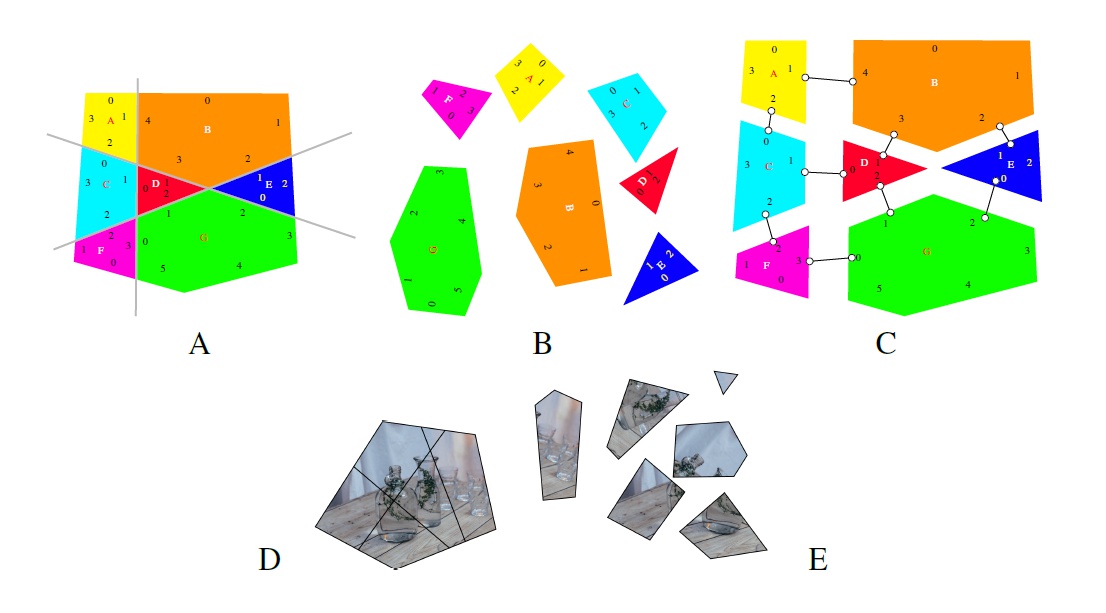}
    \CAPTION{
        The elements of a crossing cuts puzzle.
        \textbf{A:}~The puzzle is created by cutting a convex polygon using multiple (here 3) straight lines.
        \textbf{B:}~The puzzle problem constitutes of an un-ordered and arbitrarily transformed set of pieces.
        Note that different pieces may vary vastly in size (e.g., compare pieces $p_B$ and $p_D$.
         \textbf{C:}~The mating graph matches pairs of edges of two different pieces and here it includes
         $\{
         \{ e_A^1 , e_B^4 \},\allowbreak
         \{ e_A^2 , e_C^0 \},\allowbreak
         \{ e_B^2 , e_E^1 \},\allowbreak
         \{ e_B^3 , e_D^1 \},\allowbreak
         \{ e_B^3 , e_D^1 \},\allowbreak
         \{ e_C^1 , e_D^0 \},\allowbreak
         \{ e_C^2 , e_F^2 \},\allowbreak
         \{ e_D^2 , e_G^1 \},\allowbreak
         \{ e_E^0 , e_G^2 \},\allowbreak
         \{ e_F^3 , e_G^0 \}
         \}$.
         Note that pieces end up having different numbers of edges and thus different numbers of mates.
        \textbf{D:}~A pictorial crossing cuts puzzle is one with an image (or any other visual content) covering the polygon
                    that is being cut.
        \textbf{E:}~Once cut and shuffled, the pictorial pieces represent the puzzle to be solved. 
         }
\label{fig:cc_df}
\end{figure}

Unlike in square piece (and also commercial toy) jigsaw puzzles, which have a constant number of neighbors for each piece (except boundary pieces), the mating graph of a crossing cuts puzzle is more general since the number of matings a piece can have varies (see Fig.~\ref{fig:cc_df}A-C).
Moreover, the number of possible Euclidean transformations of the pieces of crossing cuts puzzles adds additional complexity, since unlike for square or toy puzzles, it is infinite in cardinality and selected from a continuous range.
Hence, on the one hand, the representation of the puzzle must account for these new degrees of freedom. On the other hand, the geometrical shape of the pieces provides more information that is not present in the square jigsaw problem and may facilitate reconstruction algorithms that rely only on the shape of the pieces.
Just as in any other type of puzzles (see Sec.~\ref{sec:related_work}), an \textit{apictorial} crossing cuts puzzle is one whose initial polygon contains no visual content
(Fig.~\ref{fig:cc_df}A,B) 
while a \textit{pictorial} crossing cuts puzzle is generated from a polygon covered with an image.
(Fig.~\ref{fig:cc_df}D,E). 
In this paper, we start our analysis with apictorial puzzles and gradually incorporate pictorial aspects while arguing that under most realistic scenarios, pictorial content is critical for successful and efficient crossing cuts puzzle solvers.

\begin{figure}[!ht]
    \centering
    \includegraphics[width=0.8\columnwidth]{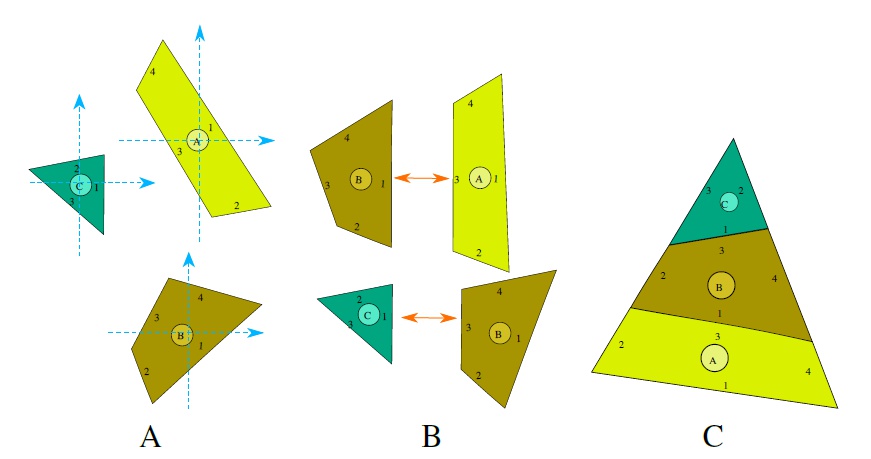}
    \CAPTION{
        The representation of a crossing cuts puzzle and its solution,
        illustrated here for the simplest 2 cuts (and 3 pieces) puzzle.
        {\bf A:}~Each piece $\{p_1,p_2,p_3\}$ is represented by its
                  vertices and edges in some arbitrary Euclidean
                  coordinate system (which conveniently may be centered at the center of mass).
        {\bf B:}~Each mating pairs two edges of two different pieces. In our case it takes just the matings $\{ \{ e_A^3 , e_B^1 \}, \{e_B^3,  e_C^1\}\}$.
        {\bf C:}~After the application of the Euclidean transformations
                 $(t_i, R_i)\quad \forall i = 1..3$, the puzzle is reconstructed (up to some global Euclidean transformation).}
\label{fig:cc_el_df}
\end{figure}

To facilitate a constructive discussion towards computational solutions to our problem, one needs to differentiate the representation of the puzzle itself (in the sense of the riddle to solve) and its possible solutions.
A crossing cuts \textit{puzzle} is thus a representation of the unordered puzzle pieces after the complete polygon was cut (Fig,~\ref{fig:cc_df}B,E). 
Formally, let $P = \{p_1, \dots p_n\}$ be a set of \textit{pieces}, where each  $p_i$ is a convex polygon of $N_i \geq 3$ vertices. By convention, we order these vertices clockwise around the polygon's center of mass and denote them
\begin{align*}
V_i = \left\{\vec{v\,}_i^{1}, \vec{v\,}_i^{2},\dots , \vec{v\,}_i^{N_i}\right\} \;.
\end{align*}
Correspondingly we label the piece edges between these consecutive vertices by
{\small
\begin{align*}
	E_i =
        \left\{ {e_i^1} , {e_i^2} ,\ldots, {e_i^{N_i}} \right\}
        =
        \left\{ (\vec{v\,}_i^1, \vec{v\,}_i^2 ) ,
                (\vec{v\,}_i^2, \vec{v\,}_i^3 ) ,\ldots,
                (\vec{v\,}_i^{N_i}, \vec{v\,}_i^1 ) \right\} \;.
\end{align*}
}
A \textit{solution} to a crossing cuts puzzle requires positioning each piece in its ''correct'' position relative to all other pieces, and while this requires the determination of a Euclidean transformation (position and rotation) for each piece, in practice this will first require to resolve the neighborhood relationships between the pieces, i.e., the ''correct'' mating graph. An algorithm to obtain a solution thus needs to determine both
\begin{itemize}

\item[i.] the pairwise matings $M = \left\{m_1, \dots m_{|M|}\right\}$ of all pieces, i.e., all unordered pairs of edges $m_q=\{ e_i^j, e_k^l \}$ of two different pieces  that should be matched (and in an ideal setting, truly overlap) in order to reconstruct the puzzle, and

\item[ii.]  the 2D Euclidean transformation of each piece $p_i$, from its given input representation $V_i$ to the one in the reconstructed puzzle.
The transformation of piece $p_i$ involves a translation $t_i \in \mathcal{R}^2$
and a rotation $R_i \in \mathcal{S}^1$.
With the rotation typically represented by an orthonormal matrix
$R_i \in \mathcal{R}^{2 \times 2}$,
the pose of the piece in the reconstructed puzzle will be
\begin{align*}
p_i' = \left\{R_i \cdot \vec{v\,}_i^1 + \vec{t\,}_i, R_i \cdot \vec{v\,}_i^2 + \vec{t\,}_i,\ldots , R_i \cdot \vec{v\,}_i^{N_i} + \vec{t\,}_i\right\}
\end{align*}

\end{itemize}
Fig.~\ref{fig:cc_el_df} illustrates both the puzzle and the aspects of its solution as just discussed. It should be noted that while the mating graph may have only one ``correct'' solution, the Euclidean transformations of the pieces can be correct up to some global Euclidean transformation that describes a rigid motion of the entire reconstructed puzzle.

\section{Mating constraints and a greedy solver}
\label{sec:mating_constraints}

\begin{figure}[!h]
  \centering
  \fontsize{6pt}{7pt}\selectfont
    \includegraphics[width=0.5\columnwidth]{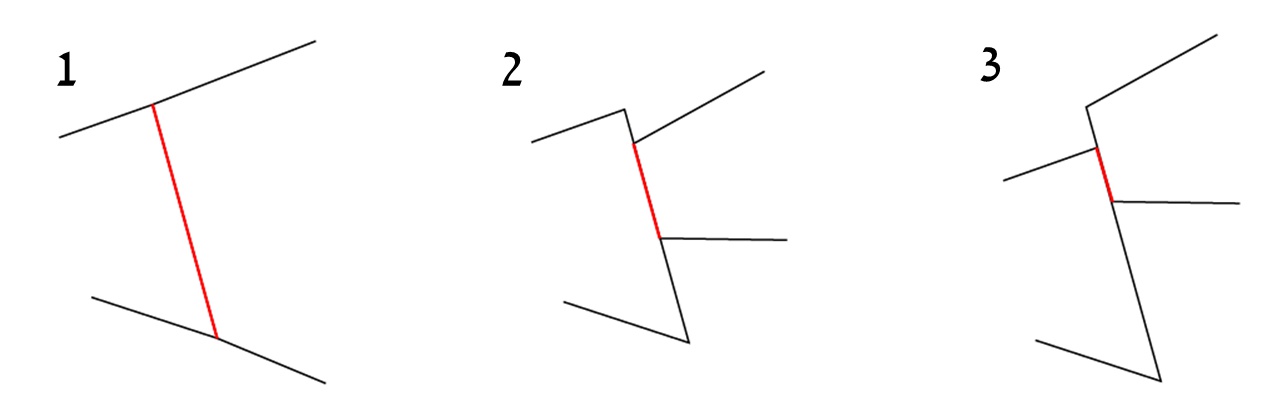}
    \CAPTION{
    In (generic) crossing cuts puzzles only matings of type 1 are allowed, while configurations of type 2 or 3 are prohibited.
    In particular, mates must be of equal length and overlap.
    }
  \label{fig:typeA}
\end{figure}

\begin{figure}[!h]
  \centering
  \fontsize{6pt}{7pt}\selectfont
    \includegraphics[width=0.5\columnwidth]{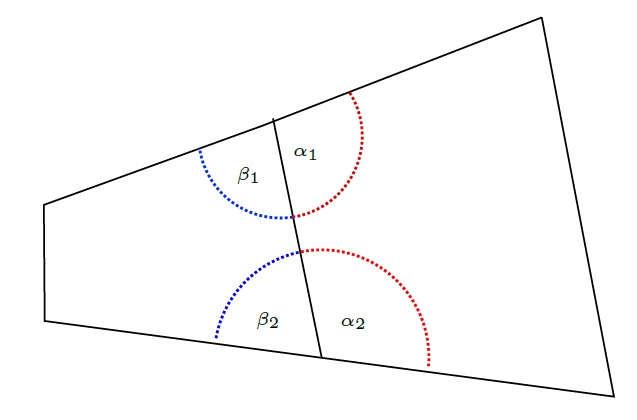}
    \CAPTION{
    The mate angle constraint dictates
    $\alpha_1 + \beta_1 = \alpha_2 + \beta_2=\pi$.
    }
  \label{fig:angle_const}
\end{figure}

With the crossing cuts puzzles defined as above, and assuming no noise, idealized infinite precision in the representation of the geometrical objects, and random uniform distribution of the crossing cuts themselves, it is immediate to observe that the probability of 
(i) more than two crossing cuts to meet at a point, and 
(ii) having more than two edges with \textit{identical} lengths, 
is nil in both cases. These properties of the \textit{generic} (i.e., non-accidental) puzzle entail two key constraints for the formation of plausible matings:
\begin{enumerate}

\item[$C_1$:] \textbf{The mate length constraint:}
Since plausible matings should match complete edges, it follows that they must match mates of the very same length (see Fig.~\ref{fig:typeA}).

\item[$C_2$:] \textbf{The mate angle constraint:} Since the mates of plausible matings have vertices emerging from just 2 crossing cuts, their adjacent edges must form a straight line (which simply overlaps with two different crossing cuts). It follows that the two pairs of adjacent angles of the neighboring pieces must complete to $\pi$, i.e., be supplementary (see Fig.~\ref{fig:angle_const}).

\end{enumerate}

In the following, we will  refer to the mating constraints also as predicates, i.e.,
\begin{align*}
\forall i\in\{1,2\} \quad C_i\left(e_i^j, e_k^l\right) = \text{true} \; \Leftrightarrow \;  e_i^j\text{~and~}e_k^l \text{~satisfy~} C_i \;.
\end{align*}
%
%
%
%
%
Clearly, the constraints just outlined entail the simple and \textit{greedy} solver in Algorithm~\ref{alg:naive} that progressively moves pieces from the set $U$  of unassigned pieces to the set $R$ of the reconstructed assembly, while forming the mating graph $G_M$. This simple algorithm uses only constraint $C_1$, but versions using $C_2$ are possible also. Either case, these greedy schemes are clearly sound, complete, and tractable, they do not need pictorial information and they can solve both apictorial and pictorial puzzles using the geometric information alone. As we discuss shortly, all this changes fundamentally once we introduce geometric noise to the puzzle.

\begin{algorithm}
\small
    $G_M \leftarrow \emptyset$ \;
    $R \leftarrow$ an initial random piece from $P$ \;
    $U \leftarrow  P \setminus R$ \;
    
    \While{$U \neq \emptyset$}{
        Pick an edge $e_i^j$ of a piece $p_i \in R$ that is not yet part of a mating \;
        \eIf {(there is an edge $j_k^l$ of some piece $p_l \in U$ that has the same length as $e_i^j$)}{
            $R \leftarrow  R \cup \{ p_l \}$ \;
            $U \leftarrow U \setminus \{ p_l \}$ \;
            Compute the Euclidean transformation that places $p_l$ properly in the reconstructed assembly \;
            $M_p \leftarrow$ ~matings of $p_l$ edges that overlap edges in $R$\;
            $G_M \leftarrow G_M \cup M_p$ \;
        }{
            do nothing (because the edge $j_k^l$ is a border edge)  \; 
        }
    }
    Return $G_M$ and the corresponding Euclidean transformations of all pieces.
 \CAPTION{
    A basic greedy algorithm for solving crossing cuts puzzles under ideal conditions
    }
\label{alg:naive}
\end{algorithm}




\section{Noisy crossing cuts puzzles}

Real-world data, its measurement, or its representation, are never completely accurate. Even if the measurement or the digital representation of the pieces were devoid of errors, \textit{real life} crossing cuts puzzles (or geometric puzzles in general) may incorporate deformations to the shapes of the pieces, as well as to their visual content (in pictorial puzzles). In fact, geometric noise also affects how pictorial information can be leveraged, even if no pictorial noise is present. For this reason we begin by formalizing the geometric noise, and extend the discussion to pictorial puzzles only later. 

Clearly, geometric noise can be modeled in many different ways, though one particular appealing is material degradation, and thus piece shrinkage, a process clearly relevant for applications involving physical pieces (e.g. in archaeology). To incorporate material degradation without escaping the crossing cuts framework, we model this deformation process by preserving the number of vertices of each piece, but shifting each of them \textit{inward} by a random distance that is distributed (in our case, uniformly) in a given range. We note that the particular distribution of such noise may affect certain statistical properties (see Chapter~\ref{chap:statistic} below), but otherwise it is less significant for the reconstruction algorithm discussed later.

\subsection{Noise formulation}
\label{sec:noise}

\begin{figure}[h!]
    \centering
    \includegraphics[width=0.6\columnwidth]{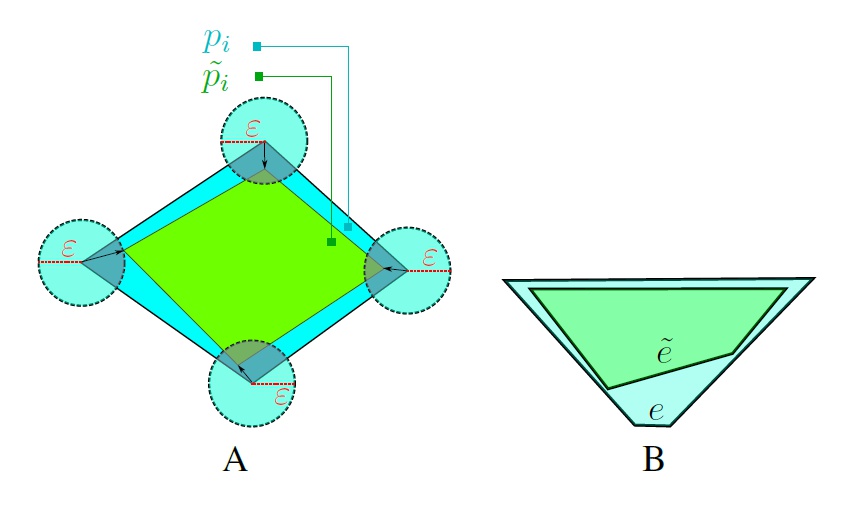}
  \CAPTION{
    The effect of noise on edge length.
    {\bf A:} Each of the vertices of a piece $p_i$ is perturbed inwards along a uniformly distributed direction $\measuredangle \vec{\epsilon\,}_i^j$ and as far as a uniformly distributed distance $||\vec{\epsilon\,}_i^j||$ to create the \enoisy piece $\tilde{p}_i$.
    {\bf B:} A case where edge $e$ increases in size after the application of noise, even though all the vertices collapsed inwards to end up as $\tilde{e}$.
    Clearly, $||\tilde{e}||$ is bounded by $||e||+2\varepsilon$.
  }
  \label{fig:noise_bound}
\end{figure}

Formally,  a vertex $\vec{v\,}_i^j$ of piece $p_i$ is perturbed inwards by a distance $\vec{\epsilon\,}_i^j$ that is bounded by some maximal noise level $\varepsilon \geq 0$. It is convenient to set that bound relative to a reference value that is based on the puzzles' geometrical properties. In our case, we use the puzzle diameter $D$, i.e., the distance between the furthest vertices. Formally, we define a relative bound $\xi$ that sets the absolute noise level at $\varepsilon=\xi \cdot D$, and let $\bar{\xi}$ be the corresponding noise level relative to the average edge length (to be derived later). An original piece $p_i = \left\{\vec{v\,}_i^1, \dots \vec{v\,}_i^{N_i} \right\}$
ends up as the \mbox{\textit{\enoisy}} piece
$ \tilde{p}_i =  \left\{ 
\vec{v\,}_i^1 + \vec{\epsilon\,}_i^1,
\dots,
\vec{v\,}_i^{N_i} + \vec{\epsilon\,}_i^{N_i}
\right\}$
where the noise magnitude
$\left\lVert \vec{\epsilon\,}_i^j \right\rVert \sim \text{U}(0,\varepsilon)$
and the noise direction is selected from the sector originating from $\vec{v\,}_i^j$ towards the two nearby vertices, namely
$\measuredangle \vec{\epsilon\,}_i^j \sim \text{U}
\left(
	\measuredangle \left(
	\vec{v_i^{j - 1}\,} -
	\vec{v_i^j\,}
	\right)
	,
	\measuredangle \left(
	\vec{v_i^{j + 1}\,} -
	\vec{v_i^j\,}
	\right)
  \right)
$.
This limits the random perturbation angle of $\vec{\epsilon\,}_i^j$ and constrains it to be \textit{inward}, i.e.,  an erosion-like process ``into'' the material.
Fig.~\ref{fig:noise_bound}A illustrates how such noise could affect the shape of a quadrilateral (4-edges) piece.

Naturally, the incorporation of noise affects the validity of our constraints on mating. In particular, the number of potential mates now increases drastically and far from uniqueness, and the implications on a reconstruction algorithm are paramount. In this sense, $C_1$ and $C_2$ must be revised, as discussed next.

\subsection{$\tilde{C}_1$ : Mate length constraint under noise}
\label{Sec:noisy_length_C}

Since now plausible matings should match edges that have been perturbed
differently, the mate length constraint must be relaxed to accommodate these independent perturbations. 
Let $e$ and $e'$ be the matching edges before applying the noise
while $\tilde{e}$ and $\tilde{e}'$ denote their corresponding \enoisy edges. 
It follows that $\tilde{e}$ and $\tilde{e}'$ might have respective lengths $\tilde{L}$ and $\tilde{L}'$ that satisfy
\begin{align}
|\tilde{L} - \tilde{L}'| \leq 4 \varepsilon \;.
\label{eq:length_err}
\end{align}
The maximum error ($4 \varepsilon$) can occur when one of the edges is shortened by $2\varepsilon$ and the other is lengthened by $2\varepsilon$. Fig.~\ref{fig:noise_bound}B exemplifies how edges may become longer even though the deformation represents the \textit{erosion} of material.

\subsection{$\tilde{C}_2$ : Mate angle constraint under noise}
\label{Sec:noisy_angle_C}

While it is clear that vertices of neighboring pieces may not meet if either sustains noise, and thus may no longer be expected to generate two supplementary angles in a strict way, one can still bound the deviation from that ideal behavior. To do so we first analyze the effect of noise on the degree of rotation of any single edge relative to its noiseless configuration and then leverage that result for the desired bound on the angles of mating edges under noise.

\begin{itemize}

\item[i.] \underline{Bound on the rotation of a single \enoisy edge}

Let $e = (\vec{u}_1, \vec{u}_2)$ be an edge (of some puzzle piece) with coordinates $\vec{u}_1= (x_1, y_1), \vec{u}_2=(x_2, y_2)$ and size ${\norm{\vec{u}_1-\vec{u}_2}=L}$, and assume (without loss of generality) that this edge is aligned with the origin and the $X$ axis of some reference coordinate frame and thus stretches
from $\vec{u}_1 = (0, 0) $ to $\vec{u}_2 = (L, 0)$. The orientation of this edge is of course $\measuredangle e=0^{o}$, as illustrated by the green edge in Fig. \ref{fig:noise_bound_C}.

\begin{figure}[t]
    \centering
  \fontsize{7pt}{8pt}\selectfont
  \def\svgwidth{0.95\columnwidth}
    \includegraphics[width=0.99\columnwidth]{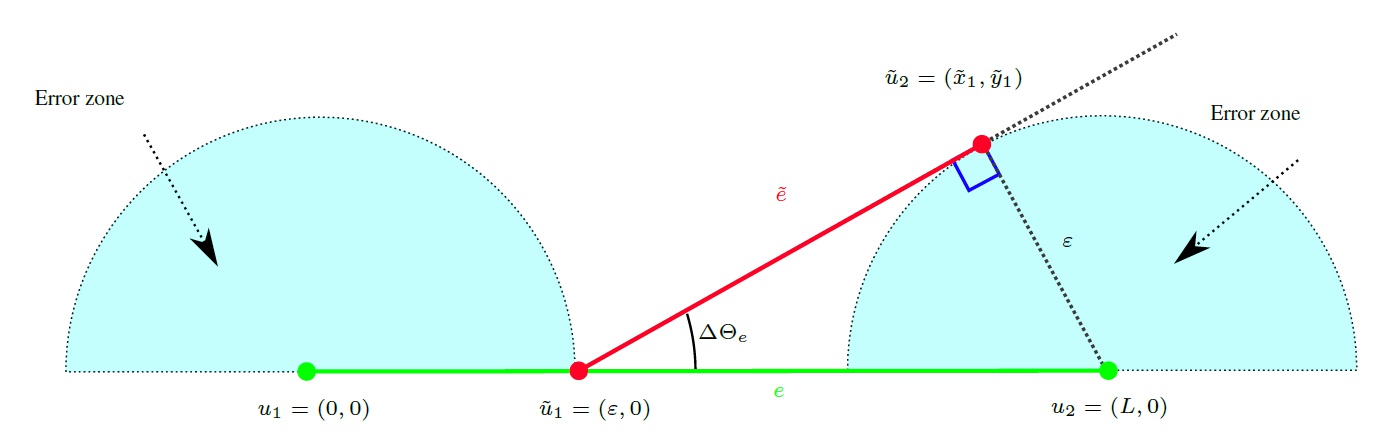}
  \CAPTION{
           If the ``clean'' edge $e$ (in green) stretches (w.l.o.g) from $u_1 = (0,0)$ to $u_2 = (L, 0)$, the vertices of the \enoisy edge must lie inside the corresponding error zones (in cyan). When considering the angle of the \enoisy edge $\tilde{e}$ (in red), the worst case occurs when one of the vertices (say, $u_1$) is only perturbed horizontally by $\varepsilon$, while the other (say, $u_2$) is perturbed to maximize the rotation, i.e, to a point $\tilde{u}_2 = (\tilde{x}, \tilde{y})$ that makes $\tilde{e}$ tangent to the error zone.  
    This bound is expressed in Eq.~\ref{eq:angle_err}.}
   \label{fig:noise_bound_C}
\end{figure}

Let us now denote  by
$\tilde{e} = 
(\vec{\tilde{u}}_1, \vec{\tilde{u}}_2) = 
((\tilde{x}_1, \tilde{y}_1),\allowbreak
(\tilde{x}_2, 
\tilde{y}_2))$
the same edge after applying the noise. Except for accidental cases, the orientation $\measuredangle \tilde{e}$ will be different than $\measuredangle e$, as was already exemplified in Fig.~\ref{fig:noise_bound}.
Let $\Delta\Theta_e(L, \varepsilon)$ be the bound on the difference between these two orientations over all possible \enoisy edges $\tilde{e}$, i.e., over all combinations of the noisy vertices $(\vec{\tilde{u}}_1, \vec{\tilde{u}}_2)$ that are possible under the noise model. In our case,
\begin{align*}
\Delta\Theta_e(L, \varepsilon)
&= 
\max_{\tilde{e}} \abs{\measuredangle \tilde{e}- \measuredangle e } = \max_{\tilde{e}} \abs{\measuredangle \tilde{e}} \;.
\end{align*}

To obtain the maximal (i.e. worst case) orientation change $\Delta \Theta_e$ while the vertices of $\tilde{e}$ remain in their respective error zones (cyan semi-disks in Fig.~\ref{fig:noise_bound_C}), it is needed to perturb one of the vertices only horizontally while the other is perturbed vertically as much as possible. This happens when $\tilde{e}$ becomes tangent to the error zone as shown in Fig.~\ref{fig:noise_bound_C} and thus the bound is:
\begin{align}
    \Delta\Theta_e(L, \varepsilon) =
    \begin{cases}
        \arcsin \left( 
        \frac{\varepsilon}{L - \varepsilon}
        \right) & L > 2\varepsilon \\
        \infty                                  & L \leq 2  \varepsilon
    \end{cases}\;\;.
\label{eq:angle_err}
\end{align}
Note that ``short'' edges ($L \leq 2  \varepsilon$) are special since the error zones intersect and thus the \enoisy edge might take arbitrary orientation or simply vanish altogether. In these cases, we set the bound to infinity ($\infty$)  to represent the fact that the angle constraint cannot contribute useful information and thus cannot be employed constructively. In practice, a finite value of $\frac{\pi}{2}$ (the bound of $\arcsin$) can serve our purpose equally well.

Eq.~\ref{eq:angle_err} requires the length of the original (``clean'') edge $L$, but in practice only $\tilde{L}$ can be measured. However, following the same arguments behind constraint $\tilde{C}_1$ (Sec.~\ref{Sec:noisy_length_C}), it holds that 
$L \geq \tilde{L} - 2\varepsilon$ and this lower bound can be used as a worst case. We therefore conclude that an \enoisy edge $\tilde{e}$ with length $\tilde{L}$ might be rotated relative to the original ``clean'' edge no more than
\begin{align}
    \Delta\Theta_e(L, \varepsilon) \leq
    \begin{cases}
     \arcsin
    \left(
    \frac
    {\varepsilon}
    {\tilde{L} - 3 \varepsilon}
    \right)
    &
    \tilde{L} > 4\varepsilon\\
    \infty & \tilde{L} \leq 4 \varepsilon
    \end{cases} \;.
\label{eq:tilde_noise}
\end{align}

\item[ii.] \underline{Bound on the angle difference of two corresponding mates}

Let $e$ and $e'$ be two ``clean'' mates  and denote the corresponding lengths of the edges \textit{before}, \textit{at}, and \textit{after} these mates as $L_{-1}, L_{0}, L_{1}$ and $L'_{-1}, L'_{0}, L'_{1}$, respectively, as illustrated in Fig.~\ref{fig:noise_bound2}A. 
Let $\alpha_1, \beta_1$  and $\alpha_2, \beta_2$ be the two pairs of supplementary angles these mates form with their adjacent edges at their vertices, also as illustrated in Fig. \ref{fig:noise_bound2}A. 
Recall that the mate angle constraint $C_2$ dictates that
\begin{align*}
    &\alpha_1 + \beta_1 = \alpha_2 + \beta_2 = \pi \;\;.
\end{align*}
\begin{figure}[b]
  \centering
  \includegraphics[width=0.7\columnwidth]{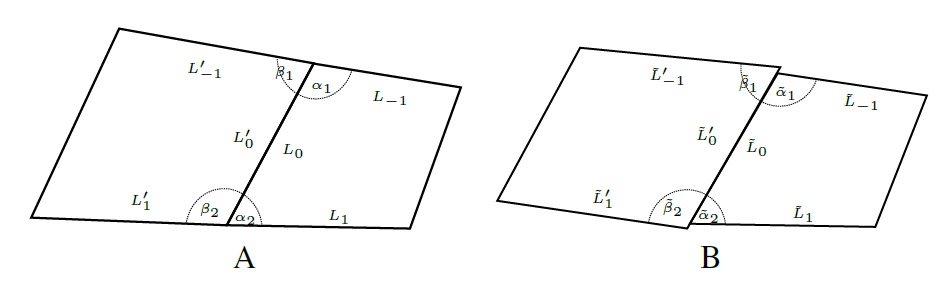}
    \CAPTION{
    The effect of noise on the mate angle constraint.
    \textbf{A: } Without noise, angles must comply to the original constraint $\alpha_1 + \beta_1 = \alpha_2 + \beta_2= \pi$. 
    \textbf{B: } After applying the noise the \enoisy angles are affected by the change in orientation in all edges that meet at both vertices of both mates, to result in the bound in the text.
    }
  \label{fig:noise_bound2}
\end{figure}
Let $\tilde{\alpha}_i, \tilde{\beta}_i$ $i\in\{1,2\}$ be the angles corresponding to $\alpha_i, \beta_i$ after applying the noise, as shown in Fig. \ref{fig:noise_bound2}B. It is clear that the bound on how different $\tilde{\alpha}_i, \tilde{\beta}_i$ from their ``clean'' versions $\alpha_i, \beta_i$ is determined by the maximal change of orientation of each of their constituent rays (i.e., edges), as expressed in Eqs.~\ref{eq:angle_err},\ref{eq:tilde_noise}. We thus get
{\scriptsize
\begin{align*}
    &|\alpha_1 - \tilde{\alpha}_1| \leq 
    \Delta \Theta_e (L_0,\varepsilon) + 
    \Delta \Theta_e (L_{-1},\varepsilon)\\
    &|\alpha_2 - \tilde{\alpha}_2| \leq 
    \Delta \Theta_e (L_0,\varepsilon) + 
    \Delta \Theta_e (L_{1},\varepsilon)\\
    &|\beta_1 - \tilde{\beta}_1| \leq 
    \Delta \Theta_e (L'_0,\varepsilon) + 
    \Delta \Theta_e (L'_{-1},\varepsilon)\\
    &|\beta_2 - \tilde{\beta}_2| \leq 
    \Delta \Theta_e (L'_0,\varepsilon) +
    \Delta \Theta_e (L'_{1},\varepsilon)
\end{align*}
}
Combining with the mate angle constraint we obtain
{\scriptsize
\begin{align*}
    |\pi - \tilde{\alpha}_1 - \tilde{\beta}_1|
    \leq &
    \Delta \Theta_e (L_0,\varepsilon) +
    \Delta \Theta_e (L_{-1},\varepsilon)
     +\\
    & 
    \Delta \Theta_e (L'_0,\varepsilon) +
    \Delta \Theta_e (L'_{-1},\varepsilon)
    \\
    |\pi - \tilde{\alpha}_2 - \tilde{\beta}_2|
    \leq &
    \Delta \Theta_e (L_0,\varepsilon) +
    \Delta \Theta_e (L_{1},\varepsilon)
     +\\
    & 
    \Delta \Theta_e (L'_0,\varepsilon) +
    \Delta \Theta_e (L'_{1},\varepsilon)    \;,
\end{align*}
}
and finally we apply the bound in Eq.~\ref{eq:tilde_noise} to reflect the fact that the true edge lengths are unknown. 
$\tilde{C}_2$, the final mate angle constraint under noise thus incorporates the following two inequalities
{\scriptsize
\begin{align*}
    \begin{split}
    |\pi - \tilde{\alpha}_1 - \tilde{\beta}_1|
    &\leq
    \Delta \Theta_e(\tilde{L}_0 - 2\varepsilon ,\varepsilon) + 
    \Delta \Theta_e(\tilde{L}_{-1} - 2\varepsilon,\varepsilon)\\
    &\quad+ 
    \Delta \Theta_e(\tilde{L}'_0 - 2\varepsilon,\varepsilon) + 
    \Delta \Theta_e(\tilde{L}'_{-1} - 2\varepsilon,\varepsilon)    
    \end{split}\\
    \begin{split}
    |\pi - \tilde{\alpha}_2 - \tilde{\beta}_2|
    &\leq
    \Delta \Theta_e(\tilde{L}_0 - 2\varepsilon,\varepsilon) +
    \Delta \Theta_e(\tilde{L}_{1} - 2\varepsilon,\varepsilon)\\
    &\quad+ 
    \Delta \Theta_e(\tilde{L}'_0 - 2\varepsilon,\varepsilon) + 
\Delta \Theta_e(\tilde{L}'_{1} - 2\varepsilon,\varepsilon) \;\;.
    \end{split} 
\end{align*}
}
\end{itemize}

To conclude this analysis, and similar to the mating constraints in the ``clean'' case, we may refer to the noisy mating constraints as predicates:
\begin{align*}
\forall i\in\{1,2\} \quad \tilde{C}_i\left(e_i^j, e_k^l\right) = \text{true} \; \Leftrightarrow \; e_i^j\text{~and~}e_k^l \text{~~satisfy~} \tilde{C}_i \;.
\end{align*}

\subsection{Noise-induced erased pieces}
\label{Sec:erased_pieces}

\begin{figure}[h]
    \centering
    \begin{tabular}{cccc}
        \includegraphics[width=0.15\columnwidth]{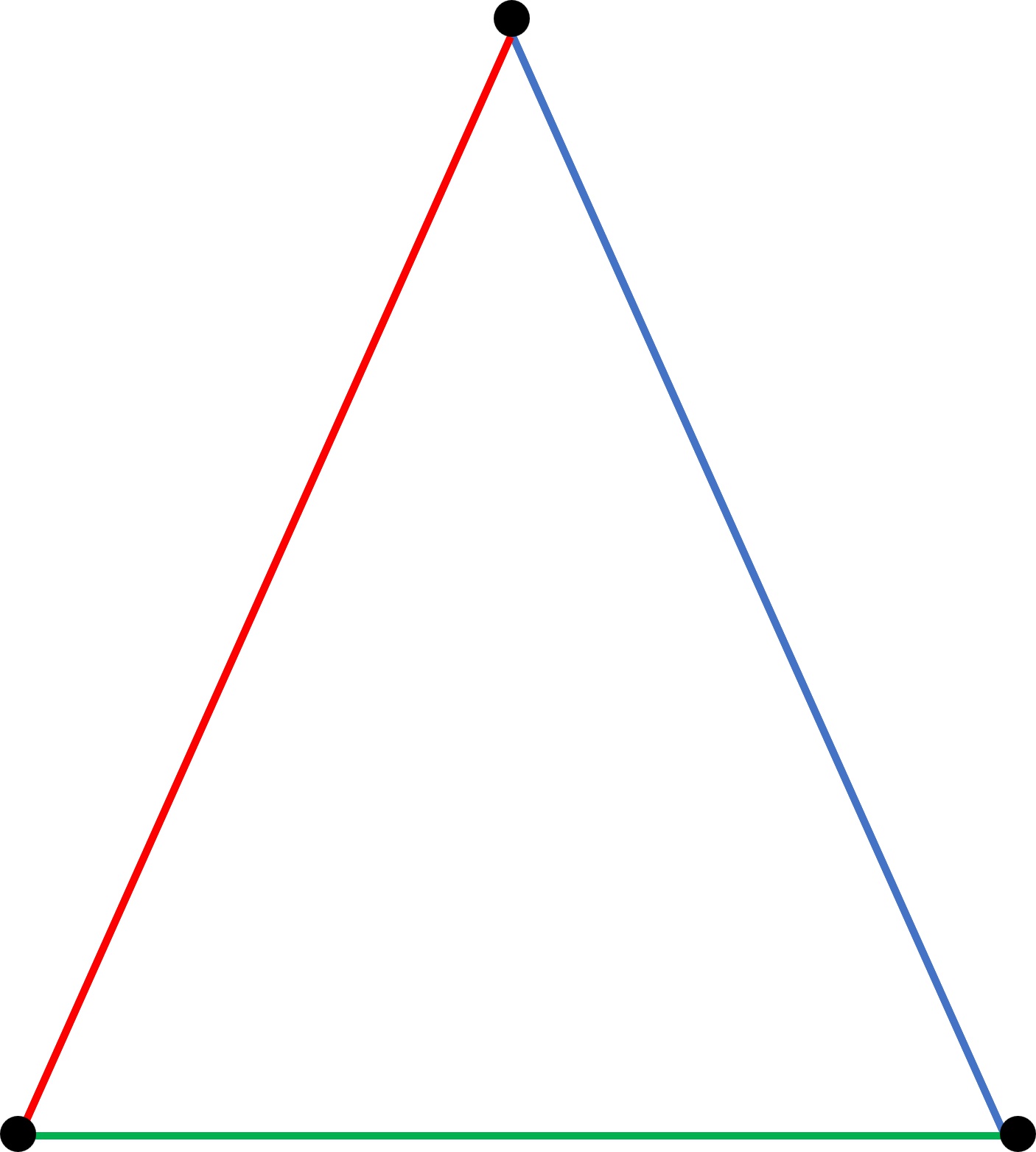}
        &
        \includegraphics[width=0.15\columnwidth]{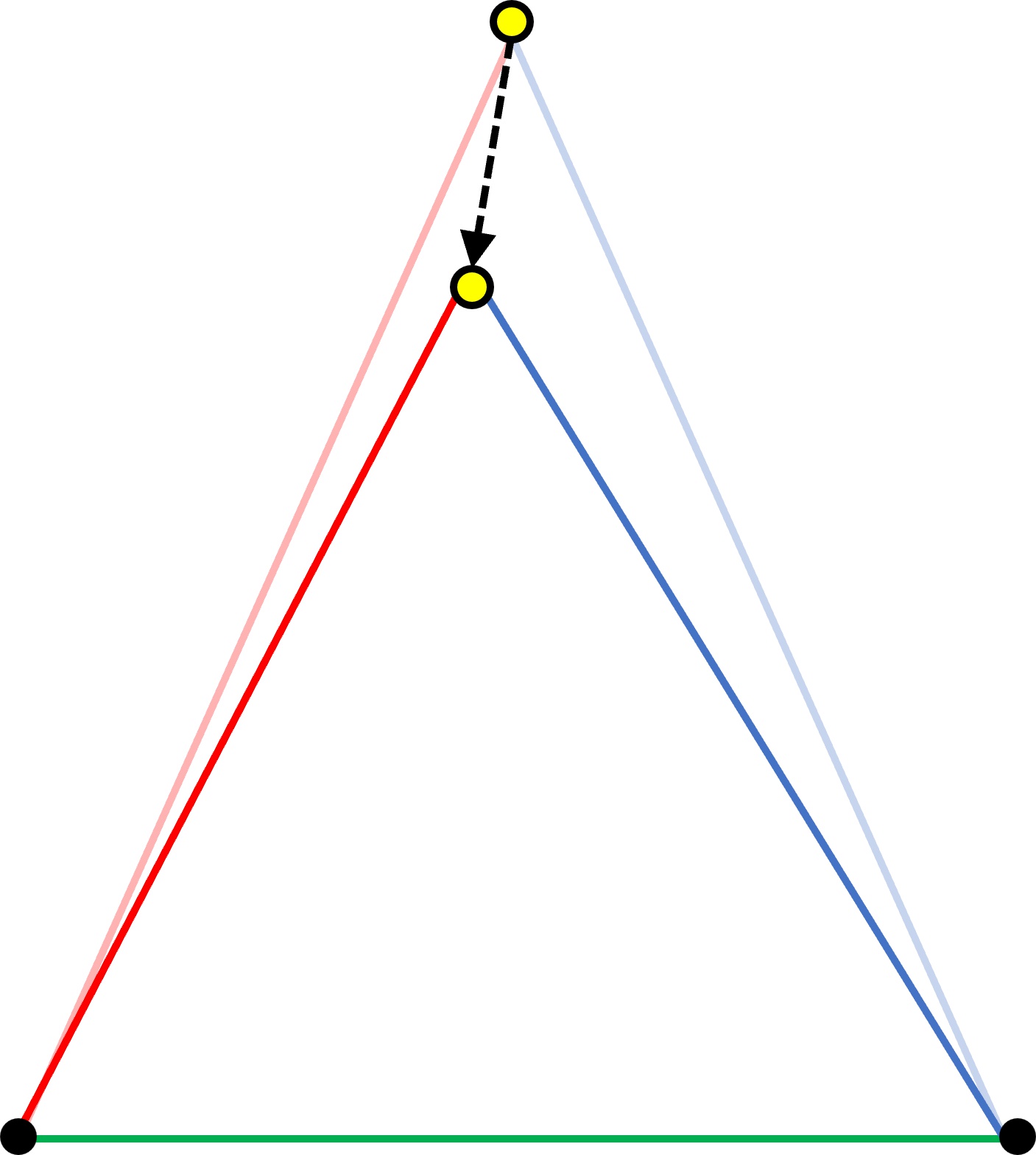}
        &
        \includegraphics[width=0.15\columnwidth]{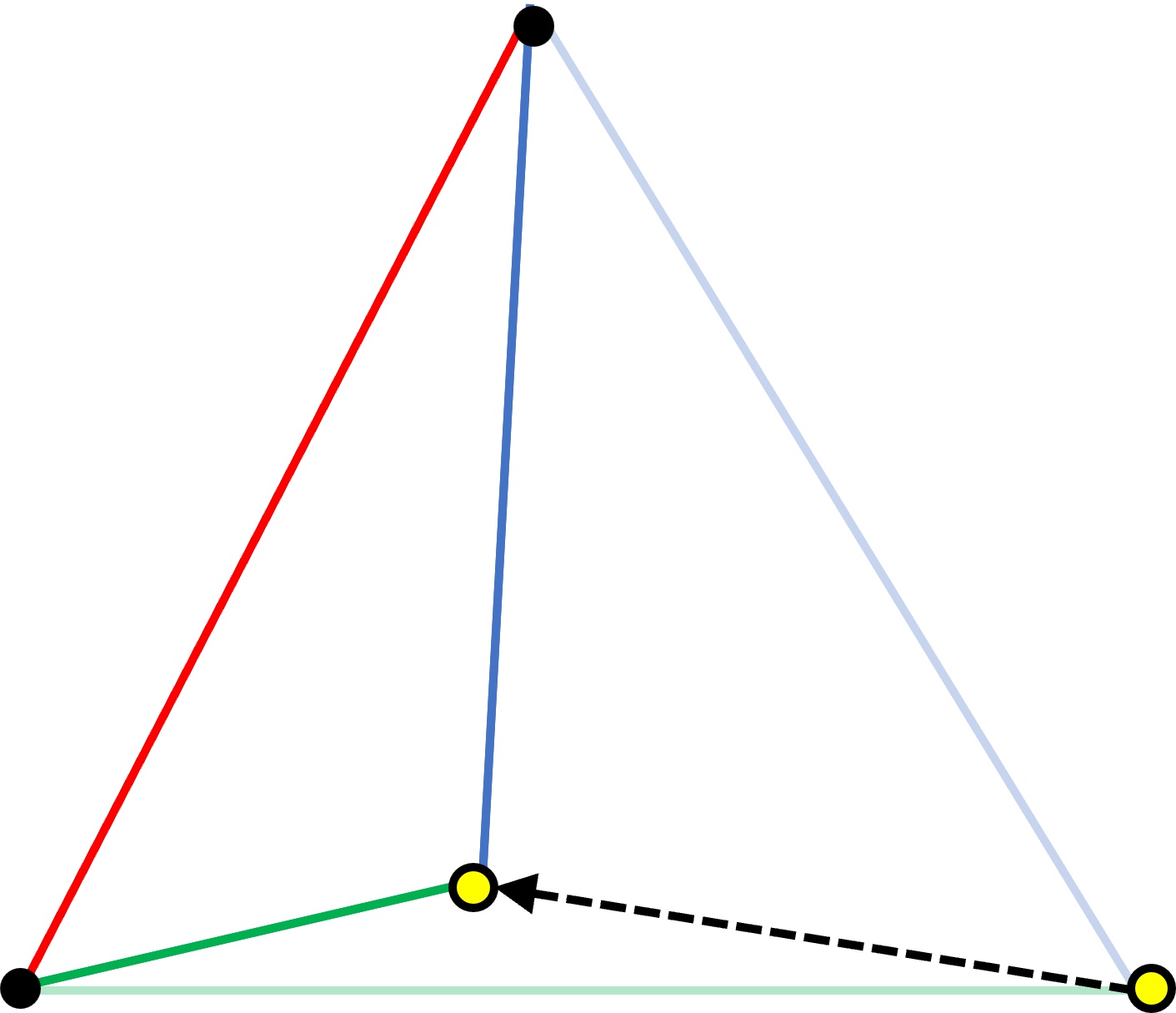}
        &
        \includegraphics[width=0.15\columnwidth]{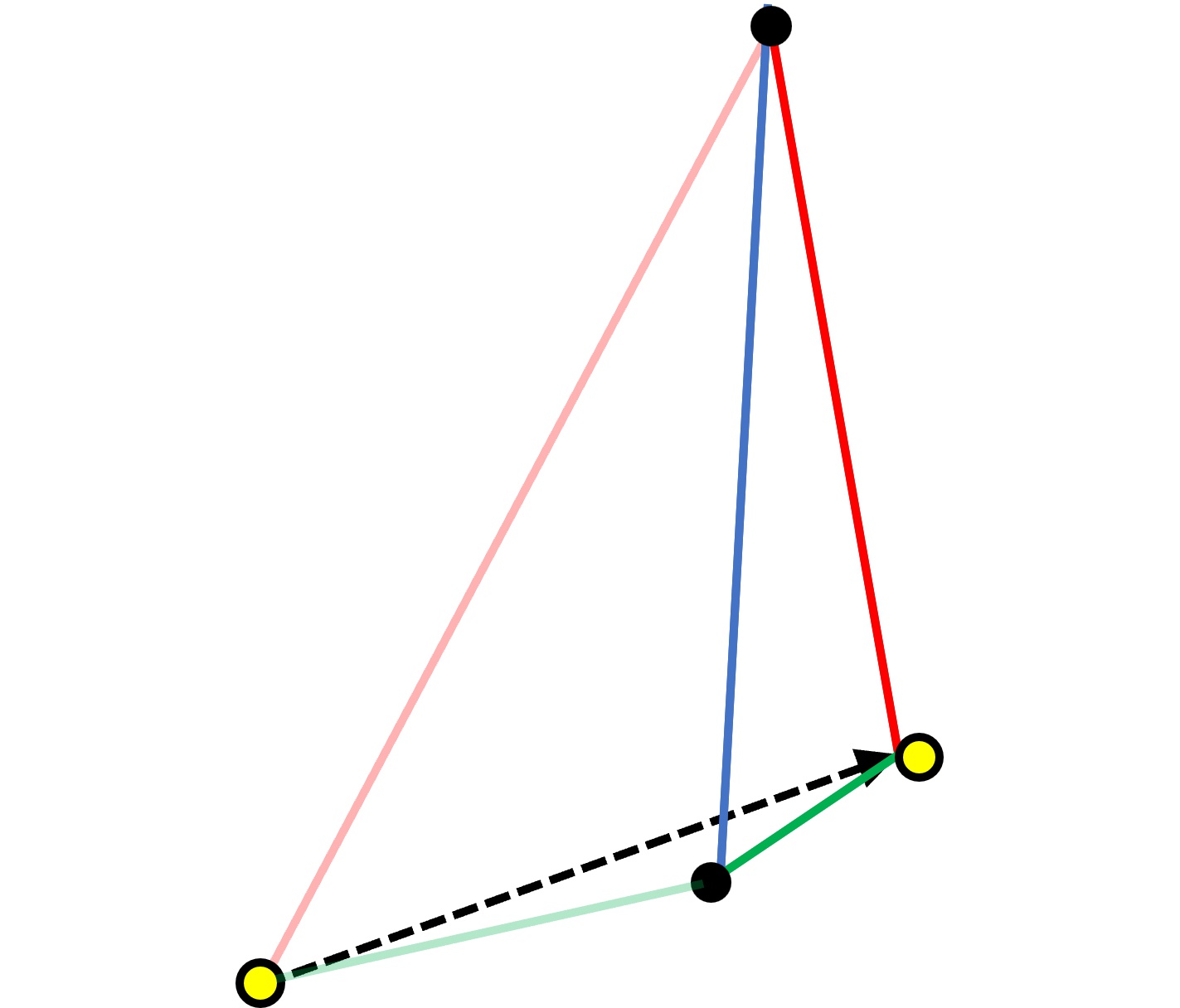} \\
        \\
        A & B & C & D
        \end{tabular}
        \CAPTION{
           Example of a triangular piece being \textit{disqualified} thus \textit{erased} as a result of being exposed to relatively large noise, applied to each of the piece's vertices in clockwise order. \textbf{A:} The original 'clean' piece. 
           \textbf{B:} Noise is applied to the first vertex, pushing it down. \textbf{C:} Noise is applied to the second vertex, pushing it left. 
           \textbf{D:} Noise is applied to the last vertex, pushing it right beyond the current piece border, thus eliminating it. }
  \label{fig:erased_piece}
\end{figure}

An inevitable consequence of applying erosion to a puzzle with pieces of various sizes is the potential risk of piece disappearance. Not unlike in the physical world, relatively smaller pieces are at greater risk of being completely eroded and thus practically disappear. In practice, in our noise model, this happens if the random inward perturbation applied to a vertex pushes it beyond some other boundary of the piece, as illustrated in Fig.~\ref{fig:erased_piece}. The result of course is a noisy puzzle with \textit{missing} pieces.

In our present work, missing pieces are not yet handled and solving puzzles with missing pieces is left for future work, as it is likely to require a completely different approach. At present, the possibility of missing pieces due to geometric noise can be reduced or even prevented by using a \textit{softer} noise model based on a smaller adaptive bound (for example, piece-adapted noise bound defined by the piece's shortest edge).


\subsection{Pictorial noisy puzzles}
\label{Sec:noisy_pictorial}

\begin{figure}[!h]
    \centering
    \begin{tabular}{cc}
        \includegraphics[height=0.25\columnwidth]{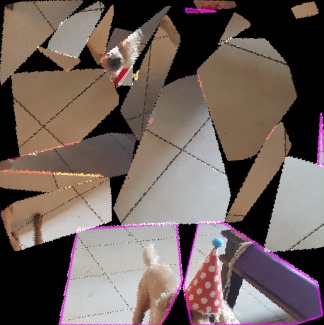}
        &
        \includegraphics[height=0.25\columnwidth]{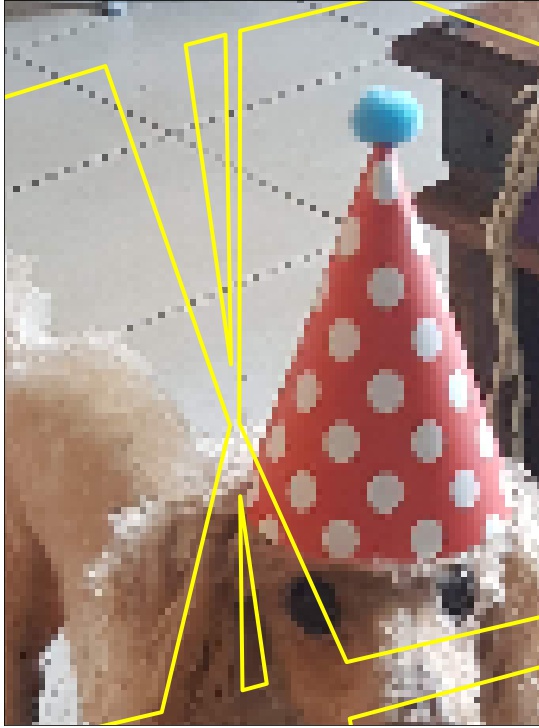}\\
        A & B
        \end{tabular}
\CAPTION{
            Example of a noisy pictorial crossing cut puzzle.
            \textbf{A:}~A noisy pictorial crossing cuts puzzle is an unordered set of pictorial \textit{noisy} pieces and recall that the noise is geometrical, not pictorial. The four pieces highlighted in purple are those shown in the next panel. 
            \textbf{B:}~A closeup on four pictorial neighbors in their original position after they are cut from the original pictorial polygon and then contaminated with geometrical noise. Note how the noise sets the available pictorial content apart, thus complicating the decision about the affinity of the corresponding edges.
}
\label{fig:pictorial_puzzle}
\end{figure}

Just like apictorial crossing cuts puzzles, their pictorial counterpart can also be contaminated by geometric noise. A typical pictorial noisy crossing cut puzzle is depicted in Fig.~\ref{fig:pictorial_puzzle}A and since it is impossible to observe the noise when the pieces are shuffled,  Fig.~\ref{fig:pictorial_puzzle}B puts several pieces in place to demonstrate the consequences.  It is easy to observe that even if the pictorial content is immune to image noise, the geometric noise distances the pictorial content that {\em is} available in different puzzle pieces and thus complicates the way it can be used to determine plausible mates. We discuss a scheme that addresses the latter challenge in Sec.~\ref{chap:puzzle_reconstruction}.

\section{Data synthesis}
\label{sec:data_synthesis}

Since there is no previous work on crossing cuts puzzles, no data or benchmark results exist either. Part of our contribution here is a mechanism for data synthesis, as well as the first public dataset of crossing cuts puzzles.
Such synthesis tools and datasets facilitate both the exploration of valuable properties of such puzzles and the experimental evaluation of reconstruction algorithms.

The synthesis process is based on a computational procedure that receives as input a description of the global polygonal shape $S$ (which could be specified by the user or selected at random; see below) and the crossing cuts $Cuts = \{c_1, \dots c_a\}$ that dissect it. It returns both the \textit{puzzle}, which can be given as input to reconstruction algorithms and the \textit{ground truth solution} that can be used to evaluate the performance of puzzle solvers.
As discussed in Sec.\ref{sec:the_puzzle}, the puzzle is a bag of polygonal pieces $P=\{p_1, \dots p_n\}$, each represented properly by its vertices in some coordinate frame of reference. The ground truth solution constitutes a representation of the mating graph (and in particular, the set $M$ of its matings), as well as the Euclidean transformations $((R_1, t_1), \dots (R_n, t_n))$ that place the pieces correctly in the reconstructed puzzle (or equivalently, the coordinates of the vertices of all pieces).

The process of synthesizing crossing cuts puzzles thus constitutes several aspects, all of which are described next for the sake of reproducibility. We note that pictorial puzzles are produced similarly to apictorial ones while the global polygonal shape is covered ahead of time by some pictorial content (e.g., from a user-provided image).

\subsection{A graph representation for planar divisions}
\label{sec:get_planar}

\begin{figure}[h!]
  \centering
    \includegraphics[width=0.4\columnwidth]{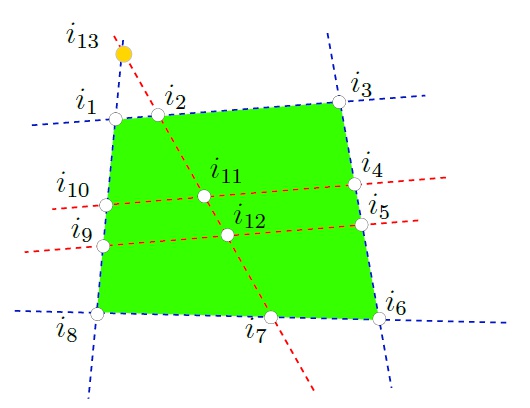}
    \CAPTION{
    The stages of representing a crossing cuts puzzle as a planar graph for the purpose of synthesis.
    In this selected example, the green quadrilateral is the global puzzle shape $S$ whose boundary is defined by four lines (dashed blue).  Three additional lines are defined as the crossing cuts (dashed red) and the intersection points of all these 7 lines \textit{that also lie inside or on the border of} $S$ are considered the nodes of the puzzle's graph (and therefore $\{i_1 \dots i_{12}\}$ \textit{are} nodes in the graph, but $i_{13}$ is \textit{not}).
    The edges of the graph are defined by pairs of nodes that lie closest on the same line, i.e., by a pair of nodes that lie on the same line such that there is no other node in between them. Hence $\{i_3, i_4\}, \{i_3, i_2\}, \{i_4, i_5\}$ \textit{are} edges but $\{i_3, i_5\}$ is \textit{not}.
    }
  \label{fig:closest_intersubsections}
\end{figure}

Let $S \subseteq R^2$ be a polygonal puzzle shape. The first stage of data synthesis is to construct a puzzle planar graph $\mathcal{G}_{puzzle} = (\mathcal{V},\mathcal{E})$ that represents both the boundary of $S$ and the cuts that go through it. Note that $\mathcal{G}_{puzzle}$ is different from the mating graph and is maintained for the synthesis process only.
Toward that end, we first combine both the boundary lines of $S$ (dashed blue lines in Fig.\ref{fig:closest_intersubsections}) and the crossing cuts themselves (dashed red lines in Fig.~\ref{fig:closest_intersubsections}) into one set of lines:
\begin{align*}
    \mathcal{C} &= Cuts \cup \{\text{edge lines of }S\}\;.
\end{align*}
The particular representation of lines is secondary, but in our case we represent each of them as a triplet $(a_1, a_2, a_3)$, where $a_i$ are the coefficients of the line equation $a_1 x + a_2 y +  a_3 = 0$ in some global coordinate system.

The nodes of $\mathcal{G}_{puzzle}$ are the intersection points of any two lines in $\mathcal{C}$ that rest inside or on the border of $S$ (see Fig.~\ref{fig:closest_intersubsections}).
Formally, this set of nodes is defined as follows:
\begin{align*}
    \mathcal{V}
    &= \left\{
    i \in S
    \Bigm|
    \exists (c_1, c_2) \in \mathcal{C} \times \mathcal{C},
     \quad
    (c_1 \neq c_2)
    \wedge
    \left(
    i = c_1 \cap c_2
    \right)
    \right\}\;.
\end{align*}
The set $\mathcal{E}$ of the edges of $\mathcal{G}_{puzzle}$ link pairs of nodes that rest on the same line with no other nodes between them, or formally:
\begin{align*}
    \mathcal{E}
    &=
    \left\{
    \{i_1, i_2\}
    \Bigm|
    \exists c \in \mathcal{C},
    \quad
    \left(
    i_1, i_2\in c \cap \mathcal{V}
    \right)
    \wedge
    \left(
    [i_1, i_2]\cap \mathcal{V} = \emptyset
    \right)
    \right\}
\end{align*}
where $[i_1, i_2]$ is the line segment (as a set of points) between node points $i_1$ and $i_2$.

\subsection{Generation of pieces and ground truth matings}
\label{sec:obtain_regions}

The extraction of the pieces from graphs that represent planar divisions has been addressed in the graph algorithms community and here we employ the optimal algorithm due to Jiang and Bunke~\cite{jiang1993optimal}.
This computational process receives the planar graph $\mathcal{G}_{puzzle}$ from Sec.\ref{sec:get_planar} and outputs all of the minimal polygonal regions, each represented as the ordered list of nodes that delineate it. One such region in Fig.~\ref{fig:closest_intersubsections} is $(i_1, i_2, i_{11}, i_{10})$.

The main construct in the algorithm is the notion of \textit{wedge}~\cite{jiang1993optimal}, defined as a pair of different edges that meet at a node (e.g., $(\{i_1, i_2\}, \{i_2, i_3\})$ so that no other edge is encountered when rotating the first edge towards the second
(e.g.  $(i_2, i_{11}, i_4)$ in Fig.~\ref{fig:closest_intersubsections} is a wedge, but $(i_{10}, i_{11}, i_4)$ is not a wedge).
A closed chain of overlapping wedges (e.g
$(
(i_1, i_2, i_{11}),
(i_2, i_{11}, i_{10}),
(i_{11}, i_{10}, i_{1})
)$ in Fig.~\ref{fig:closest_intersubsections})
defines a minimal region, and thus a puzzle piece. The sorting scheme that locates the wedge chains was shown to have $O(|\mathcal{E}| \log (|\mathcal{E}|))$ run-time complexity and $O(|\mathcal{E}|)$ memory complexity. Please refer to the original paper for more details.

The application of Jiang and Bunke~\cite{jiang1993optimal} results is a set of puzzle pieces that are positioned in their original puzzle location, and thus, if the generated puzzle is pictorial, this is the point where the geometric representation of the pieces serves to crop the original pictorial content that belongs to each piece.
Either case, the segmentation of the original polygon into pieces in their ``correct'' position is now suitable for the computation and representation of the \textit{desired} solution for the puzzle, at the very least for the evaluation of solvers output against the ground truth.  Indeed, at this point of the synthesis process, any pair of neighboring pieces is positioned such that their mating edges strictly overlap. Hence the extraction of the ground truth mating graph can be done, for example, by finding all \textit{overlapping} edges $e_i^j$ and $e_k^l$ that belong to \textit{different} pieces.
Formally, if $E_m$ represents all edges of piece $p_m$ (cf. Sec.~\ref{sec:the_puzzle}) and thus $E=\left( \bigcup_{m=1}^n E_m \right)$ is the set of all edges of all pieces, the ground truth matings are
\begin{align*}
M = \left\{
(e_i^j, e_k^l)
\in E \times E
 \Bigm|
 (p_i \neq p_k)
 \wedge
 e_i^k = e_k^l
\right\}\;.
\end{align*}

\subsection{Piece randomization and geometric noise}
\label{sec:piece_randomization_and_noise}

The final puzzle representation that is submitted to solvers should include no information about the ground truth position of the pieces.
But with all pieces generated and the ground truth secured, puzzle pieces can now be shuffled and their Euclidean transformations randomized.
To do so we first center each piece about its center of mass, i.e., each vertex is translated by the average of all vertices of the same piece. Then we apply a rotation transformation by some random angle selected uniformly in $[0,2\pi]$. Needless to say, for pictorial puzzles the pictorial content is transformed correspondingly.
If we denote the random rotation for piece $p_i$ by ${RR}_i$ and the translation to the center of mass by $\overrightarrow{tc}_i$,
The ground truth positioning of the pieces is of course the inverse transformation $\{[RR_i]^{-1}, -(\overrightarrow{tc\,}_i)\}$.

If the desired puzzle should be ``clean'', the process ends here and the list of randomly ordered and transformed pieces, each in its own coordinate system, serves as the final puzzle representation. However, if a noisy puzzle is required, each vertex of each piece is first translated by a random noise vector that obeys the constraints from Sec.~\ref{sec:noise}. If the puzzle is pictorial, the application of the noise also crops the corresponding parts of the pictorial content. Only then the list of pieces is wrapped as the puzzle representation.

\subsection{Datasets}
\label{sec:datasets}

We created several datasets using the procedure just described, where each serves a different purpose. The first dataset is tailored for the empirical exploration of statistical properties of crossing cuts puzzles while the others are designed to facilitate experimental evaluation (and if needed, of training) of crossing cuts solvers. The images for the pictorial content of all puzzles in all pictorial DBs were obtained from \texttt{\url{https://unsplash.com/}} 
and 
\texttt{\url{https://www.pexels.com/public-domain-images/}}, or taken by the authors with a digital camera.

\textbf{DB1 - circular puzzles for statistical properties:} Sec.~\ref{chap:statistic} presents a theoretical analysis of crossing cuts puzzles and their properties. To simplify and facilitate analytical analysis, it is performed on puzzles with circular global shape while the corresponding empirical properties were measured on synthesized puzzles whose shape was a unit triacontadigon (i.e., an approximation of a unit circle as a polygon of 32 sides). The random cuts in this case were selected by sampling two angles $\phi_1, \phi_2$ and then passing a line though the corresponding points on the circumference of the circle, namely $(\cos \phi_1, \sin \phi_1),\allowbreak (\cos \phi_2, \sin \phi_2)$.

Following this procedure we generated a collection of $300$ noiseless puzzles, $30$ puzzles for $10$ different numbers of crossing cuts $a\in \{10,20,\ldots,100\}$. The number of puzzle pieces that resulted from this procedure varied from $36$ to $2183$. For each ``clean'' puzzle we then generated several noisy versions, with noise level varying in $\xi \in [0\%, 0.1\%, 0.25\%, 0.5\%, 1\%, 2\%]$. Recall that $\xi$ is the noise bound relative to puzzle diameter, which in DB1's case is always 2. In absolute terms, the noise in this dataset thus varied up to $\varepsilon\leq 0.04$, but perhaps more informatively, when considered against the average edge length, the noise could approach $64\%$ (i.e., $\bar{\xi} \leq 64\%$, see Sec.~\ref{sec:expected_edge_length}).   
    
\begin{figure}[h]
    \centering
    \includegraphics[width=0.8\columnwidth]{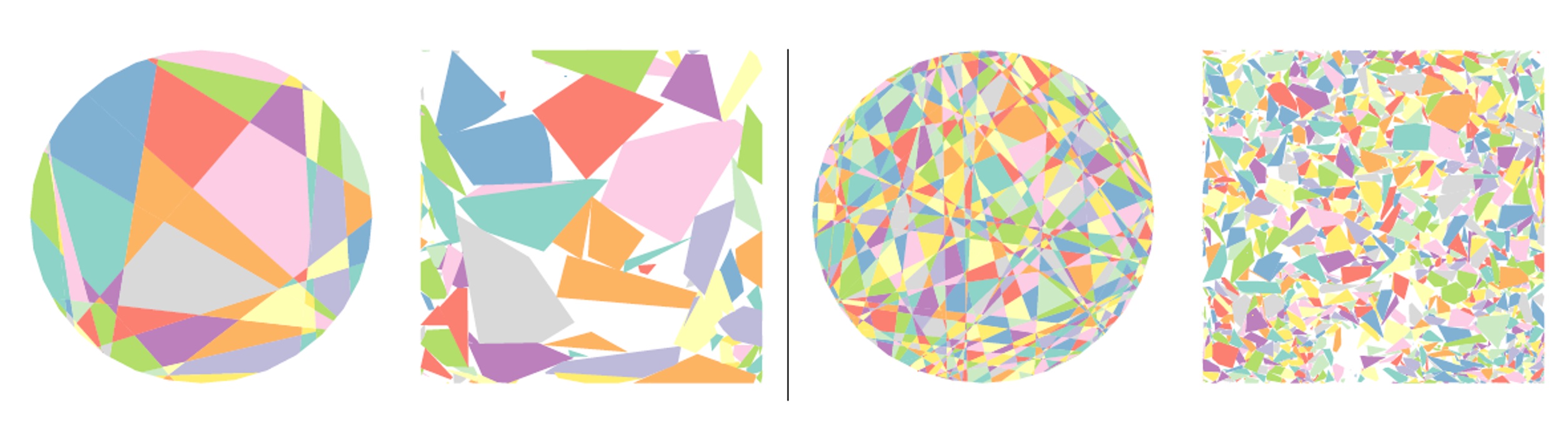}
    \CAPTION{
        Two selected synthesized circular puzzles for the analysis of puzzle properties. Shown are the puzzle as a bag of pieces and the corresponding ground truth solution. The numbers of cuts used for these examples are 20 and 100 while the corresponding number of pieces are 63 and 1806, respectively.}
    \label{fig:DB1}
    \end{figure}

With the noisy versions taken into account, the number of puzzles in DB1 thus totaled $1800$.
    For their intended use (i.e., analysis of properties), all puzzles in this dataset need not be pictorial and thus this is an apictorial dataset. Selected examples are shown in Fig.~\ref{fig:DB1}.

\textbf{DB2 - general {\em apictorial} puzzles for solvers evaluation:} Unlike the specifically crafted puzzle shape used for the analysis of puzzle properties, the evaluation of puzzle solvers requires randomly shaped (yet convex) puzzles. To achieve this goal we first sampled a random number (between 4 and 50) of randomly positioned points in some predetermined workspace $[0,W]\times[0,H]$ and then computed their convex hull to generate a random global convex polygonal shape (which in our case ended up having from $3$ to $14$ sides).
$W$ and $H$ are given as parameters to the synthesizer but they bear very little significance. In our case, we fixed them both at $W=H=100$.

    The random cuts $Cuts=\{c_1, \dots c_a\}$ were also selected as uniformly distributed random lines in the same workspace, but to ensure they indeed penetrate the random polygon we first selected two random points \textit{inside} the polygon and defined the cut as the line that goes between these points.

\begin{figure}[t!]
    \centering
    \includegraphics[width=0.7\columnwidth]{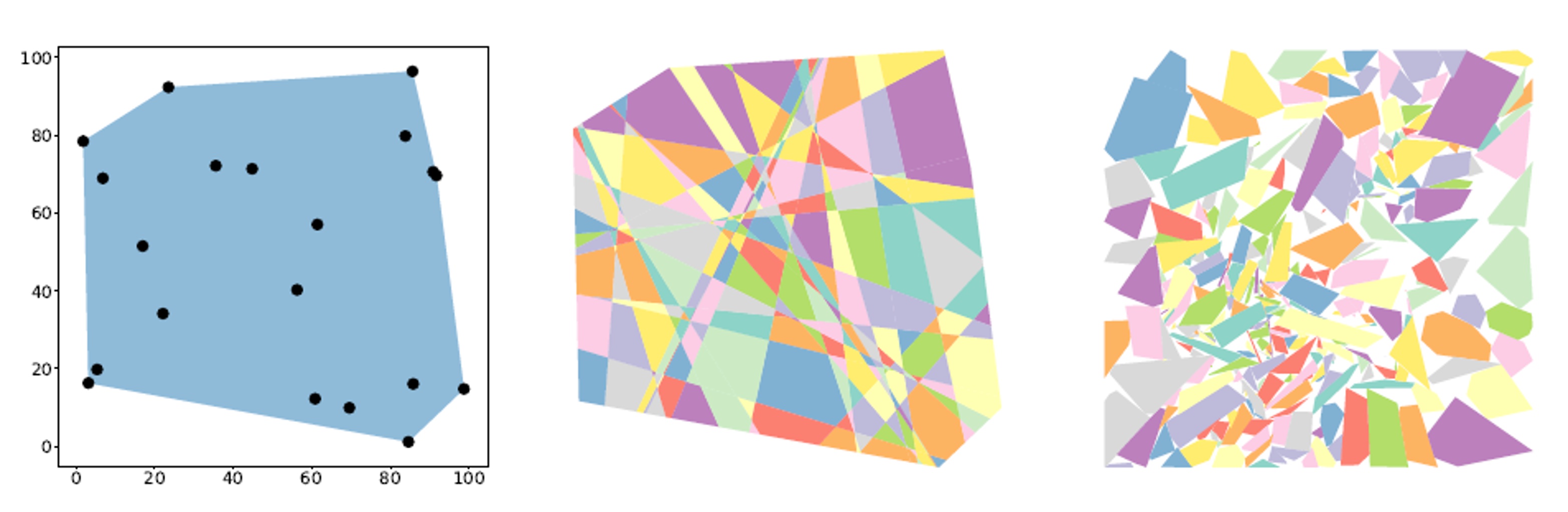}
    \CAPTION{
        One selected synthesized polygonal puzzle (30 cuts, 200 pieces) from DB2. The left column shows the process of generating the random puzzle shape in the workspace using the convex hull of random sample points. The middle column illustrates the ground truth solution and the right column shows the puzzles as a bag of pieces as provided to potential solvers. 
    } 
    \label{fig:DB2}
    \end{figure}

    While this procedure can be activated on demand and with arbitrary parameter values, we used it to generate a collection of $100$ random puzzles, whose number of cuts varies from $5$ to $50$ (10 instances from each case) and their number of pieces extends from $11$ (in the easier puzzles) to $936$ (in the more challenging ones).
    For each ``clean'' puzzle we also generated several noisy versions, with noise levels varying in $\xi \in [0\%, 0.1\%, 0.25\%, 0.5\%, 1\%, 2\%]$. With the noisy versions taken into account, the number of puzzles in DB2 thus totals $600$.
    Fig.~\ref{fig:DB2} shows one example from DB2 and aspects of its generation process.

\textbf{DB3 - Perturb grid {\em pictorial} puzzles for solver evaluation:} The procedure just described was used also for the generation of a pictorial dataset for evaluation (and if needed, also for training) of pictorial crossing cuts puzzle solvers, where the pictorial content that covers the puzzle (and consequently, it pieces) was provided as an image and handled as described earlier in Sec.~\ref{sec:data_synthesis}. However, to facilitate a better examination of the contribution of the pictorial content, in this first pictorial dataset, we reduced the role of the geometry by designating crossing cuts that generate edges of relatively similar lengths (both within and between pieces). This was done by defining the cuts to form a perturbed grid over the global polygonal shape, resulting in a narrower histogram of edge lengths and hence many more mating candidates when only geometry is considered. Without pictorial content, such puzzles will consider many more mating candidates, require a solver with significantly more computational resources, and (if the latter are bounded) may completely prohibit a solution unless pictorial considerations are incorporated too.
At the same time, with crossing cuts that are roughly parallel, we are also guaranteed that the bounded geometrical noise does not erode pieces completely, a situation that generates puzzles with missing pieces that are outside the scope of our present solver. For all these reasons we tested our pictorial puzzle solver on DB3, but already generated DB4 below for future generalizations.

\begin{figure}[h]
    \centering
    \begin{tabular}{cc}
     \raisebox{10pt}{\cellcolor[gray]{0} \includegraphics[width=0.2\columnwidth]{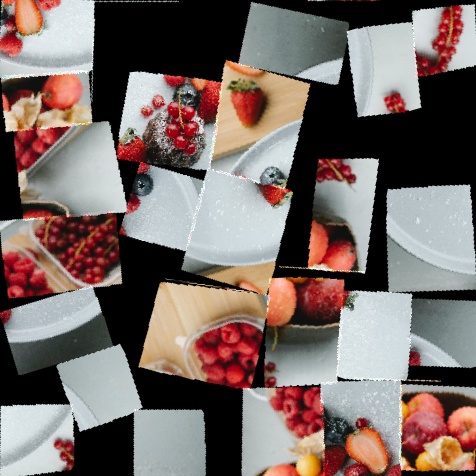}} & 
     \cellcolor[gray]{0} \includegraphics[width=0.2\columnwidth]{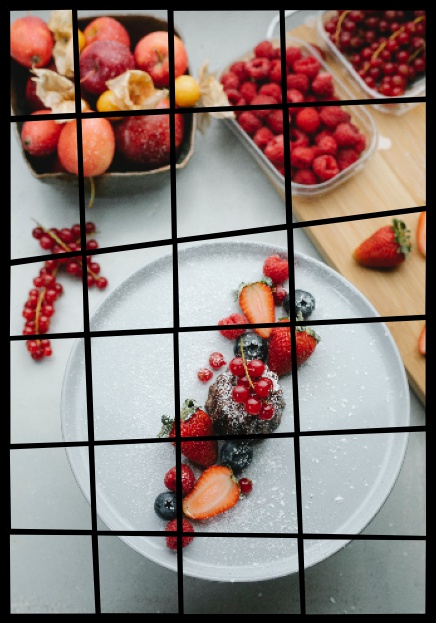} 
    \end{tabular}
    \CAPTION{
        An example of a perturbed grid pictorial puzzle (with 8 cuts and 24 pieces), both as a bag of pieces and the corresponding ground truth solution.
    }
    \label{fig:DB3}
    \end{figure}

Following this scheme, we generate a collection of $600$ random perturbed grid pictorial puzzles, whose number of cuts vary from $10$ to $100$ (with $10$ instances from each case) and number of pieces that extends from $35$ to $2601$. As before, noise level varied in $\xi \in [0\%, 0.1\%, 0.25\%, 0.5\%, 1\%, 2\%]$. Fig.~\ref{fig:DB3} shows a selected example.

In addition to the 3 datasets above, we created two additional datasets for future work by the community. These DBs are described next.

\begin{figure}[h]
    \centering
    \begin{tabular}{ccc}
     \raisebox{7pt}{\cellcolor[gray]{0} \includegraphics[width=0.2\columnwidth]{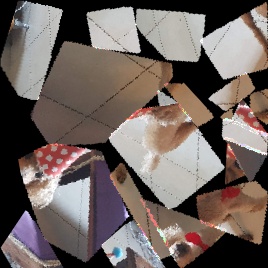}} & 
     \cellcolor[gray]{0} \includegraphics[width=0.2\columnwidth]{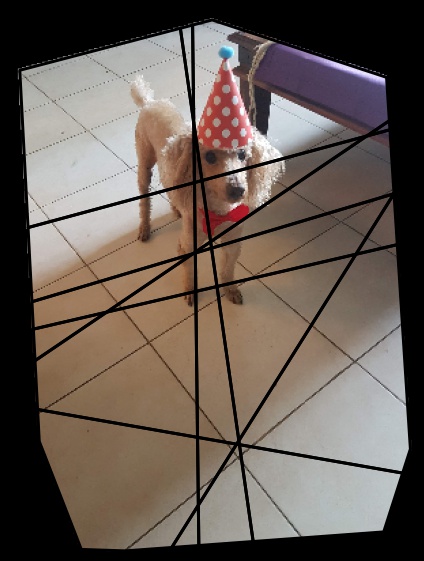}
    \end{tabular}
    \CAPTION{
        One selected example of a general pictorial puzzle (with 7 cuts and 18 pieces). Note that some small pieces appear only in the ground truth row as the noise removed them completely from the bag of pieces in the puzzle to solve.
        }
    \label{fig:DB4}
    \end{figure}

\textbf{DB4 - General {\em pictorial} dataset:} DB4 is another pictorial dataset for future research where the cuts are completely arbitrary and no special care is taken to downplay the role of geometric constraints. The importance of this dataset for future work is twofold. First, since the geometrical information becomes more significant and informative again (compared to DB3, for example), it will take more ``aggressive'' methods to exploit the pictorial content \textit{effectively}. Second, in arbitrary crossing cuts puzzles some pieces may turn small enough to completely disappear after the application of the geometrical noise (as indeed happens in $81.5\%$ of puzzles in this dataset). Thus, this DB4 also facilitates future research on crossing cuts puzzles with {\em missing pieces}.
Toward that end we generated a collection of $600$ random polygonal pictorial puzzles, whose number of cuts vary from $10$ to $100$ ($10$ instances from each case), number of pieces extends from $35$ to $3907$, and noise level in the range $\xi \in [0\%, 0.1\%, 0.25\%, 0.5\%, 1\%, 2\%]$.  A selected example of such a puzzle is shown in Fig.~\ref{fig:DB4}.

\textbf{DB5 - Square piece {\em pictorial} dataset:} As mentioned in Sec.~\ref{sec:the_puzzle}, from a geometrical point of view, strictly square piece puzzles are a very special case of crossing cuts puzzles where geometry plays no role and the pictorial content is the sole source of information for reconstruction. For ``backward compatibility'', and for their more general scheme of representation, using crossing cuts puzzles to represent square piece puzzles may be useful, especially if geometrical noise is to be allowed.
We therefore generated such a collection of $3600$ random square piece pictorial puzzles, whose number of cuts vary from $20$ to $200$ ($10$ instances from each case), number of pieces extends from $100$ to $10,000$, and noise level in the range $\xi \in [0\%, 0.1\%, 0.25\%, 0.5\%, 1\%, 2\%]$. Naturally, the different number of cuts generated pieces of different sizes, where in our cases extended from $5 \times 5$ to $30 \times 30$ pixels, yet another generalization of the prior art in square pieces puzzles that tended to focus on one size of $28 \times 28$ pixel only (though as mentioned in Sec.~\ref{sec:related_work}, one exception does exist~\cite{son2016solving}).


We note that all 5 datasets are open and available for the community at the public-domain portal \texttt{icvl.cs.bgu.ac.il{\textbackslash}polygonal-puzzle-solving}. This portal also will host additional datasets of varying characteristics as they become available.






\section{Puzzle properties}
\label{chap:statistic}

One of the advantages of the generation model that defines crossing cuts puzzles is the better
ability to analyze their properties. Since the model is stochastic, their properties are typically probabilistic, but nevertheless can provide insights on both the problem itself and about potential solutions (or limitations thereof). Here we explore such properties both analytically and, when needed, empirically. 
In this section, we assume that the global puzzle shape is a unit circle (or a polygonal approximation thereof), whose symmetry simplifies some of the analytical analyses. Most results, however, are indicative of all crossing cuts puzzles (up to a factor of half of their diameter). Empirical properties are evaluated on the DB1, the circular puzzles dataset that was described in Sec.~\ref{sec:data_synthesis}.

\subsection{Expected cut length}
\label{sec:expected_cut_length}

The first measure of interest is the length of a random cut $c_i$ through the global puzzle shape. When the latter is a unit circle, $c_i$ is determined by two points sampled uniformly on the circumference of the circle. In other words, the cut is determined by the chord between points
$\vec{p_1} = (\cos \phi_1, \sin \phi_1)$ and $\vec{p_2} = (\cos \phi_2, \sin \phi_2)$,
where the two angles are uniformly distributed random variables
$\phi_1, \phi_2 \sim \text{U}(0, 2 \pi)$. The length of cut $c_i$ is therefore another random variable defined by the function $l_{i} = \lVert \vec{p_2} - \vec{p_1} \rVert$, and one may seek its expected value.

\begin{figure}[h!]
    \centering
      \includegraphics[width=0.7\columnwidth]{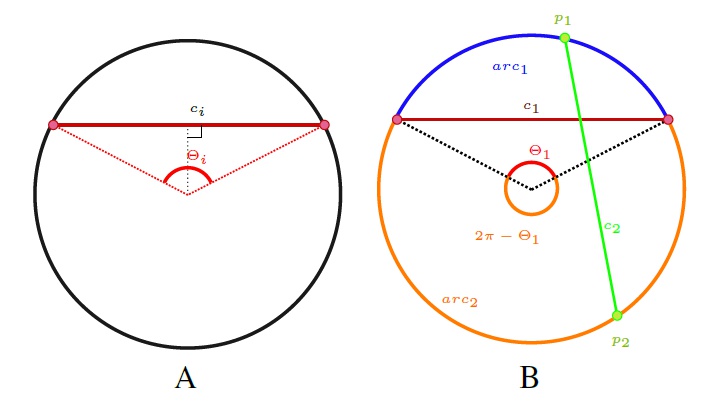}
       \CAPTION{
        Expected cut length and probability of cut intersection.
        {\bf A:} A unit circle cut $c_i$ with a central angle $\Theta_i$ can be considered w.l.o.g to be horizontal, 
                leading to an expected length as in the text. 
        {\bf B:} Two cuts $c_1$ and $c_2$ intersect if and only if the vertices of the second cut (in green) lie 
                 in different arcs (in blue and orange) generated by the first cut (in red). 
       }
       \label{fig:stats}
\end{figure}

Since circles are symmetric, without loss of generality we can align the coordinate system parallel to the cut and consider only horizontal chords that lie in the circle's upper half, i.e., when both $\vec{p_1}$ and $\vec{p_2}$ have identical positive $y$ coordinates, as in Fig.~\ref{fig:stats}A.
If we now assume (w.l.o.g) that $\phi_2>\phi_1$, then  $\Theta_i=\phi_2-\phi_1$ is the central angle of the cut and therefore $l_i = 2 \sin(\Theta_i / 2)$. Since $\Theta_i \sim \text{U}(0, \pi)$, it follows that the expected length of a random cut through the unit circle is
{\small
\begin{align*}
E[l_i]
&=
\int_{0}^{\pi} l_i(t) \cdot f_{\Theta_i}(t) d t
=
\int_{0}^{\pi} 2 \sin\left(\frac{t}{2}\right) \frac{1}{\pi} d t
=
\frac{4}{\pi}
\approx 1.273\;.
\end{align*}
}

\subsection{Probability of cut intersections}
\label{sec:prob_intersect}

Given two uniformly distributed random cuts $c_1$ and $c_2$, one may seek the probability of their intersection. This question is interesting for understanding how the number of pieces grows with the number of cuts, as intersecting cuts contribute more pieces than non-intersecting ones. Again, we can assume w.l.o.g that one of the cuts, say $c_1$, is horizontal and lying in the upper half of the circle (marked red in Fig.~\ref{fig:stats}B).
Let the central angle of $c_1$ be $\Theta_1 \sim \text{U}(0, \pi)$ and note how this cut divides the circle to two arcs - $arc_1$ of angle $\Theta_1$ (blue in Fig.~\ref{fig:stats}B) and $arc_2$ of angle $2\pi - \Theta_1$ (orange in Fig.~\ref{fig:stats}B).

Denoting the vertices of $c_2$ as $p_1$ and $p_2$, we first
note that an intersection between $c_1$ and $c_2$ occurs
if and only if $p_1$ belongs to $arc_1$ and $p_2$ belongs to $arc_2$ (or vice versa). Seeking the probability of such an event, let $I_{c_1, c_2}$ be an indicator function for the intersection between $c_1$ and $c_2$. Clearly, this function depends on the extent (or size) of the two arcs and indeed
{\small
\begin{align*}
P(I_{c_1, c_2} | \Theta_1)
&=
2 \cdot
P(p_1 \in arc_1 | \Theta_1) \cdot
P(p_2 \in arc_2 | \Theta_1)
\\
&=
2
\cdot
\frac{\Theta_1}{2\pi}
\cdot
\frac{2 \pi -  \Theta_1}{2 \pi}\\
&=
\frac{\Theta_1(2\pi - \Theta_1)}{2\pi^2}
\end{align*}
}
It follows that the expected value for the intersection event is
{\small
\begin{align*}
E[I_{c_1, c_2}]
=
P(I_{c_1, c_2})
&=
\int_{0}^{\pi}
f_{\Theta_1}(t) \cdot P(I | \Theta_1 = t) dt\\
&=
\int_{0}^{\pi}
\frac{1}{\pi}
\frac{t(2\pi - t)}{2\pi^2}
d t
=
\frac{1}{3} \;.
\end{align*}
}
Hence, we conclude that only 1 out of 3 pairs of random unit circle cuts will intersect, an event perhaps less frequent than intuitively anticipated in such circumstances. An intuitive justification nevertheless arises once we consider 4 endpoints on the shape's circumference and all 3 combinations of crossing lines they facilitate. Indeed, only one of these combinations induces a crossing.

\subsection{Expected total number of cut intersections}
\label{sec:num_intersect}

Following the probability of the cut intersection event (Sec.~\ref{sec:prob_intersect}), we now can seek the total number of intersections expected in a puzzle of $a$ cuts. Clearly, it is simply the sum of all pairs of intersecting cuts, that is
\begin{align*}
    N_{intersect}
    &=
    \frac{1}{2}
    \sum_{i=1}^{a}
    \sum_{j \neq i}
    I_{c_i, c_j} \;.
\end{align*}
The expected value of this random variable, i.e., the expected number of intersections in puzzles with $a$ crossing cuts, thus becomes:
\begin{align*}
E\left[N_{intersect}\right]
&=
\binom{a}{2}
	\underbrace{
P\left[
I_{c_1, c_2}
\right]
}_{\frac{1}{3}}
=
\frac{a(a-1)}{6} \;.
\end{align*}
Note that this number is far smaller than $\binom{a}{2}$, the maximum number of intersections possible between $a$ cuts.

\subsection{Expected number of edges}
\label{sec:expected_num_edges}

Given a crossing cuts puzzle generated by $a$ crossing cuts, we next wish to express the number of piece edges in the entire puzzle. This measure is fundamental to the number of matings and therefore is a substrate of the computational complexity of reconstruction algorithms.

First, observe that each edge is a subset of some cut between two consecutive intersections along its length. In particular, if a cut $c_i$ is intersected $k$ times, the number of edges that emerge from this cut will be $k + 1$.
To obtain the total number of edges $N_{edges}$ in the puzzle one needs to sum up the edges on all cuts, i.e.,
{\small
\begin{align}
    N_{edges}
    & =
    \sum_{i=1}^{a}
    \left(
        1+  \sum_{c_j \neq c_i} I_{c_i, c_j}
    \right) \nonumber \\
    &=
    a + \sum_{c_i} \sum_{c_j \neq c_i} I_{c_i, c_j} 
    =
    a + 2 N_{intersect}
    \;.
\label{eq:edge_num}
\end{align}
}
Since $I_{c_i, c_j}$ is a random function, so is $N_{edges}$. We can therefore seek its expected value, i.e., the expected number of edges in the entire puzzle:
{\small
\begin{align}
    E[N_{edges}]
    &=
    E\left[2 \cdot N_{intersect} \right] + a
    =
    2 \cdot \frac{a(a-1)}{6} + a
    =
    \frac{a^2+2a}{3} \;.
\label{eq:expected_edge_num}
\end{align}
}


\subsection{Expected edge length}
\label{sec:expected_edge_length}

\begin{figure}[!ht]
    \centering
    \includegraphics[width=0.5\columnwidth]{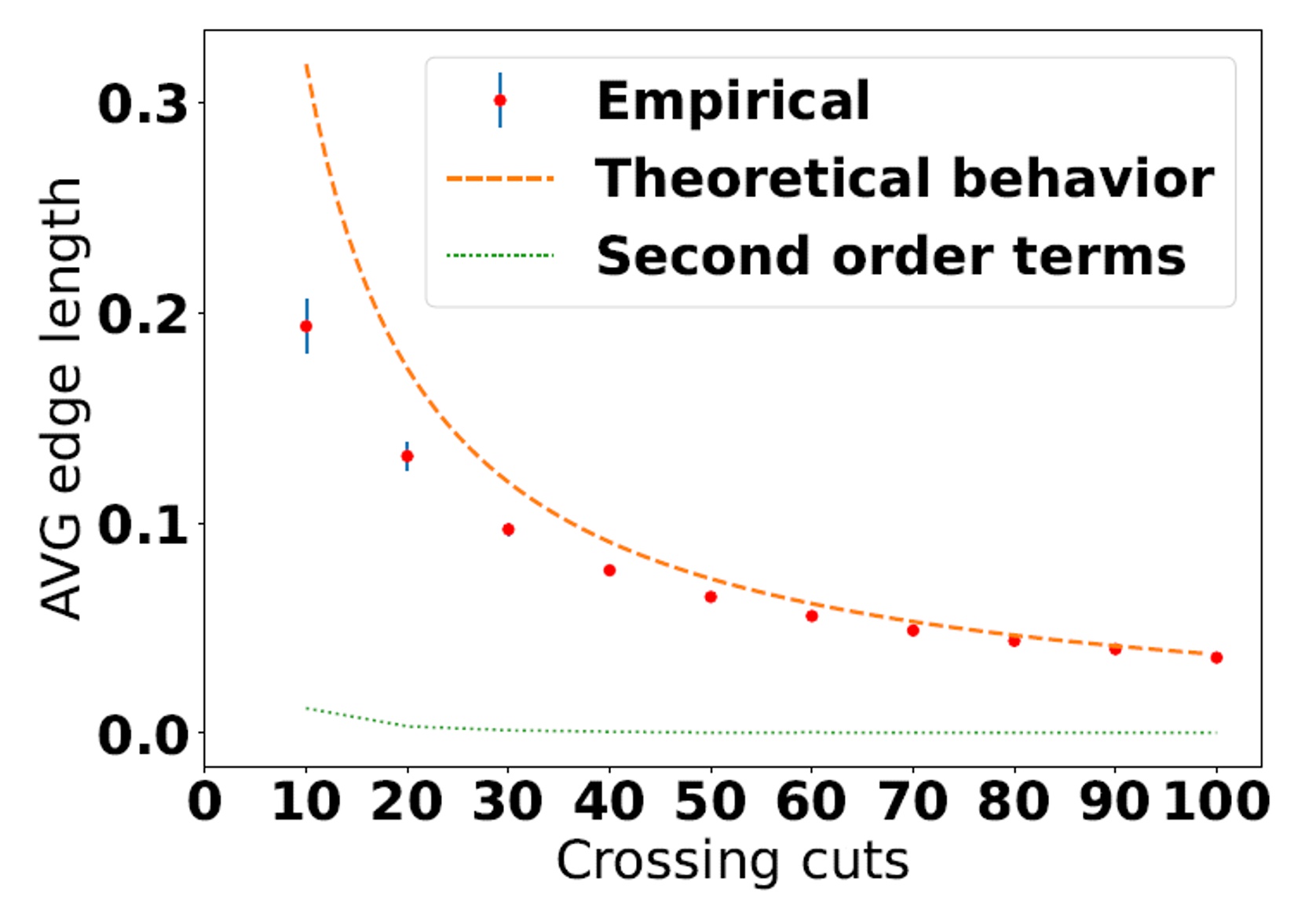}
    \CAPTION{
    Empirical average edge length with a growing number of cuts, as evaluated on DB1, compared to the first-order theoretical behavior from Eq.~\ref{eq:expectation_approximation} that improves in prediction power as the number of cuts increases. Error bars are $\pm 1$ SE.
    The green line shows the diminishing second-order terms of the Taylor approximation. 
    }
\label{fig:emp_stats_B}
\end{figure}

With the expected number of edges resolved, we can now seek the expected edge length as the expected ratio between the accumulated edge lengths to their number. Fortunately, the former is simply the summed length of all cuts and thus, if the puzzle constitutes $a$ cuts, we obtain an average edge length of
\begin{align*}
    l_{avg} = \frac{\sum_{i=1}^a l_i}{N_{edges}} \;,
\end{align*}
where $l_i$ was obtained in Sec.~\ref{sec:expected_cut_length} and $N_{edges}$ was derived in Sec.~\ref{sec:expected_num_edges}.
While the expected value of a ratio is \textit{not} the ratio of expected values, it \textit{is} its first-order Taylor approximation~\cite{Benaroya2013}. Thus:
\begin{align}
    E[l_{avg}]
    &=
    E\left[
    \frac{\sum l_i}{N_{edges}}
    \right]
    \approx
    \frac{E\left[\sum l_i \right]}{E\left[N_{edges} \right]}
    =
    \frac{12}{\pi(a + 2)}
\label{eq:expectation_approximation}
\end{align}
which conforms well with the empirical results of the same measure as shown in Fig.~\ref{fig:emp_stats_B}.
The second-order Taylor approximation
{\footnotesize
\begin{align*}
    E[l_{avg}]
    \approx 
    \underbrace{
    \frac{E\left[\sum l_i \right]}{E\left[N_{edges} \right]}
    }_{\text{First order terms}}
    - 
    \underbrace{
    \frac{
    Cov(\sum l_i, N_{edges})
    }{
    (E[N_{edges}])^2
    }
    + 
    \frac{
    Var( N_{edges}) \cdot E\left[ \sum l_i \right]
    }{
    ( E[N_{edges}] )^3
    }
    }_{\text{Second order term}} 
\end{align*}
}
constitutes two second-order terms that turn out to cancel each other to a diminishing sum as the number of cuts increases, thus facilitating the approximation in Eq.~\ref{eq:expectation_approximation}. This also is exemplified empirically in Fig.~\ref{fig:emp_stats_B}.
As mentioned in the introduction to this section, empirical results are evaluated on DB1, the circular puzzles dataset, described in Sec.~\ref{sec:data_synthesis}

\subsection{Edge length distribution}
\label{sec:dist_edge_length}

\begin{figure}[h!]
    \centering
    \includegraphics[width=0.85\columnwidth]{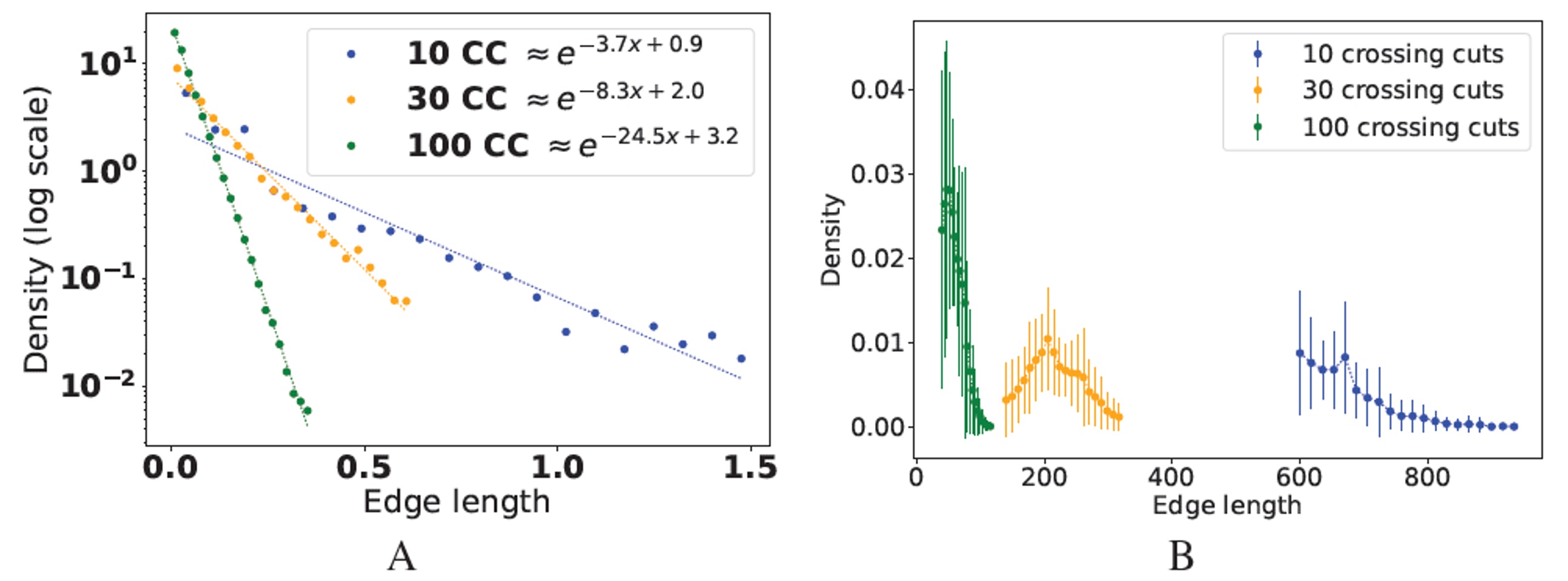}
    \CAPTION{
    Probability density of edge length in crossing cuts puzzles.
    {\bf A:} Empirical distribution for growing number of cuts, as evaluated on {\em noisy} puzzles from DB1.  Note how the distribution gets narrower with the number of crossing cuts, as depicted by a steeper slope in the log scale.
    {\bf B:} Empirical edge length distribution of perturbed square puzzles with a growing number of cuts, as evaluated on DB3.  Observe how the distribution of edge lengths in these puzzles is no longer exponential, with a modal behavior that is dominated by a particular edge length. 
    }
\label{fig:emp_edge_length_dist}
\end{figure}

While the expected edge length can be computed analytically, it is far more complicated to do so for the entire distribution of edge lengths. The importance of this distribution lies in how it influences the number of possible mates under geometric noise, a quantity that is likely to increase the narrower the distribution becomes. We have therefore measured this property empirically using our synthesized datasets and Fig.~\ref{fig:emp_edge_length_dist}A reports these findings. Note how in general the distribution is exponential, preferring shorter edges and (not surprisingly) becoming narrower with a larger number of cuts.  
Clearly, when cuts are no longer selected uniformly, the distribution can definitely change shape. For example, strictly square noiseless puzzles will of course have a delta distribution for their edge lengths. Perturbed square noisy puzzles, i.e., those in our DB3, exhibit the distribution shown in Fig.~\ref{fig:emp_edge_length_dist}B.

\subsection{Min, max, and expected number of pieces}
\label{sec:expected_pieces}

One of the significant properties of jigsaw puzzles that clearly affects the complexity of their representation (and thus of possible solutions) is their number of pieces. Clearly, even if the number of crossing cuts is set, different cut patterns can create puzzles with varying numbers of pieces.
To estimate this number, and inspired by Moore~\cite{moore1991using}, we use Euler's Formula for planar graphs:
\newtheorem{theorem}{Theorem}
\begin{theorem}[Euler's Formula]
If $G = (V, E)$ is any planar graph, then G has $|E| - |V| + 2$ regions where $|E|$ is the number of links in the graph and $|V|$ is the number of nodes.
\label{th:euler}
\end{theorem}
Note that in our crossing cuts puzzle case, the number of nodes for Euler's formula is the number of inner intersections ($N_{intersect}$) plus the $2a$ intersections of the cuts with the boundary of the puzzle. The number of links is the number of internal edges ($N_{edges}$) plus the $2a$ piece sides generated by the cuts along the puzzle boundary.
Using Euler's formula, and applying Eq.~\ref{eq:edge_num}, we thus get
\begin{align}
    N_{pieces}
    &=
    \underbrace{
    (N_{edges} + 2a)}_{|E|}
    -
    \underbrace{
    (N_{intersect} + 2a)
    }_{|V|}
    \label{eq:n_regions_raw}
    + 2 - 1\\
    &=
    N_{intersect} + a + 1
\label{eq:n_regions}
\end{align}
Note that the subtraction of the last 1 in Eq.~\ref{eq:n_regions_raw} is required since Euler's formula also counts the region outside the puzzle/graph.

\begin{figure}[!ht]
    \centering
    \includegraphics[width=0.5\columnwidth]{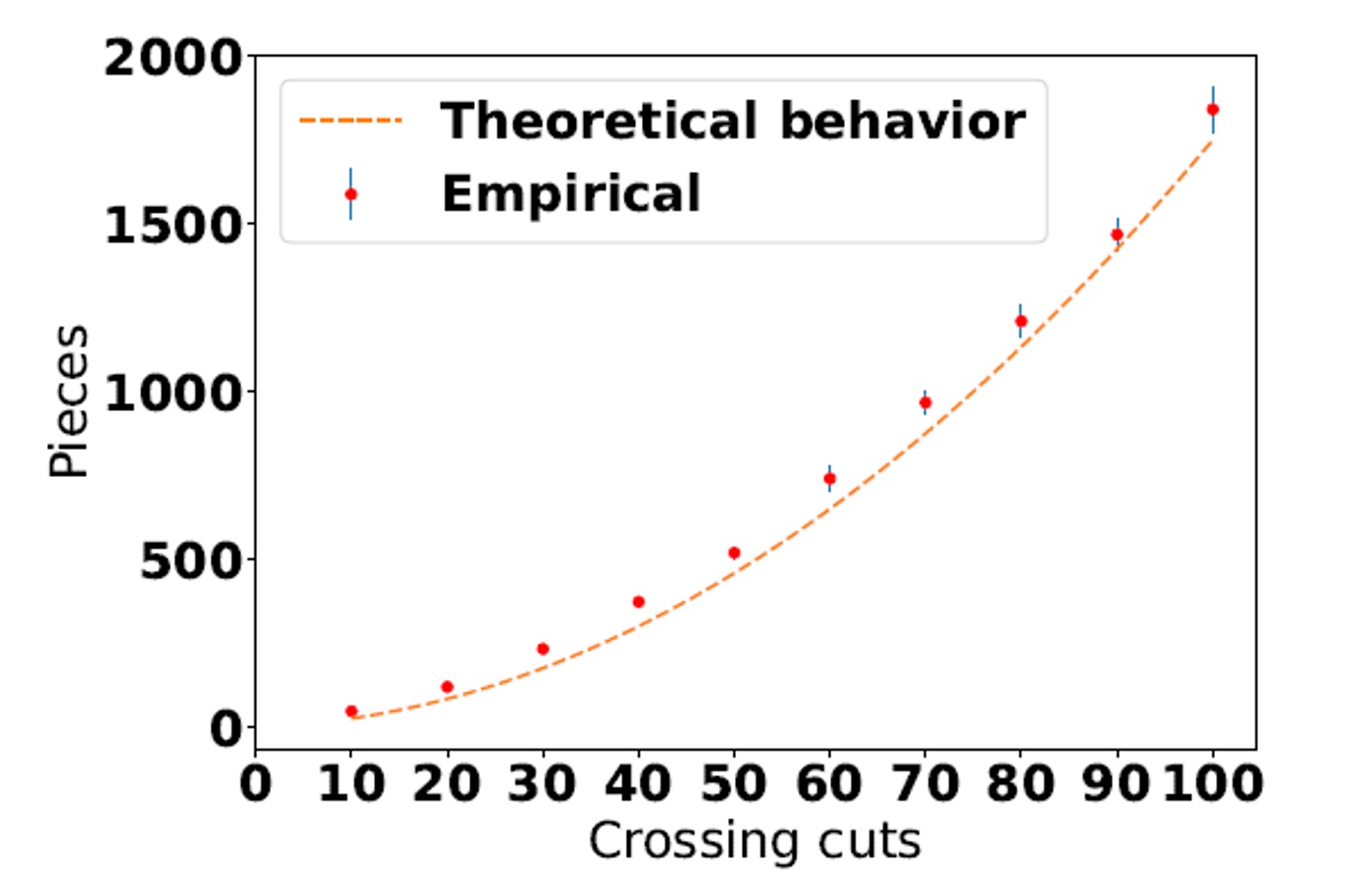}
    \CAPTION{
    The empirical expected number of puzzle pieces with a growing number of cuts, compared to the theoretically expected behavior (Eq.~\ref{eq:n_regions}). Error bars are $\pm 1$ SE.
    }
\label{fig:emp_stats_A}
\end{figure}

With this in mind, we next observe that one extreme case includes puzzles where no cut intersects others ($N_{intersect}=0$), and thus the minimal number of pieces is $N_{pieces}=a+1$.
At the other extreme, every cut intersects all others, and the $\binom{a}{2}$ intersections yield the following quadratic upper bound on the number of pieces (which is exactly the Lazy caterer's sequence)
\begin{align*}
    \max_{c_i, \dots c_a} N_{pieces}
    &=
    \binom{a}{2} + a + 1
    =
    \frac{a^2}{2}
    + \frac{a}{2}
    +
    1\;\;.
\end{align*}
However, with  $N_{intersect}$ being a random variable (that depends on the random cuts), it is more interesting to examine the \textit{expected} number of pieces:
\begin{align}
    E\left[ N_{pieces} \right]
    &=
    E[N_{intersect}]
    +a
    +1 \nonumber \\
    &=
    \frac{a(a-1)}{6}
    +a
    +1
    =
    \frac{a^2}{6} + \frac{5a}{6} + 1 \;.
\label{eq:exp_n_pieces}
\end{align}
This behavior can also be verified empirically, as shown in Fig.~\ref{fig:emp_stats_A}.
Finally, as the number of cuts increases, and when $a \to \infty$, the ratio between the expected and the maximum number of pieces becomes
\begin{align*}
\lim_{a\to\infty} \frac{E[N_{pieces}]}{\max N_{pieces}} = \frac{1}{3}
\end{align*}
which is the same as the probability for cut intersection found in Sec.~\ref{sec:prob_intersect}.

\subsection{Expected number of edges per piece}

As discussed in Sec.~\ref{sec:the_puzzle}, the crossing cuts puzzle model cuts the puzzle shape into convex polygonal pieces. Clearly, these pieces can have a different number of edges and there is no a-priori inherent limit to this number (except the number of cuts, of course).

\begin{figure}[!ht]
    \centering
    \includegraphics[width=0.5\columnwidth]{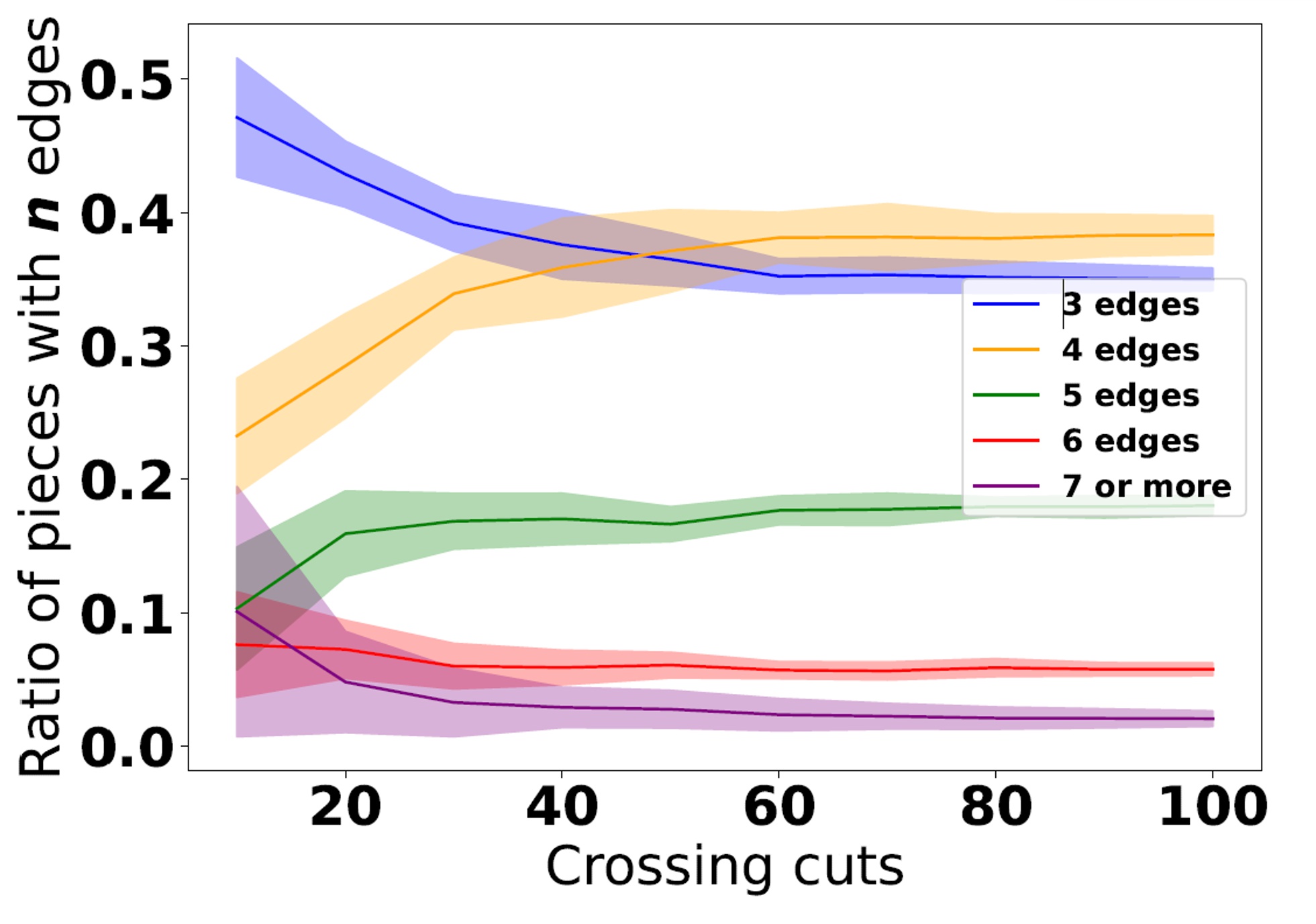}
    \CAPTION{ 
        Expected ratios of pieces with a particular number of edges as a function of the number of crossing cuts. Note how quadrilaterals are always the majority, followed closely by triangular pieces and the less frequent pentagons. These three classes of polygons quickly converge to account for approximately 95\% of all pieces. Note how the ratios converge quickly and remain essentially invariant to the number of cuts. Shaded bands represent $\pm 1$SE.}
\label{fig:emp_stats_C}
\end{figure}

To explore this property we conducted an empirical evaluation on DB1, i.e., by using the $30$ synthesized circular crossing cuts puzzles synthesized for each of the different number of cuts. Empirically, the most frequent pieces are either quadrilateral or triangular, depending on the number of cuts, whereas asymptotically, quadrilaterals are the most abundant. The probability of encountering puzzle pieces with more than 6 edges is diminishing quickly from $10\%$ in puzzles with few cuts, to approximately $2\%$ as the number of cuts increases.
The results up are shown in Fig.~\ref{fig:emp_stats_C} and demonstrate that the distribution converges quickly and then remains stable from approximately 60 cuts.

\subsection{Number of potential matings per edge}
\label{sec:stat_potential}

Since any puzzle reconstruction algorithm will seek (as part of its different computations) to match the edge of a given piece to edges of other pieces, the complexity of such an algorithm will relate intimately to the number of potential matings each edge may have. Clearly, the higher the number of admissible candidate mates, the more difficult the identification of the correct one is likely to be.
Following the discussion in Sec.~\ref{sec:noise}, the raw number of geometrically admissible mates each edge may have is determined by the two mating constraints $\tilde{C}_1$ and $\tilde{C}_2$ and it is naturally affected by the level of the noise. In fact, since the number of expected edges in the puzzle is quadratic in the number of cuts (cf. Sec.~\ref{sec:expected_num_edges}), a naive extension of Algorithm~\ref{alg:naive} from Sec.~\ref{sec:mating_constraints} that also incorporates backtracking when wrong matings are identified, will grow intractably in complexity by a factor of $O\left( k^{a^2} \right)$ if the number of potential matings per edge is $k$.

We empirically explored the expected average number of matings by counting the average number of possible matings for each puzzle in DB1 while employing $\tilde{C}_1$ and $\tilde{C}_2$.
Not unexpectedly, the results provided in Fig.~\ref{fig:emp_stats_D}A indicate that the noise level affects the number of potential mates very rapidly and very drastically (where the decline after the peak is because greater noise erodes more pieces completely, as shown in Fig.~\ref{fig:emp_stats_D}B).

\begin{figure}[h!]
    \centering
        \begin{tabular}{cc}
        \includegraphics[width=0.45\columnwidth]{avg_number_of_potential_matings_per_edge.jpeg}
        &
        \includegraphics[width=0.45\columnwidth]{avg_ratio_of_erased_pieces.jpeg}
        \\
        A & B 
        \end{tabular}
    \CAPTION{
        \textbf{A:}~The average number of geometrically admissible matings for each edge as a function of noise level. Each graph shows the potential number of mates that satisfy both $\tilde{C}_1$ and $\tilde{C}_2$, summed over all edges. Both the initial rapid growth and the subsequent gradual decline (which is particularly visible in puzzles with more cuts) indicate the harmful effect of noise. Shaded bands are $\pm 1$ SE.
        \textbf{B:}~The average ratio of completely eroded pieces due to applied noise, out of the puzzles' original (noiseless) number of pieces. 
        As expected, the application of larger noise completely erodes more pieces, thus also decreasing the number of potential matings per edge.   
    }
\label{fig:emp_stats_D}
\end{figure}

Here it is also worth re-emphasizing that the noise levels in our model are measured relative to the puzzle size, or its \textit{diameter}, and therefore might appear small. In reality, they are not small at all, because noise affects individual pieces, that typically are very much smaller than the entire puzzle. Thus, considering also the average edge length (cf. Sec.~\ref{sec:expected_edge_length}),  noise level $\xi$ relative to the puzzle diameter is comparable to the following bound
\begin{align}
    \bar{\xi} = \frac{4 \cdot \xi \cdot \pi(a + 2)}{12}
\label{eq:noise_to_edge_length}
\end{align}
relative to average edge length, which is perhaps a more tangible and informative measure. For example, in a puzzle of 20 crossing cuts ($84$ pieces on average) and a noise level of $\xi=1\%$, the noise relative to average edge length is~$\bar{\xi} \approx 10\%$, namely a rather significant noise.

Indeed, the high number of potential matings in the presence of noise suggests a similarly high branching factor in a naive ``search and backtrack'' algorithm, which will clearly become intractable for handling noisy (i.e., realistic) crossing cuts puzzles, even if the number of cuts is modest. Our goal is to seek heuristics that make the reconstruction more manageable after all, and as we will see later on, this can be achieved by utilizing multiple geometric constraints \textit{simultaneously}, and by leveraging the \textit{pictorial} content to generate and apply yet more constraints on the matings.




\section{Puzzle reconstruction under noisy conditions}
\label{chap:puzzle_reconstruction}

\begin{figure}[h!]
    \centering
    \includegraphics[width=0.7\columnwidth]{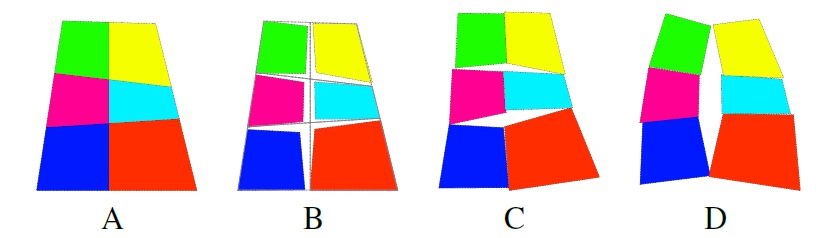}
    \CAPTION{
             The placement of noisy pieces can be ambiguous. When noise is applied to this simple crossing cuts puzzle (A,B), there could be different placements of the noisy pieces that might count as ``correct'' (C,D).}
    \label{fig:spring_motivation}
\end{figure}

Recall that a ``realistic'' crossing cuts puzzle constitutes a representation of the input pieces (and some bound on the erosion noise),  and it seeks as output both the correct matings and the geometric transformation of each piece.
As mentioned above, at first sight one may wish to extend the initial greedy Algorithm~\ref{alg:naive} from Sec.~\ref{sec:mating_constraints} while using the relaxed ``noisy'' constraints ($\tilde{C}_1$ and $\tilde{C}_2$) to find candidate matings, and if needed employ backtracking upon failures (e.g., piece collisions). However, as analyzed above, the expected number of candidate matings per edge (cf. Sec.~\ref{sec:stat_potential}) clearly makes this naive extension intractable. Moreover, under noise, it is unclear what is the desired position (i.e., Euclidean transformation) of each piece, or how to compute it in the first place, even if the mating relationships are resolved correctly. Fig.~\ref{fig:spring_motivation} illustrates some of these challenges.

To address these difficulties we approach the problem in stages, and in particular, we begin with the simpler problem of solving the puzzle \textit{when the correct matings are given also}.
More concretely, we first suggest a solution to this sub-problem by representing it as a multi-body spring-mass system where energy minimization is sought while the spring attractive forces apply between corresponding vertices. The solutions obtained this way are then used as scores for searching and determining the correct matings while incorporating a hierarchical (and progressively growing) set of circular constraints among adjacent pieces.
For pictorial puzzles, we also add another set of pictorial constraints on top of the geometrical ones.


\subsection{Noisy puzzle solving with \textit{known} matings}
\label{sec:psy}

Let $P=\{p_1,p_2,\ldots,p_n\}$ be the set of pieces and let $M=\{m_1,\ldots,m_{|M|}\}$  be a set of (known) pairwise matings $m_q=\{ e_i^j, e_k^l \}$ between their corresponding edges.  We seek a computational scheme that obeys the given matings and places the pieces in some ``optimal'' or ``good'' way next to each other.
Intuitively, we would like to do so in a way that minimizes the total distance, i.e., the $L_2$ displacement error, between corresponding mating vertices, or more formally, to find the set of Euclidean transformations $(R_i,\vec{t}_i)$ that satisfy
\begin{align}
    \underset{\left(R_i, \vec{t\,}_i\right)}{\mathrm{argmin}}\;\;
    \sum_{(\vec{v\,}_i^j, \vec{v\,}_k^l)}
    \left\lVert
    \left(R_i \vec{v}_i^j + \vec{t}_i\right) -
    \left(R_k \vec{v}_k^l + \vec{t}_k\right)
    \right\rVert^2
    \;,
\label{eq:displacement_error}
\end{align}
where $\vec{v\,}_i^j$ and $\vec{v\,}_k^l$ are the corresponding vertices of the matings defined by $M$ while $(R_i, t_i)$ and $(R_k, t_k)$ are the euclidean transformations of pieces $p_i$ and $p_k$ that own these vertices. Unfortunately, this is no simple least squares minimization, as the unknowns include rotation matrices and the sought-after transformations must satisfy the constraint that they are {\em identical for all vertices of the same piece}.

\begin{figure}[t!]
    \centering
      \includegraphics[width=0.95\columnwidth]{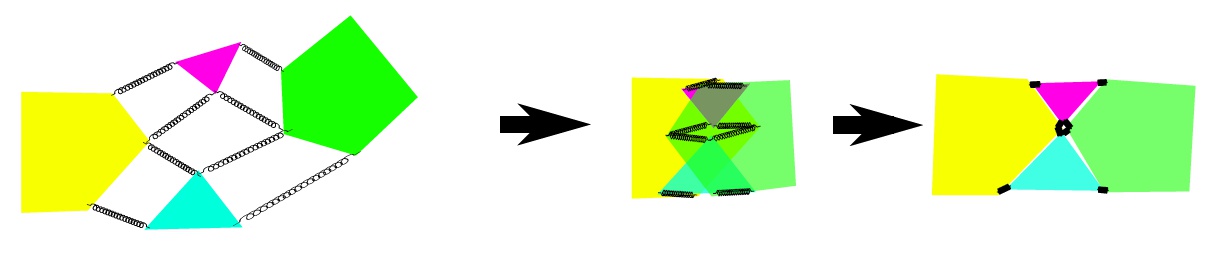}
      \CAPTION{
               The puzzle with given matings is abstracted as a spring-mass system that evolves over time towards a convergence state.
               If the pieces are far apart (left), the springs pull them closer. When the pieces overlap (middle), the springs pull them apart again. With some damping (i.s., loss of energy due to friction), the system eventually converges to a state of minimal energy.
               }
  \label{fig:joint_def_A}
\end{figure}

As a result of its specifications, this optimization problem defies analytical solutions and we therefore resort to tools from other disciplines. In particular, we propose to abstract the rearrangement problem as a \textit{multi-body spring-mass system}. To do so we first represent our puzzle pieces as 2D rigid bodies with uniform density, and therefore with mass that is proportional to their area.  We then connect all pairs of corresponding vertices (i.e., those matched by the matings) with springs of zero length and identical elasticity (i.e., having the same \textit{spring constants}).
Since the elastic potential energy of such a spring-mass system is $U(x) = \sum_{l} \frac{1}{2} k x_l^2$, where $x_l$ is the displacement from equilibrium length of spring $l$, it is identical (up to a constant) to our objective function in Eq.~\ref{eq:displacement_error}. We therefore apply numerical methods for solving multi-body spring-mass problems, while the initial pose (position and rotation) of each piece is chosen randomly inside the arena. The physical system is then set loose and with some damping (i.s., loss of energy due to friction) it converges to its minimal energetic state, as illustrated in Fig.~\ref{fig:joint_def_A}.

\begin{figure}[h!]
    \centering
     \includegraphics[width=0.8\columnwidth]{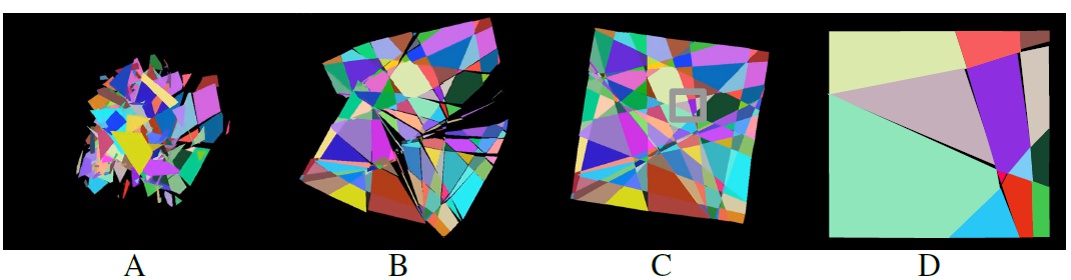}
        \CAPTION{ 
            Several snapshots of the numerical simulation for a noisy puzzle of 25 cuts and 940 pieces.
            {\bf A:} Initial state.
            {\bf B:} Intermediate state.
            {\bf C:} Final state.
            {\bf D:} A zoomed section of the region marked in the final state shows the approximated final placement due to the noise.
        }
  \label{fig:joint_def_B}
\end{figure}

In practice there are off-the-shelf tools to solve the above system numerically, practically simulating the dynamical process that the system undergoes from initial condition until convergence, and here we use the Box2D physics engine~\cite{box2d}. For puzzles, it is undesired to obtain solutions with overlapping pieces, but adding this constraint to a random initial state is unstable numerically. We therefore run the process first while allowing the pieces to overlap. The convergence state of this run is energetically minimal but might include small overlaps. We then use it as the initial state for a second run,  this time while forbidding overlaps. The end result is our solution and Fig.~\ref{fig:joint_def_B} shows several snapshots from this dynamical process.

\subsection{Noisy puzzle solving with \textit{unknown} matings}
\label{sec:noisy_reconstruction}

Let $P=\{p_1, \dots p_n\}$ be the set of puzzle pieces and let $\varepsilon$ denote the noise level.
Unlike the conditions in the previous section, we now assume no knowledge of the matings and thus our goal is twofold: to find the correct matings $M=\{m_1,\ldots,m_{|M|}\}$ between the edges \textit{and} the geometrical transformation of each piece.
To do so, we endow the basic constraint matching procedure (based on $\tilde{C}_1$ and $\tilde{C}_2$) with a modified version of a hierarchical loops scheme~\cite{son2018solving}, where the mass-spring minimization approach from Sec.~\ref{sec:psy} is used to score the loops based on their success to position the pieces properly, as defined below. If the puzzle is pictorial, we also rank and filter those matches using the pictorial content next to the geometrical one.

\subsubsection{Hierarchical layered loops}

As is usually done in jigsaw puzzle solvers, we start by finding candidate mates for each edge by aggregating the set of all unordered pairs of edges that satisfy the constraints $\tilde{C}_1, \tilde{C}_2$ (cf. Sec.\ref{sec:noise}). We denote this set by $\tilde{M}$
{\small
\begin{align*}
    \tilde{M} =
    \left\{
    \{e_i^j, e_k^l\} \Bigm|
    e_i^j, e_k^l \in E  \;\wedge\;
    \tilde{C}_1\left(e_i^j, e_k^l\right) \;\wedge\;
    \tilde{C}_2\left(e_i^j, e_k^l\right)
    \right\}\;,
\end{align*}
}
and recall that the higher the noise level, the more numerous are the potential matings, as analyzed in Sec.~\ref{sec:stat_potential}.

\begin{figure}[h!]
  \centering
    \includegraphics[width=0.8\columnwidth]{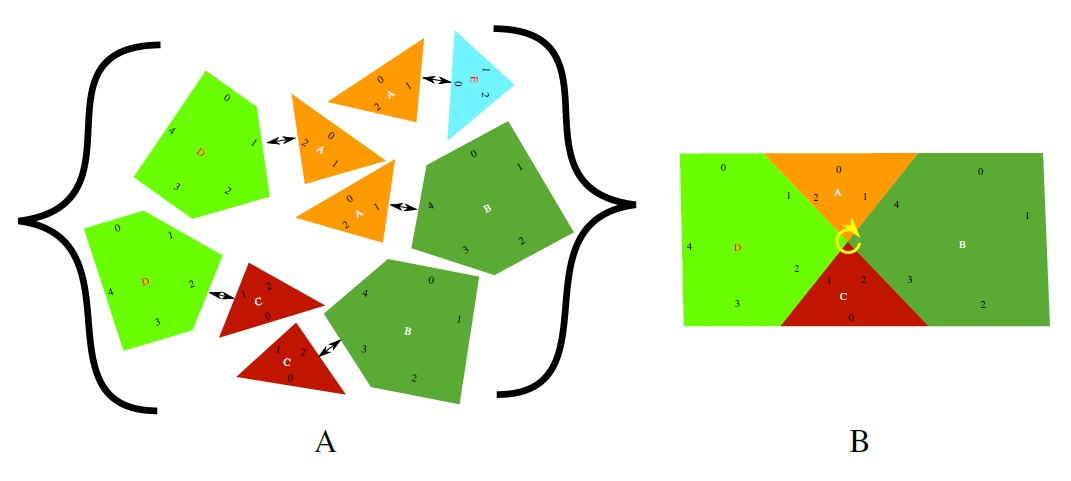}
  \CAPTION{
    0-loops formation from pairwise matings.
    \textbf{A:}~A bag $\tilde{M}$ of 5 potential matings, of which $(e_A^1, e_E^0)$ is wrong but included because it satisfied both $\tilde{C}_1$ and
               $\tilde{C}_2$.
     \textbf{B:}~The loop $(e_A^1, e_B^4) \to (e_B^3, e_C^2) \to (e_C^1, e_D^2)\to (e_D^1, e_A^2)$ is identified and supports the plausibility of its
                constituent matings. Note that the path ending with $(e_A^1, e_E^0)$ does not close a loop because the mating $(e_E^{2}, e_C^2)$ is not present in the bag.
}
\label{fig:0loops}
\end{figure}

As mentioned earlier, in crossing cuts puzzles with uniformly distributed random cuts, the probability of more than two cuts meeting at a point is nil (cf. Sec.\ref{sec:mating_constraints}).
It directly follows that all \textit{inner} puzzle junctions constitute exactly four pieces.
We utilize this property to identify ordered lists of 4 mating candidates that form such junctions, or \textit{loops}, as illustrated in Fig.\ref{fig:0loops}.
Formally, a mating loop in the \textit{clockwise} direction is a 4-tuple
{\small
\begin{align}
 & (m_1, m_2, m_3, m_4)
= \nonumber \\
& \left(
\left\{e_A^{j_A}, e_B^{i_B}\right\},
\left\{e_B^{j_B}, e_C^{i_C}\right\},
\left\{e_C^{j_C}, e_D^{i_D}\right\},
\left\{e_D^{j_D}, e_A^{i_A}\right\}
\right)
\label{eq:4_tuple_loop}
\end{align}
}
such that $m_k \in \tilde{M}\; \forall k=1..4$ and the following conditions hold:
\begin{description}
    
     \item[Cond 1:] No piece appears twice, i.e. $p_A \neq p_B \neq p_C \neq p_D$ (i.e., $A\neq B\neq C\neq D$).

     \item[Cond 2:]  If a mating in the loop ``enters'' a piece $p$ though its $e_p^i$ edge, the consecutive mating ``exists'' the same piece through the adjacent edge $e_p^j=e_p^{(i-1) \mod N_p}$, where $N_p$ is the number of $p$'s edges (and also vertices; cf. Sec.~\ref{sec:the_puzzle}). In other words,  it ``exits'' through an edge immediately \textit{counterclockwise} to $e_p^i$ along the piece border.
     See edges $e_{B}^4$ and $e_{B}^3$ in Fig.~\ref{fig:0loops}B for an example.

     \item[Cond 3:]  The loop begins and ends with the same piece. This is in fact true by the definition in Eq.~\ref{eq:4_tuple_loop} as both the first and last matings contain the same edge of piece $p_A$.

\end{description}

Since these basic loops are the building blocks for the puzzle reconstruction, and since their number is polynomial in the number of candidate matings,  we search for them exhaustively among all $O\left(|\tilde{M}|^4 \right)$ possible mating 4-tuples, keeping only those that satisfy all of the above constraints. However, to nevertheless spare $75\%$ of combinations and avoid searching and storing all 4 circular shift permutations of the same loop, we force loops to start with the edge having the lowest index.

\begin{figure}[b!]
 \centering
    \includegraphics[width=0.95\columnwidth]{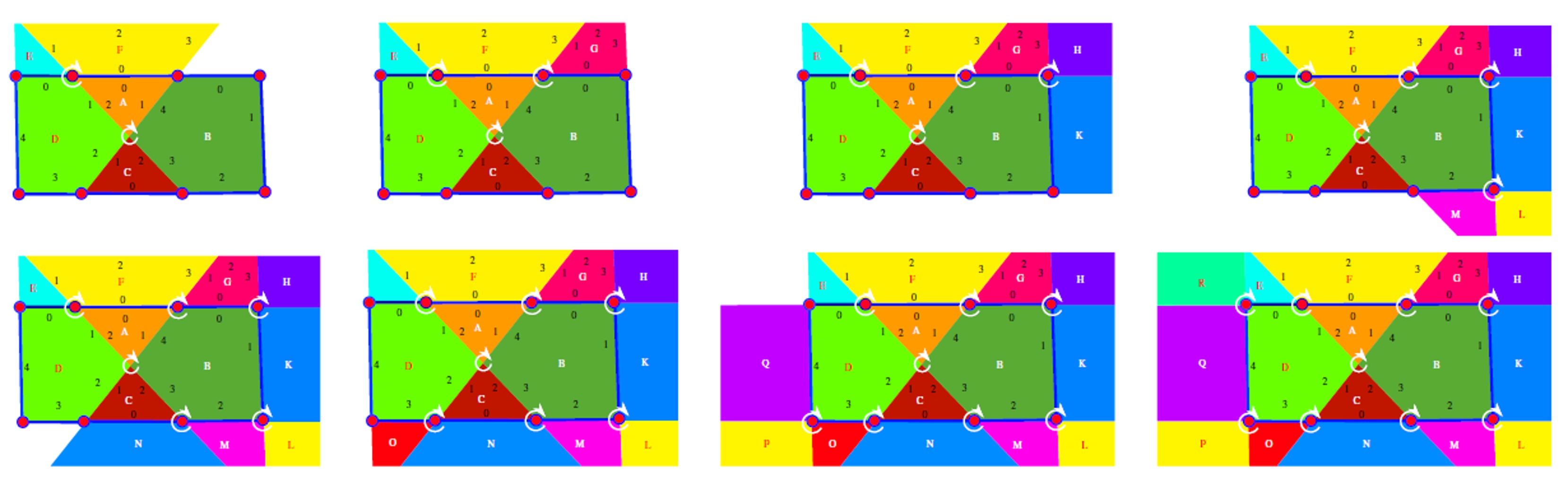}
  \CAPTION{
    Hierarchical loops formation. Shown left to right, top to bottom, the formation of a higher-level loop (in this case, 1-loop) is done by scanning the boundary of the lower-level loop (in this case, 0-loop) until looping all around it with consistent 0-loops that match existing matings and pieces. In particular, the last added 0-loop much also match the first added one.
}
\label{fig:xloops}
\end{figure}

Let now $\mathcal{L}$ be the bag of basic loops computed as above. We now exploit partial \textit{overlaps} between loops to identify \textit{correct} matings more robustly instead of relying on $\tilde{M}$ matings alone.
More specifically, the next stage of the puzzle reconstruction algorithm is searching for ``higher-order'' loops, i.e., loops of loops, or \textit{hierarchical loops}~\cite{son2018solving}.
Denoting the basic 4-tuple loops in $\mathcal{L}$ as 0-loops, we  now seek all possible $x$-loops by trying to \textit{fully enclose} $(x\!\!-\!\!1)$-loops with partially overlapping $0$-loops, as illustrated  in Fig.~\ref{fig:xloops}.
Toward that end, let $(e_1,\allowbreak e_2\allowbreak \dots e_k)$ be the list of edges along the boundary of some $(x\!\!-\!\!1)$-loop. For example, the boundary of the 0-loop in Fig.~\ref{fig:0loops}B is
$(e_A^0, e_B^0,\allowbreak e_B^1, e_B^2, e_C^0, e_D^3,\allowbreak e_D^3, e_D^4, e_D^5)$.
Starting with $e_1$ and ending with $e_k$, we progressively construct a higher level $x$-loop by searching and merging a proper 0-loop from $\mathcal{L}$ that matches a \textit{sub-loop} of the current $x$-loop around $e_i$.
For example, if we start from the boundary edge  $e_A^0$ in Fig.~\ref{fig:0loops}B, we look for 0-loops that not only include that edge but also include edges from piece $p_B$, i.e., the mating $\left\{ e_B^4, e_A^1 \right\}$ and the edge $e_B^0$.
As shown in Fig.~\ref{fig:xloops}, the loop that was found in this particular example constitutes
$
\left(
\left\{ e_{A}^0, e_F^0 \right\},
\left\{ e_{F}^3, e_G^1 \right\},\allowbreak
\left\{ e_{G}^0, e_B^0 \right\},\allowbreak
\left\{ e_{B}^4, e_A^1 \right\}
\right)
$.
Typically, and unless it is near the corner of the $(x\!\!-\!\!1)$-loop, the 0-loops that are identified for merging will need to match an existing sub-loop of at least 3 edges and at least one mating. And yet, despite these multiple constraints, it is possible that more than one 0-loop in $\mathcal{L}$ will match around some boundary edge of the current $(x\!\!-\!\!1)$-loop, and consequently, it is possible that more than one $x$-loop will fit around a given $(x\!\!-\!\!1)$-loop. In such cases, we generate and store them all for subsequent processing.

\begin{figure}[h]
    \centering
  \includegraphics[width=0.85\columnwidth]{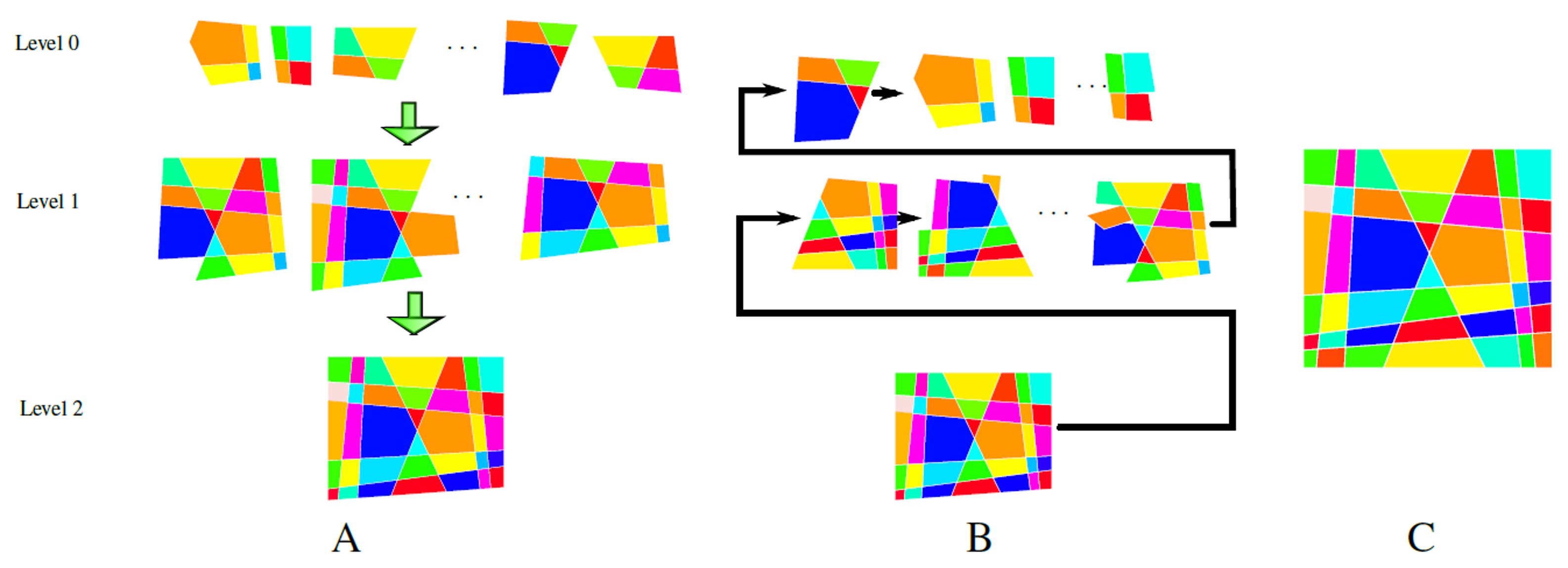}
    \CAPTION{
             The outline of the reconstruction process.
             \textbf{A:}~Hierarchical loops of all levels are found, where each level is used as a starting point to search for the next level. Note that the max-level loop may not cover the entire puzzle.
             \textbf{B:}~The hierarchical loops are merged by iterating over the loop levels in decreasing order. Each level is merged in increasing order based on the score.
             \textbf{C:}~The merged pieces are positioned using the spring-mass system.
    }
    \label{fig:whole}
\end{figure}

The process just described constructs the hierarchical loops in ``layers'' to produce a bag of $x$-loops for each layer $x$. Each of the $0$-loops in $\mathcal{L}$ may produce several $1$-loops, each of them may produce several $2$-loops, and so forth, until a layered representation is established, as illustrated in Fig.~\ref{fig:whole}A.
This process terminates at level $x_{\max}$ if not even a single $(x_{\max}\!\!+\!\!1$)-loop can be constructed, an event likely to happen if such loops overflow beyond the true (though unknown) puzzle boundary.

\subsubsection{Ranking hierarchical loops}
\label{sec:ranking_loops}

Although hierarchical loops require simultaneous consensus between growing numbers of participating matings, and thereby reduce significantly the possibility of wrong combinations, false positives are still possible due to the noise.
To rank better and worse loops,  we utilize the fact that each of them is a small noisy puzzle of pieces $P_{loop}$ and (known) matings $M_{loop}$ (cf. Sec. 6.1), and that ``correct'' loops can be "solved" for their spatial transformations with little to no overlaps even when collisions are allowed when we follow the multi-body spring-mass mechanism from Sec~\ref{sec:psy}.
We therefore employ this scheme and rank the different $x$-loops by their convergence state. We first define the following ``quality'' measure
{\footnotesize
\begin{align}
\label{eq:Qoverlap}
Q_{overlap}(P_{loop}, M_{loop})
&=
    \sum_{p_i\in P_{loop}}
	\frac{
	\left|
	  A(p_i) \cap \left(
	  \bigcup_{p_j \neq p_i}A(p_j)
	  \right)
	  \right|
	  }
	{\left| A(p_i) \right|}
\end{align}
}
where $A(p_i)$ represents the \textit{region} (as a set of points) of piece $p_i$ in its final pose $(R_i,\vec{t\,}_i)$ and the measure as whole is a modified Dice coefficient~\cite{dice1945measures} between each piece and the rest of the pieces.
Since the distance between all adjacent vertices in ``correct'' loops also must be small, we also consider the distances between corresponding vertices as defined by $M_{loop}$ measured \textit{after} collisions are prohibited:
{\small
\begin{align*}
Q_{dist}(P_{loop}, M_{loop})
&=
\sum_{\vec{v\,}_i, \vec{v\,}_{i'}} \lVert \vec{v\,}_i - \vec{v\,}_{i'} \rVert^2 \;.
\end{align*}
Combining both scores into one rank we get:
\begin{align*}
Q_{loop}(P_{loop}, M_{loop})
= &
w_1 \cdot Q_{overlap}(P_{loop}, M_{loop}) 
+ \nonumber\\
& w_2 \cdot Q_{dist}(P_{loop}, M_{loop})
\end{align*}
}
and while the weights can prioritize one score over the other, in our evaluation we found that $w_1=w_2=1$ produces excellent results and that sensitivity to these values is very small.

\subsubsection{Merging hierarchical loops}

Even with the best hierarchical loop found at the maximum level, the process of puzzle reconstruction is not yet finished since the maximum level of hierarchical loops does not necessarily cover the entire puzzle (e.g. the 2-loop in Fig.~\ref{fig:whole}A).
To complete the process and obtain the matings for the complete puzzle we now attempt to merge hierarchical loops. The $x$-loops are first sorted at each level $x$ according to their rank $Q$ (Sec.~\ref{sec:ranking_loops}), and this list is then scanned from the best and highest level loops
(Fig.\ref{fig:whole}B).

More formally, let $P_{agg}, M_{agg}$ denote the pieces and matings of the merging (or aggregation) process, initialized to be the best $x_{max}$-loop. Scanning now the sorted list of all $x$-loops, each is merged into the aggregated structure if several conditions hold. Assuming the pieces of the current $x$-loop under consideration are $P_{loop}$, and they are connected with $M_{loop}$ matings,  this loop is merged into  $P_{agg}, M_{agg}$ if

\begin{itemize}
    \item at least one piece is shared with the aggregated structure, i.e $P_{agg} \cap P_{loop} \neq \emptyset$,
    \item at least one piece is novel, i.e $P_{agg} \cup P_{loop} \neq P_{agg}$, and
    \item there is no contradiction between the matings in $M_{agg}$ and $M_{loop}$, i.e. if $\{e_A^i, e_{B}^j\} \in M_{loop}$ then
          either $\{ e_A^i, e_{B}^j\} \in M_{agg}$ \textit{or} none of the matings in $M_{agg}$ contains edges $e_A^i$ or $e_B^j$.
\end{itemize}
The merging process continues through the lowest ranked $0$-loop, and is then repeated from the start until $M_{agg}$ no longer changes during a full scan. This process must converge since the aggregation can include each possible mating at most once.

After the aggregated structure converges, the multi-body spring-mass process is performed one last time to position all the pieces $P_{agg}$ properly based on the obtained mating $M_{agg}$. The result is the final reconstructed crossing cuts puzzle.


\subsection{Incorporating \textit{Pictorial} constraints}

As the analysis of puzzle properties showed, larger geometrical noise increases rapidly the number of potential mates that are found using the geometrical constraints ($\tilde{C}_1$ and $\tilde{C}_2$) (cf. Sec~\ref{sec:stat_potential}). Similar effect is induced by increasing the number of cuts.
In these cases, using the pictorial content of the piece can provide a big advantage.
In particular, while the initial set $\tilde{M}$ of potential matings can be obtained using geometrical constraint, scoring and ranking these matings based on pictorial content may drastically reduce admissible matings and thus the computational effort of the reconstruction algorithm discussed in Sec.~\ref{sec:noisy_reconstruction}.

It should be emphasized from the outset that, unlike geometrical constraints, pictorial content alone cannot exclude matings with full certainty, as two genuinely neighboring pieces may legitimately have drastically different pictorial content even along their abutting boundaries. A solver can thus ''take risks'' and heuristically excludes pictorial matches below some predefined fidelity threshold, but strictly speaking, the pictorial content can help only in \textit{prioritizing} certain matings over others, and therefore it can merge naturally into the ranking process described in Sec.~\ref{sec:noisy_reconstruction}.

Similar to methods proposed for solving other pictorial puzzles in the literature, mostly in the context of square jigsaw puzzles (see Sec.~\ref{sec:related_work}), a pictorial compatibility score can be based on some dissimilarity measure of the colors along the \textit{edges} or \textit{margins} of puzzle pieces, while paying less attention to the pictorial information deeper inside each piece. However, unlike in the common case studied in the square jigsaw puzzle literature, here our setting is far more challenging, for several reasons.
First, pieces in crossing cuts puzzles are essentially never aligned with the pixel grid, making both the representation of the pictorial content and its use in a comparison measure, ill-defined and prone to aliasing (among other problems).
Second, and even more critical, is the fact that the geometric noise renders the information that is vital for the comparison simply missing. In fact, it forces us to do what the square jigsaw puzzle literature has usually been avoiding deliberately, namely to use pictorial information \textit{further away} from the piece boundaries.
And third, the geometric noise also introduces uncertainty about the proper offset between neighboring pieces in the direction of the mates. In addition to Gur and Ben-Shahar~\cite{gur2017square}, who introduced the last consideration in their brick wall puzzle setup, only a handful of works address square piece puzzles with gaps between their pieces (i.e., eroded pieces that might have no direct contact), including Paumard et at.~\cite{paumard2020deepzzle} that employes a deep network to predict the position of the pieces, and Song et al.~\cite{songetal2023erodedgaps} that employs two types of deep networks and a genetic algorithm.

\begin{figure}[!ht]
    \centering
    \begin{tabular}{ccc}
        \includegraphics[width=0.30\columnwidth]{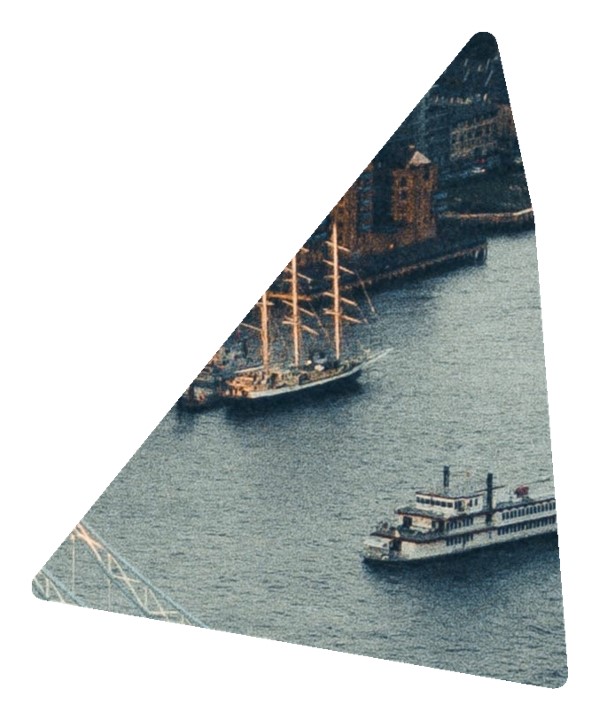}
        &
        \includegraphics[width=0.30\columnwidth]{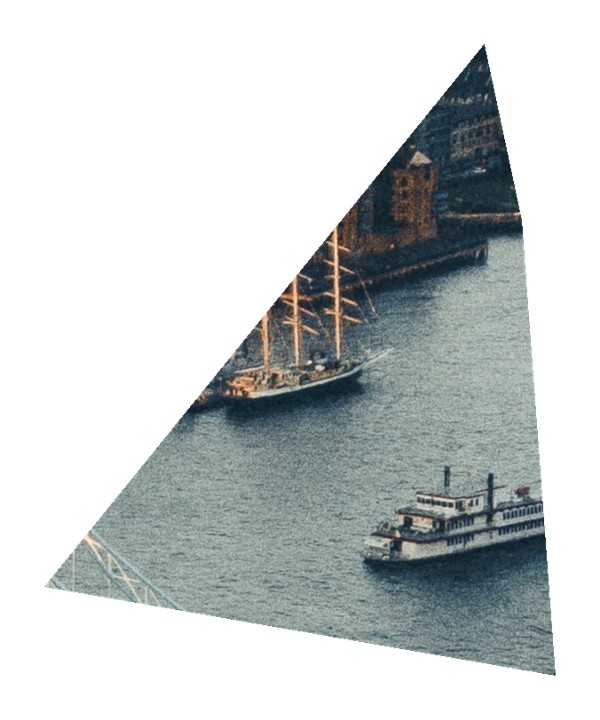}
        &
        \includegraphics[width=0.30\columnwidth]{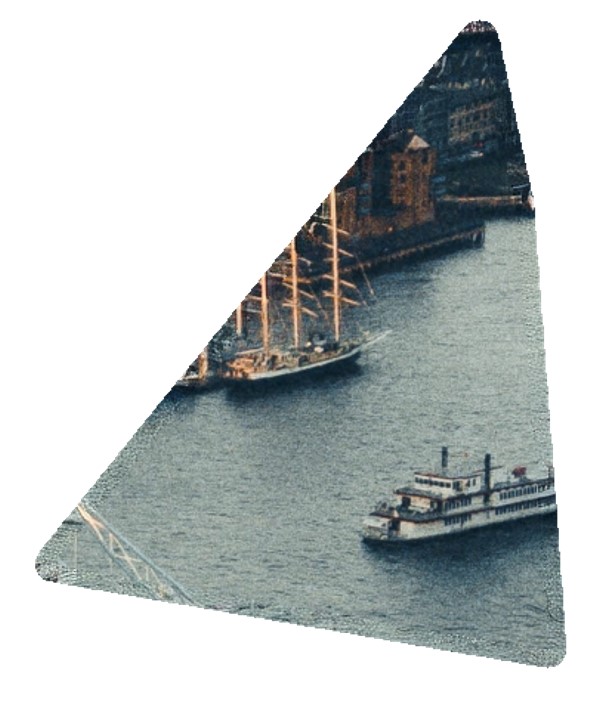}
        \\
        A & B & C
        \end{tabular}
        \CAPTION{
            Extrapolation of visual information.
            \textbf{A:} An original piece.
            \textbf{B:} A geometrically noisy piece.
            \textbf{C:} The extrapolated piece for extrapolation radius of $\varepsilon=25$ pixels. 
}
    \label{fig:border_extrapolate}
\end{figure}

To deal with all these problems simultaneously, and inspired by similar ideas in the literature~\cite{liu2011automated,derech2021solving}, we score a candidate mating $m = \{e_i^j, e_k^l\}$ by extrapolating the information of the two corresponding puzzle pieces $p_i$ and $p_k$ to a spatial band beyond their boundaries and thus obtaining ''dilated'' pictorial pieces on which a compatibility measure can be applied more safely. There are many ways of doing such extrapolation, but most of those we experimented with perform too poorly to provide a reliable visual outcome that in turn can facilitate reliable pictorial compatibility score $S(m) = S(\{e_i^j, e_k^l\})$ for mating $m$. 
In our work, we first applied the basic inpainting method due to Telea ~\cite{telea2004inpainting}. We then fed the results, after resizing and padding, to a pre-trained deep Stable-Diffusion network~\cite{Rombach_2022_CVPR} to extrapolate the pieces beyond their original boundaries, as far as a band whose thickness is defined by the bound $\varepsilon$ on the geometric noise. Fig.~\ref{fig:border_extrapolate} illustrates a selected result of the pictorial extrapolation and compares it to the original pictorial content. It goes without saying that in our case the available information is taken from the \enoisy (i.e., ``eroded'') piece while the pictorially extrapolated (i.e., ``diluted'') piece usually extends even beyond the boundaries of the original noiseless piece.

With the extrapolated pieces computed, one can conceive many different ways to measure the compatibility of any two mates even without knowing the details of the geometric noise that affected them. For example, we note that by design of the extrapolation procedure, there must be some overlapping content between the two extrapolated pieces. Hence, one can attempt to register the two pieces and find the relative Euclidean transformation that places them next to each other with proper pictorial overlap along the extrapolated boundaries. This also may provide some information about the localization of pieces in the reconstructed puzzle, but at the same time, this approach is very sensitive and prone to errors (as the overlapping extrapolated pictorial information available for registration is both scarce and hypothetical) and therefore does not make the global optimization from Sec.~\ref{sec:psy} redundant. Because of such observations, it may be more effective to design a scoring mechanism that does not pretend to localize the pieces but is rather \textit{invariant} to the relative transformation, for example by producing a \textit{scalar} score for the dissimilarity embodied in any candidate mating. And while numerous such measures can be developed, we currently elected to implement the following simple scheme.

\begin{figure}[!ht]
    \centering
    \begin{tabular}{cc}
        \cellcolor[gray]{0}
        \includegraphics[width=0.25\columnwidth]{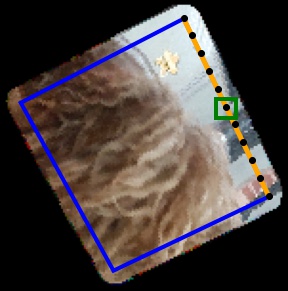}
        &
        \cellcolor[gray]{0}
        \includegraphics[width=0.25\columnwidth]{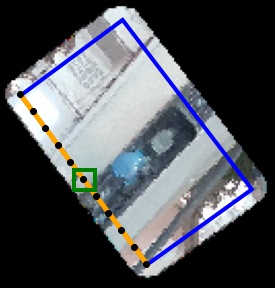}\\
        \end{tabular}
        \CAPTION{
             The pictorial scoring process in action.
             Two running windows (in green) are simultaneously scanning the extrapolated pictorial content of the two candidate mates (in orange). The sample points are shown in black.
             The differences in window averages at all corresponding locations are aggregated into a $G+1$ dimensional vector, whose $L_1$ norm serves as the dissimilarity measure $S(m)$. Note that the pictorial descriptor based on these window averages is invariant to the relative transformation, and although being simple it is descriptive enough to allow effective scoring and filtering or matings.
}
\label{fig:window}
\end{figure}

Given a candidate mating $m=\{e_i^j, e_k^l\}$ from two different pieces $p_i$ and $p_k$, we wish to compare the pictorial information around different corresponding points along $e_i^j$ and $e_k^l$. For that, we sample both edges an equal number of times from one end to the other and compare visual windows around corresponding samples.
More specifically, denote by $W(\vec{v}) \in \mathcal{R}^{h \times h}$ the square pixel window around position $\vec{v}$ and let $F(\vec{v}) \in \mathcal{R}$ be its average color across the channels, i.e., 
{\small
\begin{align*}
F(\vec{v}) = \frac{1}{|W| \cdot 3} \sum_{\vec{w} \in W(\vec{v})} \left( R(\vec{w}) + G(\vec{w}) + B(\vec{w}) \right) \;.
\end{align*}
}
To evaluate the pictorial affinity of the two edges, we sample both $e_i^j$ and $e_k^l$ evenly $G+1$ times, including at their vertices $(v_i^{j},v_i^{j + 1})$ and $(v_k^{l },v_k^{l + 1})$. We next consider all the mean color values of the windows $W(\vec{v})$ along either $e_i^j$ and $e_k^l$  as two vectors in a $G+1$ dimensional space and compute the $L_1$ norm of their difference. The result serves as our measure of dissimilarity $S(m)$. Formally,
{\scriptsize
\begin{align*}
S(m) =
\sum_{k = 0}^{G}
\left|
F\left(
  \frac{k}{G} \cdot  v_i^{i + 1} +
  \left(
    1 - \frac{k}{G}
  \right) \cdot v_i^{i}
\right)
-
F\left(
  \frac{k}{G} \cdot v_k^{l} +
  \left(
    1 - \frac{k}{G}
  \right) \cdot v_k^{l + 1}
\right)
\right| \;.
\end{align*}
}
Note that the running windows may not be completely synchronized in relative position since the edges may have slightly different noisy lengths. In addition, the possibly different transformations of the two pieces suggest that the two pictorial windows will exhibit different aliasing and thus slightly different pixel values.
However, the low pass filtering embedded implicitly in $S(m)$ and the fact that a pictorial descriptor based on scalar window averages is invariant to the different relative transformations of each piece, provide robustness to both confounds. Clearly, one can conceive numerous other ways of implementing compatibilities for polygonal pieces and future work is likely to put additional attention on this challenge. However, despite being relatively simple, the $S(m)$ measure is already descriptive enough to allow effective pictorial scoring of matings. A depiction of this process is provided in Fig.~\ref{fig:window}.

With a pictorial score for each candidate mating, and keeping in mind that $\tilde{M}$ denotes all matings that satisfy the noisy \textit{geometric} constraints, we now define the \textit{pictorially constrained mating set} $\tilde{M}_p$ by considering for each edge pair only the $T$ matings that scored the best (i.e., lowest) $S(m)$, with $T$ being some predefined number or an absolute percentile, that could depend on available computational or time resources.
Formally, if $H_T(X)$ denotes the set of $T$ matings with the lowest $S(m)$ score in a given set $X \subseteq \tilde{M}$, then $\tilde{M}_p$ is defined as follows
{\small
\begin{align*}
\tilde{M}_p =
\bigcup_{\{e \in E\}}  H_T(\{m | m \in \tilde{M} \wedge e \in m\})\;\;.
\end{align*}
}
%



It goes without saying that the higher the geometrical noise (expressed by its bounds $\varepsilon$ or alternatively, $\xi$), the more significant the pictorial compatibility and pictorial mating filtering become. This quickly allows a drastic (typically an order of magnitude or larger) decrease in the number of potential matings, thus making much larger puzzles potentially solvable. Once the pictorially constrained set $\tilde{M}_p$ is computed, the reconstruction process can proceed exactly as described for the apictorial case (cf. Sec.~\ref{chap:puzzle_reconstruction}), including the global considerations encapsulated in the hierarchical layered loopy constraints (Sec.~\ref{sec:noisy_reconstruction}).





\section{Evaluation metrics and experimental results}

This paper presented a new visual puzzle model, analyzed its properties, and suggested a solution scheme for both apictorial and pictorial variants. To test our approach, and having no prior work on crossing cuts or polygonal puzzles in the literature, our experimental evaluation focus on the formulation of performance metrics and reporting qualitative and quantitative results on the novel benchmark datasets presented in Sec.~\ref{sec:data_synthesis}. Note that these datasets include both pictorial and apictorial puzzles with varying global shape, different numbers of crossing cuts, and a range of noise levels. Results of the naive algorithm for ``clean'' puzzles are not reported as it always reconstructs the puzzles perfectly.


\subsection{Evaluation metrics}

As mentioned in Sec.~\ref{chap:puzzle_reconstruction}, under geometric noise it is unclear what is the desired position (i.e., Euclidean transformation) of each piece in the reconstructed puzzle, and the multi-body spring-mass system aspires to obtain a solution that optimizes an intuitive objective. It still remains to score such solutions as to allow their quantitative evaluation and comparison, and for that purpose, one can assume the availability of a ground truth solution against which the evaluation is performed. As discussed in Sec.~\ref{sec:the_puzzle}, any solution, be it the ground truth or one computed by a solver, constitutes both a mating graph and the Euclidean transformation of each piece, and thus the evaluation must take both into account. Unfortunately, this is not a straightforward task.

The evaluation of the mating graph is perhaps clearer, as we wish to compare two graph structures that could differ only in their set of links\footnote{Recall that we reserved the term 'edges' for the boundary segments of the puzzle pieces while the edges of the mating graph are termed `links' to avoid confusion.}. Inspired by the Neighbor Comparison Metric from the square jigsaw puzzle literature (e.g., \cite{cho2010probabilistic,pomeranz2011fully,sholomon2013genetic}) we therefore define an evaluation metric for the computed matings as an area-weighted precision and recall measures of the computed matings: 
{\small
\begin{align}
    Q_{precision} = 
    \frac{
        \sum_{{e_i^j, e_k^l} \in M_{gt} \cap M_{sol}}{
        \left(\left| A(p_i) \right| + \left| A(p_j)\right| \right)}
    }
    {
        \sum_{{e_i^j, e_k^l} \in M_{gt}}{
        \left( \left| A(p_i)\right| + \left|A(p_j)\right| \right)}
    }
    \\
    Q_{recall} = 
    \frac{
        \sum_{{e_i^j, e_k^l} \in M_{gt} \cap M_{sol}}{
        \left(\left| A(p_i)\right| + \left| A(p_j)\right| \right)}
    }
    {
        \sum_{{e_i^j, e_k^l} \in M_{sol}}{
        \left(\left| A(p_i)\right| + \left| A(p_j)\right| \right)}
    }
\label{eq:mating_metrics}
\end{align}
}
where $M_{gt}$ are the ground truth matings, $M_{sol}$ are the matings of the solution found by the reconstruction algorithm, and $A(p_i)$ represents the \textit{region} (i.e., set of points) of piece $p_i$ in its final pose, as in Eq.~\ref{eq:Qoverlap}. 

The scoring of the Euclidean transformation of pieces is more tricky. For example, we observe that even qualitatively perfect solutions by the spring-mass system may differ by a global Euclidean transformation due to arbitrary choice of a coordinate system in the representation of the pieces (cf. Sec.~\ref{sec:the_puzzle} and Fig.~\ref{fig:cc_el_df}A). The situation becomes significantly more ambiguous once the solutions are not perfect (as is always the case under noise) and scoring needs to consider the placement of each and every piece of the puzzle separately.

With such challenges in mind, assume we have a solution we wish to score, i.e., the mating graph and the Euclidean transformation of all pieces in a reconstructed puzzle.  Let $\vec{u\,}_i^j$ be the vertices of piece $p_i$ in the ground truth and $\vec{v\,}_i^{j}$ the corresponding vertices of $\tilde{p}_i$ in the obtained solution.
We first globally align the obtained solution with the ground truth before comparing the placement of individual pieces. In other words, we wish to find a global Euclidean transformation $(R^*,t^*)$ that aligns the reconstructed pieces ``as close as possible'' to the ground truth so they can be compared.
To do so we employ  SVD for Least-Squares Rigid Motion~\cite{sorkine2017least} to solve the following weighted minimization
{\small
\begin{align}
    (R^*,t^*) = \underset{(R,t)}{\mathrm{argmin}}
    \sum_{i=1}^n \sum_{j=1}^{N_i}
    w_i \lVert (R \vec{v\,}_i^{j} + t) - \vec{u\,}^j_i \rVert^2
\label{eq:min_svd_rigid}
\end{align}
}
where the weights $w_i$ are set to be proportional to the area of each piece  to reflect the greater importance of larger pieces on the shape of the puzzle, i.e.,
{\small
\begin{align}
    w_i = \frac{\left|A(p_i)\right|}{\sum_{k=1}^n \left| A(p_k) \right|}\;.
\label{eq:min_svd_rigid_weights}
\end{align}
}

Qualitatively, the more similar the mating graphs of the reconstructed puzzle and the ground truth, the better the global alignment will be and thus a better (i.e., smaller) score will be achieved by the optimal global transformation $(R^*,t^*)$.
However, Eq.~\ref{eq:min_svd_rigid} in itself is not a convenient metric for the quality of the overall solution since it depends on the specific puzzle evaluated. In that sense, that score may allow the comparison of different solutions (say by different solvers) to the \textit{same} puzzle, but it provides hardly any insights about the solution quality to an arbitrary puzzle, it does not allow ordering the solutions of different puzzles, and it prohibits aggregation of many solutions into statistical measures on whole datasets.

\begin{figure}[!ht]
    \centering
    \includegraphics[width=0.7\columnwidth]{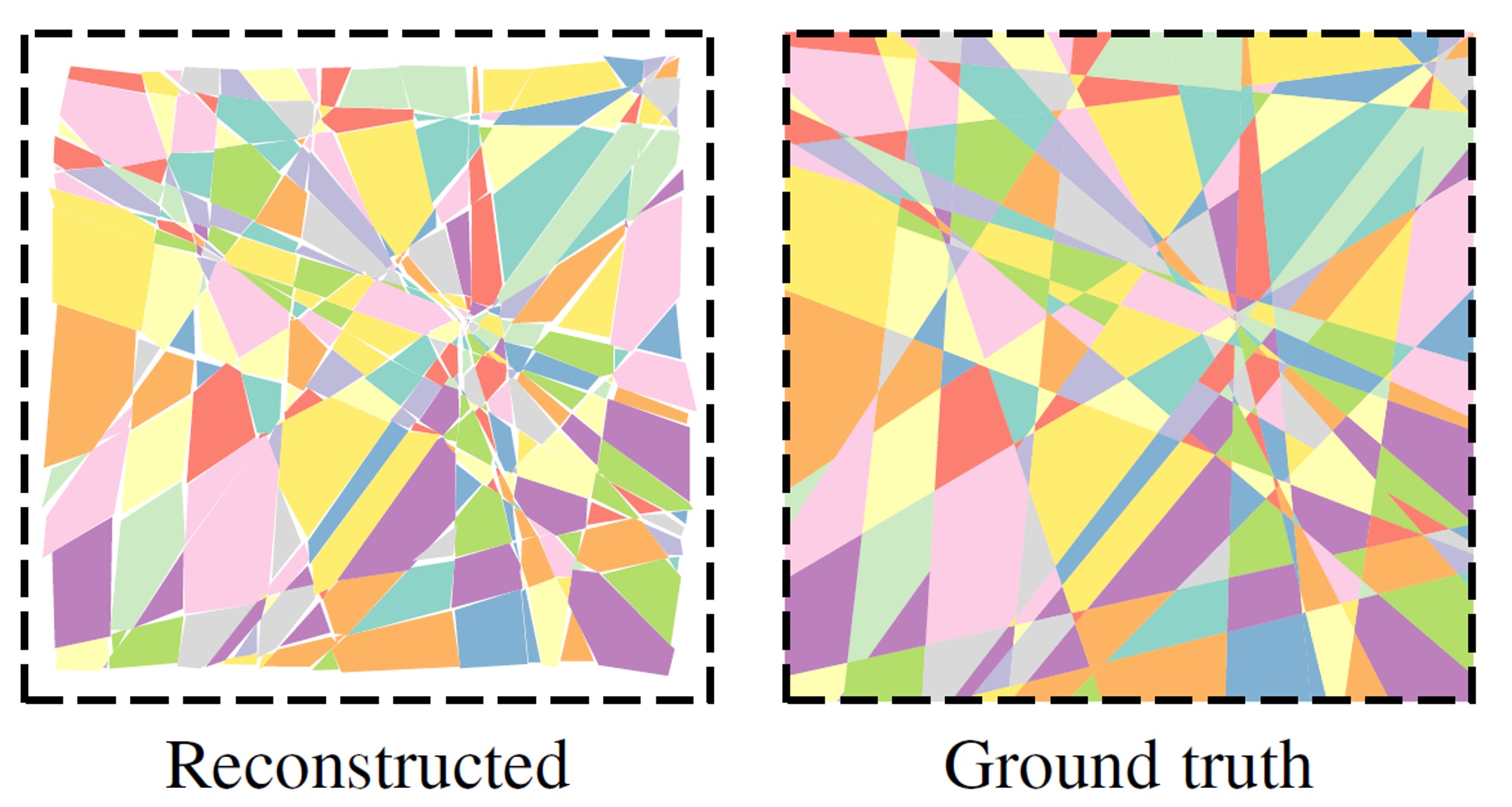}
    \CAPTION{
            Being conservative, the performance metric can score a solution low even if intuitively it is good. In this 30-cut and 326-piece noisy puzzle, the reconstructed solution is qualitatively similar to the noiseless ground truth and should be considered a success. And yet, this solution scores just 0.56/1.00 by Eq.~\ref{eq:solution_score} because the noise is rather big, the gaps created between the eroded pieces is significant, and the springs tend to pull the pieces closer over the gaps and clump them in the center of the noiseless puzzle boundary (dashed line).
    }
\label{fig:pos_bench_B}
\end{figure}

To overcome all these difficulties, we seek a more informative measure that is \textit{normalized} to some canonical range (say $[0,1]$). We therefore consider the degree of area overlaps between the pieces in the solution vs. their ground truth counterpart, after the two solutions have been aligned with Eq.~\ref{eq:min_svd_rigid}. 
Formally, if $\tilde{p}_i'$ is the noisy piece $\tilde{p}_i$ \textit{after} being placed in the reconstructed puzzle, i.e.,
\begin{align*}
\tilde{p}_i' = \left\{R_i \cdot \vec{v\,}_i^1 + \vec{t\,}_i, R_i \cdot \vec{v\,}_i^2 + \vec{t\,}_i,\ldots , R_i \cdot \vec{v\,}_i^{N_i} + \vec{t\,}_i\right\}
\end{align*}
then we define
{\small
\begin{align}
Q_{pos}(R_1, t_1 ,\dots R_n t_n) =
    \sum_{i=1}^n
    w_i \cdot \frac{
    \left|
    A(p_i) \cap A(\tilde{p}_i')
    \right|
    )
    }{
    \left|
    A(\tilde{p}_i)
    \right|
    } \;\;.
\label{eq:solution_score}
\end{align}
}
where the weights are as defined in Eq.~\ref{eq:min_svd_rigid_weights}. This measure is conservative, in the sense that high scores always imply good solutions, but good solutions do not always receive high scores, as illustrated in Fig.~\ref{fig:pos_bench_B}. Future research may wish to explore improved metrics for such circumstances.

In summary, we use the mating measures $Q_{precision}$ and $Q_{recall}$ from Eq.~\ref{eq:mating_metrics} and the positions metric $Q_{pos}$ from Eq.~\ref{eq:solution_score} as our quantitative measures for evaluating puzzle solutions. Next, we apply them to different test cases.


\subsection{Experimental evaluation of piece positioning}

We first tested our crossing cuts solver for positioning puzzle pieces while assuming the matings are known, i.e., we only evaluated the degree to which the abstraction as a multi-body spring-mass system (Sec.~\ref{sec:psy}) provides desired results, both qualitatively and quantitatively. Clearly, for this evaluation, it is irrelevant if the puzzle is pictorial or apictorial.

To implement this test we extracted from DB2 and DB3 puzzles their set $\tilde{P}$ of noisy pieces \textit{and} the ground truth matings $M_{gt}$, applied the positioning system (Sec.~\ref{sec:psy}) and obtained the euclidean transformation $(R_i, t_i)$ of each piece $\tilde{p}_i$ in the solution.  Evaluation of the result was then based on Eq.~\ref{eq:solution_score}.

\begin{figure}[h!]
    \centering
    \includegraphics[width=0.8\columnwidth]{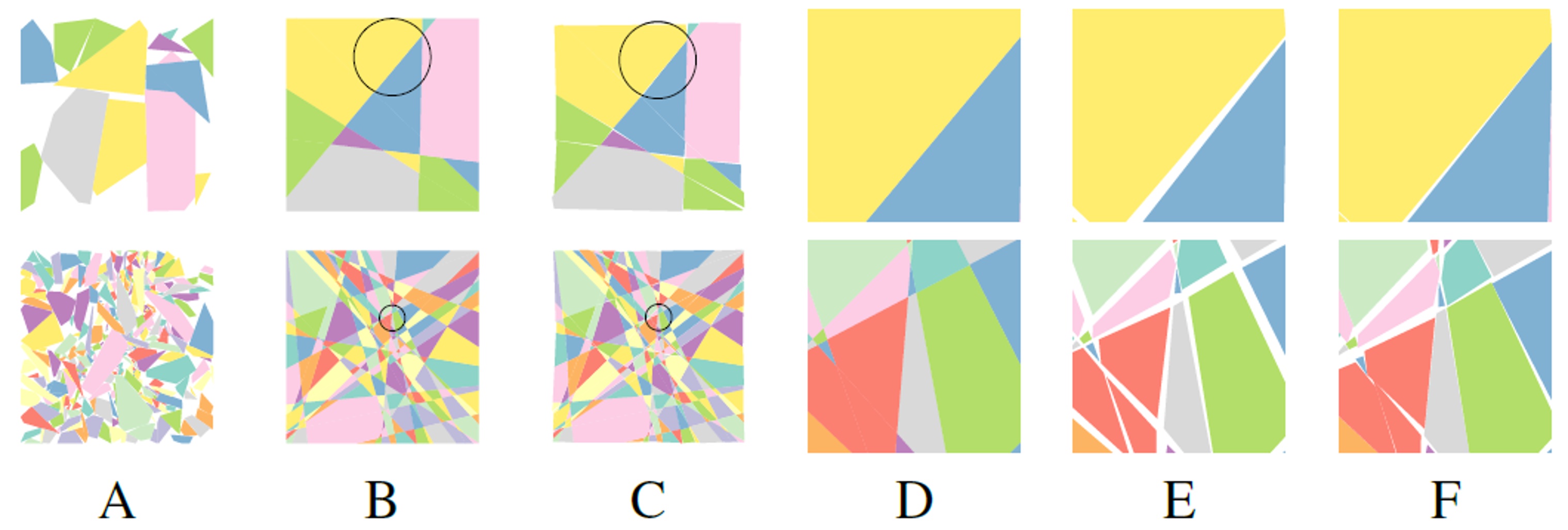}
    \CAPTION{
            Qualitative performance of piece positioning for known matings. Since pictorial information plays no role in this test, we omit pictorial examples.
            {\bf A:} Initial random placement of pieces. Keep in mind that despite the messy organization the mates (i.e., the springs) are set correctly and we are interested to examine whether such a random initial state can lead to undesired local minima.
            {\bf B:} Ground truth assembly.
            {\bf C:} Computed assembly after convergence of the second phase of the spring-mass system. The circle marks the area shown in the closeup section in the next column.
            {\bf D:} A closeup on the original ground truth.
            {\bf E:} A closeup on the corresponding section of the noised ground truth.
            {\bf F:} A closeup on the corresponding section of the reconstructed assembly.
            }
\label{fig:position_test_qualitative}
\end{figure}

\begin{figure}[h!]
    \centering
    \includegraphics[width=0.95\columnwidth]{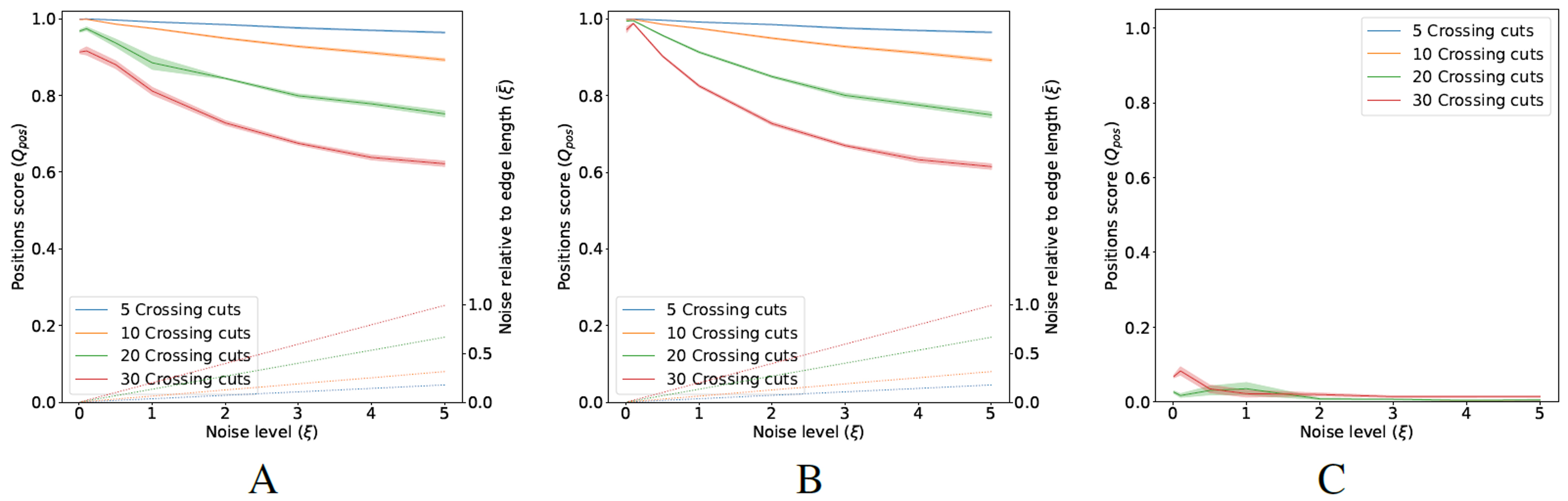}
    \CAPTION{
            Quantitative performance of piece positioning for known matings. The dotted lines show the value of $\bar{\xi}$ to appreciate how high it can get.
            {\bf A:} Average positioning score of the two-phase dynamical system as a function of noise level, computed from 10 random puzzles for selected numbers of crossing cuts/pieces. Shaded bands are $\pm 1$ SE.
            {\bf B:} Positioning score after the application of the second phase of the dynamical system from the ground truth initial positions. 
            {\bf C:} The difference of the first two graphs indicates that in most cases the initial state of the positioning systems has no effect, except possibly for small noise cases. 
            }
\label{fig:pos_bench_A}
\end{figure}

The first of these evaluations examined the ability of the spring-mass system to converge to the desired spatial configuration from the same initial state suggested by the algorithm, namely with the initial pose (position and rotation) of each piece chosen randomly inside the arena. Recall that the first run of the dynamical system allows pieces to overlap (a near-certain event under random piece positions). Upon convergence the same system is restarted but now while piece overlaps are prohibited. Fig.~\ref{fig:position_test_qualitative} shows the initial and final configurations next to the ground truth of selected puzzles, and Fig.~\ref{fig:pos_bench_A}A presents the quantitative score $Q_{pos}$ for puzzles with various noise levels. We were particularly interested in examining if the system might converge to improper local minima that depart qualitatively from the desired organization of pieces. This never happened and as shown in the examples, convergence is qualitatively correct even in the most complex cases.

The second test aims to empirically quantify a lower bound on the deviation from ground truth positions induced by the spring-mass system. To do so we applied the positioning computational to the same set of puzzles, though this time the initial state of the pieces was the ground truth position of the noisy pieces (where $Q_{pos}=1$), and the only computational step executed is the second phase where overlaps are prohibited. Intuitively, there could not be a better initial state for the pieces before the computation begins and thus Fig.~\ref{fig:pos_bench_A}B represents the best positioning scores possible.  Fig.~\ref{fig:pos_bench_A}C shows the performance difference between an optimal and a random initial states, thus representing how the deviation from the optimal positions is reflected in the positioning score. Note that in most cases the initial state of the positioning systems has negligible effect.

\subsection{Experimental evaluation of \textit{apictorial} puzzle solutions}

With a system to evaluate the solutions by the multi-body spring-mass system established, we turn to evaluate the full algorithmic solution under \textit{unknown} matings (Sec.~\ref{sec:noisy_reconstruction}), where the input is just the noisy pieces $\tilde{P}$ (and the bound on the noise level $\xi$) while the output includes both the matings graph $M$ and the Euclidean transformations $(R_i, t_i)$ of each piece $\tilde{p}_i$ in the solution.  Here we first focus on apictorial puzzles and seek to evaluate both parts of the solution using the two evaluation measures, i.e., both the precision and recall from Eq.~\ref{eq:mating_metrics} and $Q_{pos}$ from Eq.~\ref{eq:solution_score}.

\begin{figure}[h!]
    \centering
    \includegraphics[width=0.85\columnwidth]{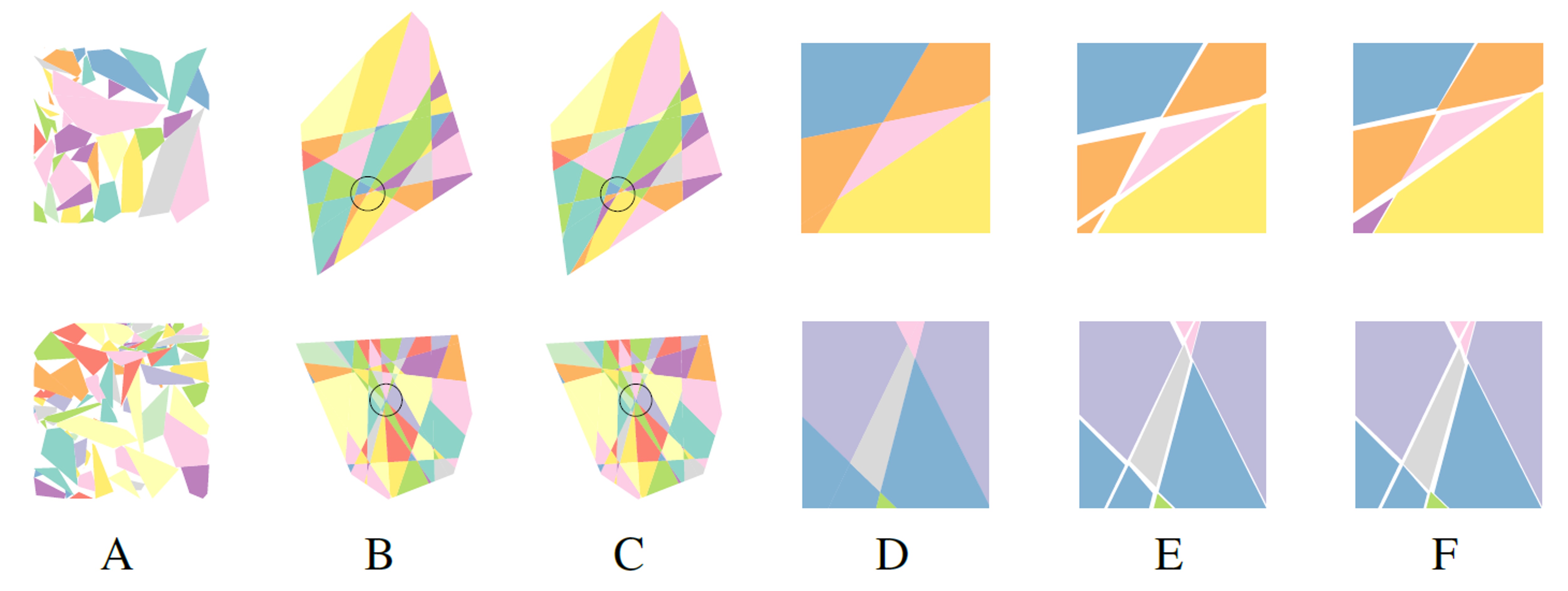}
    \CAPTION{
        Examples of successful reconstruction results.
        \textbf{A:}~The unordered bag of puzzle pieces that the solver receives as input.
        \textbf{B:}~The ground truth assembly of (noiseless) pieces
        \textbf{C:}~The reconstruction result of the noisy puzzle.
        \textbf{D, E, F:}
                A zoomed area of the noiseless ground truth, the noisy ground truth, and the solution. Unlike in the ground truth,  the pieces in the noisy ground truth and in the solution do not abut each other, as would be expected because of the noise in their shape. In the solution, the gaps are determined automatically by the multi-body mechanical system while minimizing the global elastic energy of the springs.
    }
    \label{fig:exp_res_good}
\end{figure}

First qualitatively, Fig.~\ref{fig:exp_res_good} presents visual examples of successful reconstructions of selected puzzles of different global shapes, number of cuts, and noise levels. Note how the solution remains loyal to the (unknown) ground truth puzzle both in terms of its global shape and he organization of the pieces. The closeup insets show how the positioning system places the pieces at some distance, as would be desired due to the noise. Indeed, the configuration is not necessarily identical and in fact slightly perturbed relative to the ground truth, where the vertices' position (and thus the gaps) are determined automatically by the multi-body mechanical system while minimizing the energy of the springs.

Fig.~\ref{fig:pos_bench_C}A shows aggregated quantitative performance selected subsets of DB2. These results indicate that the mechanism based on the hierarchical loops, loop ranking, and loop merging obtains excellent results. Still, since the problem is intractable we cannot expect the heuristics to provide a perfect solution always, and Fig~\ref{fig:pos_bench_C}B-D exemplifies one such unlikely failure.

\begin{figure}[h!]
    \centering
    \includegraphics[width=0.9\columnwidth]{fig34.jpg}
    \CAPTION{
        Reconstruction results on selected puzzles from DB2.
            \textbf{A:} Results tested on 4 subsets of DB2, each containing 10 selected puzzles, sampled randomly from the original dataset with 5-50 cuts, $\xi$ (noise level relative to the puzzle diameter) varying between 0 to 0.25\% and $\bar{\xi}$ (noise level relative to the average edge length) varying between 0 to 1.36\%.
            The results show the positioning score (in blue), precision score (in orange), and recall score of the matings (in green).
    \textbf{B-D:}An uncommon reconstruction failure.
                 Shown are the unordered bag puzzle pieces that the solver receives as input (B), the ground truth solution (C), and the faulty reconstruction result (D).
    }
    \label{fig:pos_bench_C}
\end{figure}


It should also be mentioned that although the jigsaw problem is NP-complete, and thus complete or optimal solvers are expected to be exponential, relying on the crossing cuts geometrical constraints, and the looping and merging heuristics, can decrease the practical complexity significantly. Several of the steps become polynomial, and in particular, establishing 0-loops is bounded by a $4^{th}$ order polynomial of the number of pieces. However, \textit{in the noisy case} the number of possible mating combinations and the search in the merging step (cf. Sec.~\ref{sec:noisy_reconstruction}) remain exponential, where in practice they are influenced by the noise bound (cf. Sec.~\ref{sec:stat_potential}) and the number of cuts (cf. Sec.~\ref{chap:statistic}). This is why pictorial constraints are so valuable, and the better they can be utilized and reduce the number of candidate matings, the more efficient the solver can become for a given puzzle.


\subsection{Experimental evaluation of \textit{pictorial} puzzle solutions}
\label{sec:exp_pictorial}

\begin{figure}[h!]
    \centering
    \begin{tabular}{cccc}
        \cellcolor[gray]{0} \includegraphics[width=0.16\columnwidth]{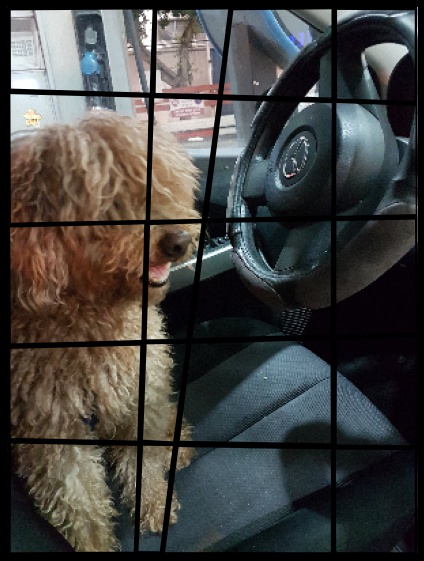} &
        \cellcolor[gray]{0} \includegraphics[width=0.16\columnwidth]{sqpic_7_ground_truth_n.jpg} &
        \raisebox{18pt}{\cellcolor[gray]{0} \includegraphics[width=0.16\columnwidth]{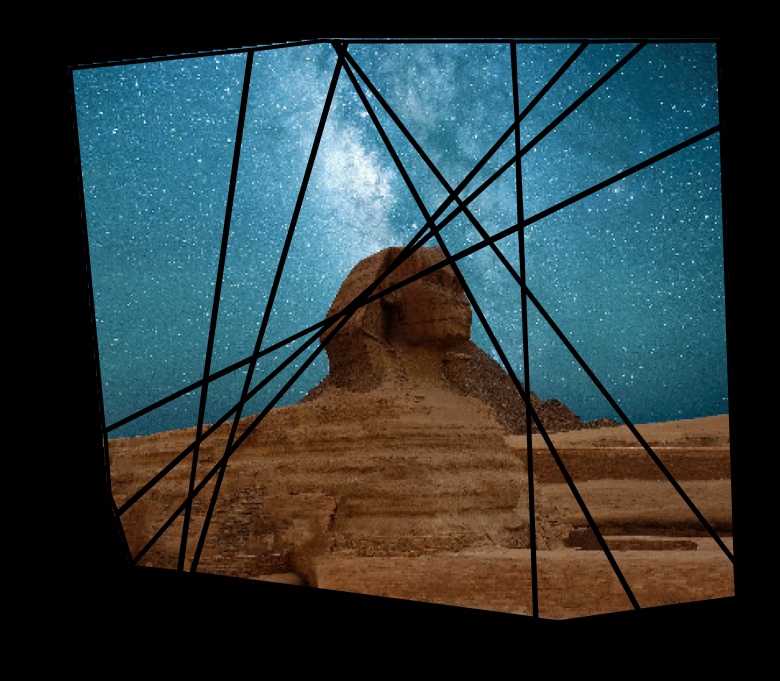}} &
        \raisebox{09pt}{\cellcolor[gray]{0} \includegraphics[width=0.16\columnwidth]{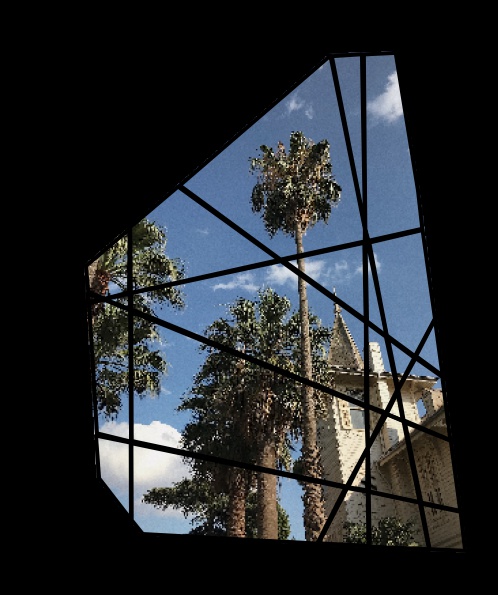}}
        \\
        
        \cellcolor[gray]{0} \includegraphics[width=0.16\columnwidth]{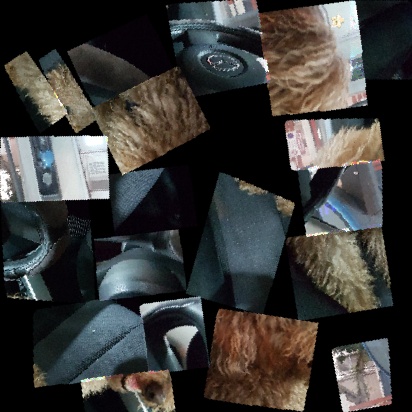} &
        \cellcolor[gray]{0} \includegraphics[width=0.16\columnwidth]{sqpic_7_Pictorial_puzzle_bag.jpg} &
        \cellcolor[gray]{0} \includegraphics[width=0.16\columnwidth]{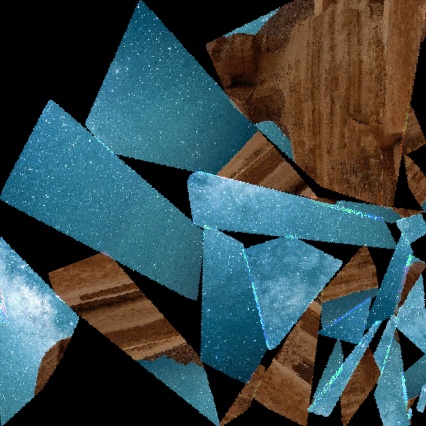} &
        \cellcolor[gray]{0} \includegraphics[width=0.16\columnwidth]{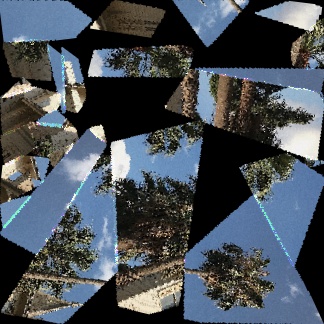}
        \\
        
        \raisebox{18pt}{\cellcolor[gray]{0} \includegraphics[width=0.16\columnwidth]{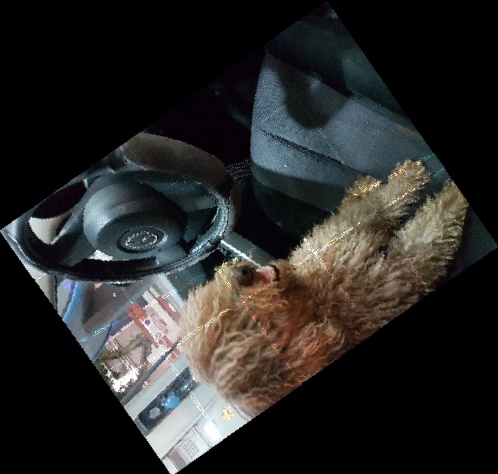}} &
        \raisebox{30pt}{\cellcolor[gray]{0} \includegraphics[width=0.16\columnwidth]{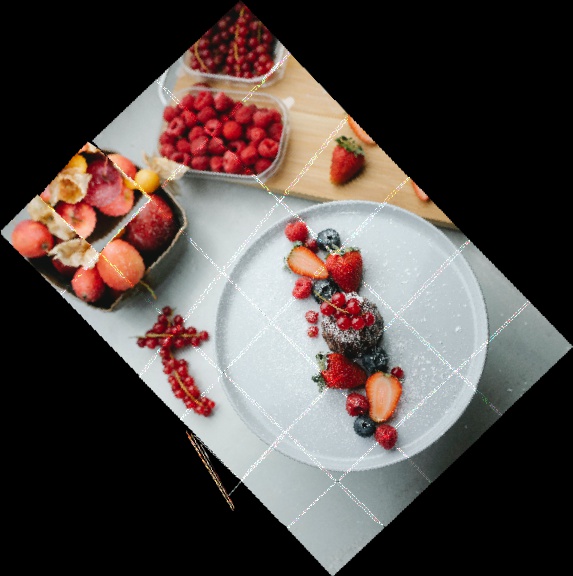}} &
        \raisebox{15pt}{\cellcolor[gray]{0} \includegraphics[width=0.16\columnwidth]{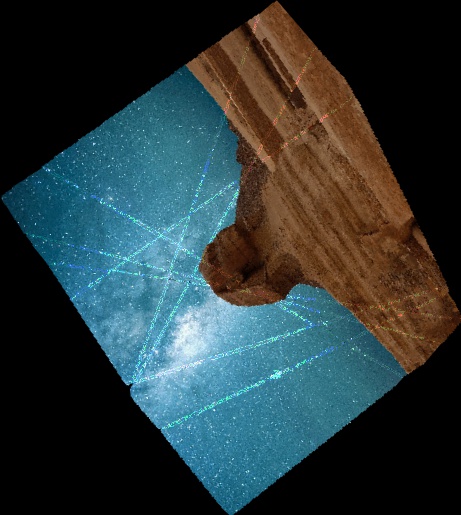}} &
        \cellcolor[gray]{0} \includegraphics[width=0.16\columnwidth]{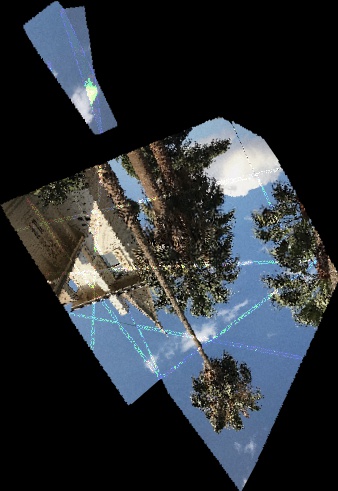}
        \\
        A & B & C & D \\
        \end{tabular}
        \CAPTION{
                 Selected examples of pictorial puzzle solutions ({\bf A,B} - perturbed square jigsaw puzzles, {\bf C, D} - general crossing cuts puzzles) having various numbers of cuts, piece size,  noise levels, and visual contents.
                 Top row shows the original image/puzzle. The second row represents the puzzle that was submitted to the solver, and the bottom row shows the solution (slightly scaled down to fit the allotted space). Recall that
                 solutions can be perfect up to a global Euclidean transformation. {\bf D} shows one unusual failure caused by $T$ being too restrictive thus excluding the correct mating.}
\label{fig:qualitative_pictorial_evaluation}
\end{figure}

\begin{figure}[h]
    \centering
    \begin{tabular}{cc}
        \includegraphics[width=0.30\columnwidth]{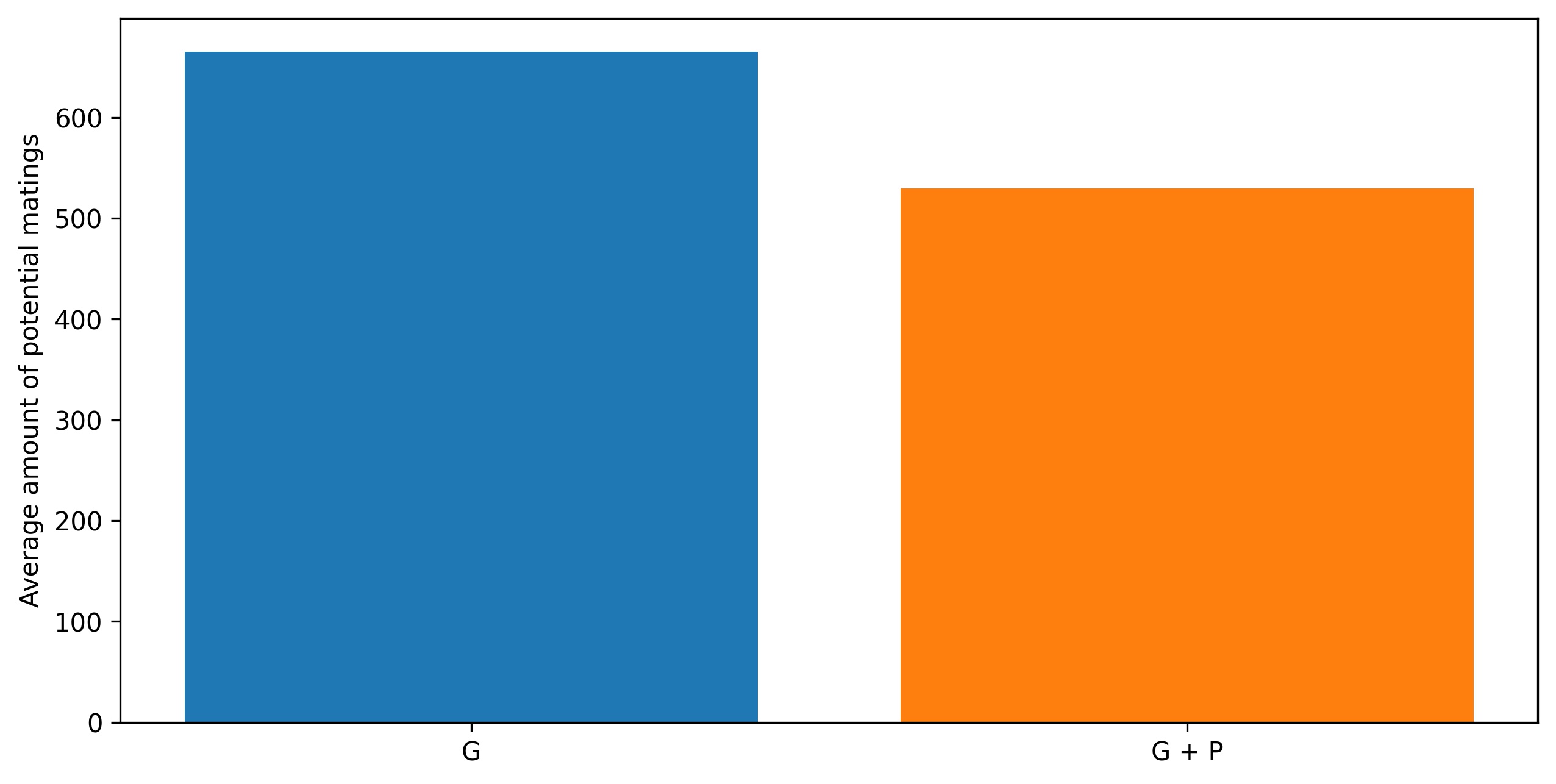} &
        \includegraphics[width=0.30\columnwidth]{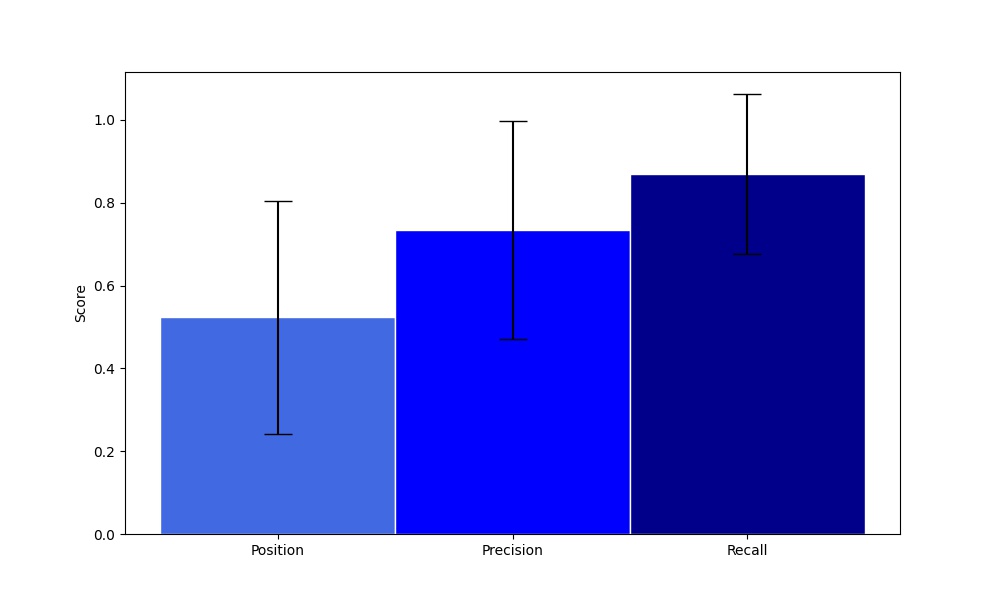} 
        \\
        A & B
        \end{tabular}
        \CAPTION{
                 Pooled quantitative performance on pictorial puzzles.
                 {\bf A:} Averaged on 10 DB3 puzzles, applying the pictorial constraints (P) 
                          on top of the geometrical ones (G) saves $\tilde 20\%$ of the original potential matings to test. This rate is of course based on the method used to filter the potential matings, based on their pictorial compatibility scores, and may vary accordingly. In this case, all of the potential matings scored above a certain value (30\%) were kept.
                 {\bf B:} Precision and recall (Eq.~\ref{eq:mating_metrics}) and position score (Eq.~\ref{eq:solution_score}) for the pictorial puzzles from DB3. It should be noted that in cases where real puzzle matings had an unusually low pictorial compatibility score, they were filtered out of the potential matings, leading the solver to eventually fail to reconstruct the puzzle as a whole. 
                 }
\label{fig:quantitative_pictorial_evaluation}
\end{figure}

We finally turn to examine the reconstruction of pictorial puzzles and assess the role of pictorial constraints using puzzles from DB3 and DB4. Recall from Sec.~\ref{sec:data_synthesis} that DB4 is a general pictorial puzzle set, while DB3 is designed to play down the role of geometrical constraints by having pieces whose edge length histogram is sharper (a condition that implies that each mate will have many more geometrically compatible matches). In such puzzles, the number of matings that satisfy constraints $\tilde{C}_1$ and $\tilde{C}_2$ is approaching the unfiltered set of matings, and thus the number of 0-loops, hierarchical loops, and possible geometrical solutions has easily overwhelmed the memory resources of the hardware we used for evaluation, which was a desktop computer with a 12th Gen Intel(R) Core(TM) i7-12700K Processor with a base clock speed of 3.60 GHz, and 32.0 GB RAM. 
Towards this end, we tested 10 puzzles from DB3 and DB4, all of which were solved only when the pictorial content was considered too. 
Selected qualitative solutions are shown in Fig.~\ref{fig:qualitative_pictorial_evaluation} while the average saving in potential matings due to the pictorial constraints are depicted in Fig.~\ref{fig:quantitative_pictorial_evaluation}A. 
These results are designated preliminary both because the pictorial filter is still simple, and because our present algorithm is not designed to deal with missing pieces, a condition that applies to many puzzles in DB4, once the level of the applied noise is increased.

 Next to several successful solutions, Fig.~\ref{fig:qualitative_pictorial_evaluation} shows one rare failure result. Failures such as this could happen when the value of $T$ is too small and correct matings are discarded by the pictorial filter. Indeed, although the filter eliminated many mating candidates, the aggregated quantitative performance on the tested puzzles from both DBs is good, as reported in Fig.~\ref{fig:quantitative_pictorial_evaluation}B,C. Performance on DB3 puzzles is slightly lower since as implied above, they can hardly utilize the geometrical constraints.

\section{Conclusions and future work}

We introduced a new jigsaw puzzle model and analyzed its properties and the inherent challenges in solving them once pieces are perturbed with noise. To cope with such difficulties and keep the problem tractable, we abstracted it as a multi-body spring-mass dynamical system method endowed with hierarchical loop constraints and a merging process of layered puzzle loops. Results exhibit excellent solving power but also suggest that future work should utilize pictorial data on the pieces much more strongly to drastically reduce the number of potential mates per edge and turn the problem more tractable and thus truly suited for real-life applications. This work also introduces a new class of puzzle generation models that are partially constrained, well formulated, and have enough expressive power to allow more real-life applications while being subjected to more rigorous analysis. We hope this type of thinking about ``restricted modeled puzzles'' can expand puzzle-solving literature to new directions, where future work should also address even more general (but formal) generation processes and noise models, as well as how to handle missing pieces in all these cases.


\section*{Acknowledgements} 

This work has been funded in part by the European
Union’s Horizon 2020 research and innovation
programme under grant agreement No 964854 (the RePAIR project). We also thank the
Helmsley Charitable Trust through the ABC Robotics
Initiative and the Frankel Fund of the Computer Science
Department at Ben-Gurion University for their generous support.

\renewcommand{\appendixname}{Appendix}
\begin{appendices}


\section*{Appendix - Glossary}          
\label{appendix:glossary}   

\noindent
The following summarizes the different symbols and notations used in the paper, sorted alphabetically when applicable.  

{\centering
\renewcommand{\arraystretch}{0.85}  
\begin{longtable}{| p{.1\textwidth} | p{.85\textwidth} |}
\hline
\textbf{Symbol}         &  \textbf{Meaning} \\ \hline 

$a$                     &  Number of crossing cuts that generate a puzzle  
\\
$A(p_i)$                & The spatial region of piece $p_i$ in its pose $(R_i,\vec{t\,}_i$ in the puzzle 
\\
$c_i$                   &  cut $i$ that participated in generating the puzzle  
\\
$Cuts$                  &  The set of cuts that generated a puzzle  
\\
${C}_1,\tilde{C}_1 $    & The mate length constraint under ideal and noisy conditions
\\
${C}_2,\tilde{C}_2$     & The mate angle constraint under ideal and noisy conditions
\\
$D$                     & puzzle diameter (distance between furthest vertices)    
\\
$\Delta\Theta_e(L, \varepsilon)$ & The bound of the orientation difference between $\measuredangle \tilde{e}$ and $\measuredangle e$, i.e., between the orientation of the \enoisy edge and its original noiseless edge of length $L$   
\\
$E$                     & The set of all piece edges of the puzzle  
\\
$E_i$                   & The set of edges of piece $i$ of the puzzle  
\\
$e_i^j$                 & Edge $j$ from $\vec{v\,}_i^j$ to $\vec{v\,}_i^{j+1}$ of piece $i$ of the puzzle  
\\
$\mathcal{E}$           & The edges (line segments between nodes) of  $\mathcal{G}_{puzzle}$   
\\
$\varepsilon$           & Bound on geometric noise in absolute units   
\\
$\vec{\epsilon\,}_i^j$  & Perturbation vector of vertex $j$ of piece $i$ due to the geometric noise 
\\
$G_M$                   &  The mating graph of the puzzle
\\
$\mathcal{G}_{puzzle}$  & The planar graph that represents a synthesized crossing graph puzzle    
\\
$L,\tilde{L}$           & The length of a clean and noisy edge, respectively   
\\
$\mathcal{L}$           & The bag of puzzle loops obeying the geometrical constraints 
\\
$M$                     & The set of all matings that constitute a mating graph $G_M$
\\
$M_{loop}$              & The set of all matings in a hierarchical loop 
\\
$\tilde{M}$             & The set of all matings candidates that satisfy $\tilde{C}_1, \tilde{C}_2$
\\
$\tilde{M}_p$           & The set of all matings candidates that satisfy $\tilde{C}_1, \tilde{C}_2$ and ranked best pictorially
\\
$m_q$                   & Mating $q$ in a mating graph  
\\
$N_i$                   &  Number of vertices (and edges) in piece $i$ of the puzzle   
\\
$P$                     &  The set of puzzle pieces  
\\
$P_{loop}$              & The set of pieces of an $x$-loop
\\
$p_i$                   &  Piece $i$ of the puzzle   
\\
$\tilde{p}_i$           &  \enoisy piece $i$ of the puzzle   
\\
$Q_{loop}$               & The quality score of a hierarchical loop 
\\
$Q_{pos}$               & The normalized quality score (in $[0,1]$) for piece positions in a puzzle solution  
\\
$R_i$                   & The rotation (matrix) applied to the vertices of piece $i$ 
\\
$S(m)$                  & Pictorial compatibility score of mating $m$   
\\
$\vec{t\,}_i$           & The translation (vector) applied to the vertices of piece $i$   
\\
$V_i$                   &  The set of vertices of pieces $i$ of the puzzle   
\\
$\vec{v\,}_i^j$         & Vertex $j$ of piece $i$ of the puzzle, ordered clockwise   
\\
$\measuredangle \vec{v}$& Angle of the vector $\vec{v}$   
\\
$\mathcal{V}$           & The nodes (intersection points) of  $\mathcal{G}_{puzzle}$ 
\\

$\xi$                   & Bound of geometric noise relative to puzzle diameter  
\\
$\bar{\xi}$             & Bound of geometric noise relative to average (expected) edge length  
\\
\hline
\caption*{}
\label{tab:glossary}
\end{longtable}
}\end{appendices}

\clearpage
\bibliographystyle{acm}
\bibliography{references}

\end{document}